\documentclass[journal]{IEEEtran}

\usepackage{cite}

\ifCLASSINFOpdf
  \usepackage[pdftex]{graphicx}
  \graphicspath{{./fig/}}
  \DeclareGraphicsExtensions{.pdf}
\else
\fi

\usepackage{algorithmic}

\ifCLASSOPTIONcompsoc
 \usepackage[caption=false,font=normalsize,labelfont=sf,textfont=sf]{subfig}
\else
 \usepackage[caption=false,font=footnotesize]{subfig}
\fi

\usepackage{algorithm}
\usepackage{amsfonts}
\usepackage{color}
\usepackage{footnote}
\usepackage[inline]{enumitem}
\usepackage{mathtools}
\usepackage{textcomp, gensymb}
\usepackage{booktabs}
\usepackage{threeparttable}
\usepackage{academicons}
\usepackage{scalerel}
\usepackage{tikz}
\usepackage{ragged2e}
\usepackage{color,soul}
\usepackage{xcolor}

\usetikzlibrary{svg.path}

\definecolor{orcidlogocol}{HTML}{A6CE39}
\tikzset{
  orcidlogo/.pic={
    \fill[orcidlogocol] svg{M256,128c0,70.7-57.3,128-128,128C57.3,256,0,198.7,0,128C0,57.3,57.3,0,128,0C198.7,0,256,57.3,256,128z};
    \fill[white] svg{M86.3,186.2H70.9V79.1h15.4v48.4V186.2z}
                 svg{M108.9,79.1h41.6c39.6,0,57,28.3,57,53.6c0,27.5-21.5,53.6-56.8,53.6h-41.8V79.1z M124.3,172.4h24.5c34.9,0,42.9-26.5,42.9-39.7c0-21.5-13.7-39.7-43.7-39.7h-23.7V172.4z}
                 svg{M88.7,56.8c0,5.5-4.5,10.1-10.1,10.1c-5.6,0-10.1-4.6-10.1-10.1c0-5.6,4.5-10.1,10.1-10.1C84.2,46.7,88.7,51.3,88.7,56.8z};
  }
}

\newcommand\orcidicon[1]{\href{https://orcid.org/#1}{\mbox{\scalerel*{
\begin{tikzpicture}[yscale=-1,transform shape]
\pic{orcidlogo};
\end{tikzpicture}
}{|}}}}

\usepackage{hyperref} %

\makesavenoteenv{tabular}

\makeatletter
\newcommand*{\bdiv}{%
  \nonscript\mskip-\medmuskip\mkern5mu%
  \mathbin{\operator@font div}\penalty900\mkern5mu%
  \nonscript\mskip-\medmuskip
}
\makeatother

\renewcommand{\algorithmicrequire}{\textbf{Input:}}
\renewcommand{\algorithmicensure}{\textbf{Output:}}

\begin{document}

\title{Spatial Context Awareness for Unsupervised Change Detection in Optical Satellite Images}

\author{Lukas~Kondmann \orcidicon{0000-0002-2253-6936},
        Aysim Toker, Sudipan Saha \orcidicon{0000-0002-9440-0720}, Bernhard Schölkopf \orcidicon{0000-0002-8177-0925}, Laura Leal-Taixé \orcidicon{0000-0001-8709-1133},
        and Xiao~Xiang~Zhu \orcidicon{0000-0001-5530-3613}%
\thanks{The work is jointly supported by the Helmholtz Association through the joint research school “Munich School for Data Science - MUDS", the Framework of Helmholtz AI (grant  number:  ZT-I-PF-5-01) - Local Unit ``Munich Unit @Aeronautics, Space and Transport (MASTr)'' and Helmholtz Excellent Professorship ``Data Science in Earth Observation - Big Data Fusion for Urban Research''(grant number: W2-W3-100), by the European Research Council (ERC) under the European Union's Horizon 2020 research and innovation programme (grant agreement No. [ERC-2016-StG-714087], Acronym: \textit{So2Sat}),  and by the German Federal Ministry of Education and Research (BMBF) in the framework of the international future AI lab "AI4EO -- Artificial Intelligence for Earth Observation: Reasoning, Uncertainties, Ethics and Beyond" (grant number: 01DD20001).
}
\thanks{\emph{Corresponding author: Xiao Xiang Zhu.}}
\thanks{Lukas Kondmann and Xiao Xiang Zhu are with the Remote Sensing Technology Institute, German Aerospace Center (DLR), Wessling, Germany, and also with Data Science in Earth Observation, Technical University of Munich, Munich, Germany (e-mail: \{lukas.kondmann,xiaoxiang.zhu\}@dlr.de).}
\thanks{Sudipan Saha is with Data Science in Earth Observation, Technical University of Munich (TUM), Munich, Germany (e-mail: sudipan.saha@tum.de)}
\thanks{Aysim Toker and Laura Leal-Taixé are with the Dynamic Vision and Learning Group, Technical University of Munich, Munich Germany. (e-mail: \{leal.taixe,aysim.toker\}@tum.de)}
\thanks{Bernhard Schölkopf is with the Max Planck Institute for Intelligent Systems, Tübingen, Germany.}%
}

 \markboth{Submitted to IEEE Transactions on Geoscience and Remote Sensing}%
 {Shell \MakeLowercase{\textit{et al.}}: Bare Demo of IEEEtran.cls for IEEE Journals}

\maketitle

\begin{abstract}
Detecting changes on the ground in multitemporal Earth observation data is one of the key problems in remote sensing. In this paper, we introduce Sibling Regression for Optical Change detection (SiROC), an unsupervised method for change detection in optical satellite images with medium and high resolution. SiROC is a spatial context-based method that models a pixel as a linear combination of its distant neighbors. It uses this model to analyze differences in the pixel and its spatial context-based predictions in subsequent time periods for change detection. We combine this spatial context-based change detection with ensembling over mutually exclusive neighborhoods and transitioning from pixel to object-level changes with morphological operations. SiROC achieves competitive performance for change detection with medium-resolution Sentinel-2 and high-resolution Planetscope imagery on four datasets. Besides accurate predictions without the need for training, SiROC also provides a well-calibrated uncertainty of its predictions. This makes the method especially useful in conjunction with deep-learning based methods for applications such as pseudo-labeling.

\end{abstract}


\IEEEpeerreviewmaketitle

\section{Introduction} \label{sec:1}

\IEEEPARstart{C}{hange Detection} (CD) is at the heart of many impactful applications of remote sensing. Studying differences in land cover and land use over time with remote sensing imagery can shed light on urbanization trends \cite{lu2011detection,ji2019building}, ecosystem dynamics \cite{chen2011airborne}, surface water and sea ice trends \cite{gao2019transferred,ROKNI2015226} and damages through natural disasters \cite{gupta2019creating,moya2020detecting}. Because of rising spatial and temporal resolution of Earth observation imagery, the possibilities of multi-temporal analysis have increased significantly \cite{zhu2017deep}. Combined with the open data policy of the Copernicus program, it is, for example, possible to acquire a Sentinel-2 image with 10m resolution per pixel of any region of interest on any continent every 5 days \cite{drusch2012sentinel} free of charge. Commercial providers of satellite imagery can even offer almost daily coverage with high-resolution imagery for large parts of the planet \cite{kwan2018assessment}. These trends emphasize the increasing opportunities in monitoring Earth from space and the relevance of CD as a field within remote sensing. Obtaining labeled data for CD, however, is costly in terms of time and effort, especially at scale. Therefore, a large focus of attention in the design of CD algorithms is unsupervised methods that do not require ground truth \cite{saha2019unsupervised}. 

The applicability of unsupervised CD methods in multispectral satellite images varies depending on the spatial resolution of input images. For very-high-resolution (VHR) imagery with a spatial resolution up to 0.5m, deep learning-based methods tend to be in general preferable because of their elaborate capacity to model spatial context \cite{saha2019unsupervised} although most of the work in this area focuses on supervised methods \cite{zhan2017change,lyu2016learning,mou2018learning,zhang2016change,saha2020semisupervised,gong2019generative}. Since for VHR imagery an object such as a building consists of a number of pixels, modeling spatial context is essential to provide accurate unsupervised change segmentations. Saha et al. (2019) introduce Deep Change Vector Analysis (DCVA) \cite{saha2019unsupervised}, a  VHR CD framework that combines ideas from image differencing with feature extraction based on pre-trained neural networks. DCVA has also been combined with self-supervised pre-training of the feature extractor specifically for remote sensing images \cite{saha2020deepjoint}. MSDRL \cite{zhan2020unsupervised} is a scale-driven unsupervised method that uses deep feature extraction to obtain a pseudo-classification of change superpixels. Superpixels with high certainty pseudo-labels are then taken as input to train a support vector machine which eventually classifies the uncertain superpixels. 
Such pre-classification schemes where pseudo-labels are obtained based on another method have also been presented in conjunction with methods for unsupervised change detection in Synthetic Aperture Radar images \cite{gong2017feature,gao2019sea,gao2019transferred} such as PCANet \cite{gao2016automatic,li2018sar}. 
Gong et al. \cite{gong2017generative} introduced modeling the difference image with a generative adversarial network (GAN). While the deep learning methods above were primarily designed for high-resolution imagery, some of them can be applied to medium-resolution imagery as well. In the case of DCVA, there also exists a variant adjusted to the spatial and spectral scale of Sentinel-2\cite{saha2020unsupervised}. 

For medium-resolution CD, non-deep learning methods based and improved on 
Change Vector Analysis (CVA) can still compete. CVA takes the difference of radiometric values or features derived from it over time \cite{saha2019unsupervised} and applies a threshold to this difference image. Examples of features that have been derived from radiometric values as input for image differencing are vegetation indices \cite{singh1989review} or tasseled cap transformation features \cite{correa2014change}. Otsu thresholding \cite{otsu1979threshold} has been shown to be effective for thresholding the difference image \cite{thonfeld2016robust} although a variety of approaches exist \cite{bruzzone2000automatic,bruzzone2000minimum,celik2009unsupervised}. Beyond, binary change detection,  the signal in the CVA difference image can also be used to uncover the type of change \cite{bovolo2006theoretical,bovolo2011framework}. 

CVA-based methods can still be insightful especially with medium-resolution because the size of objects in these images is typically assumed to be similar to the spatial resolution of a pixel. However, extensions of CVA still fall short of the deep learning-based DCVA for unsupervised CD on the OSCD benchmark. Still, the relative performance of traditional methods based on CVA improves with medium-resolution imagery compared to higher resolutions. More recent versions of CVA that can also be applied to higher resolution imagery tend to include close spatial context of pixels to some extent. Parcel CVA (PCVA) includes surrounding information of pixels by independent hierarchical segmentation at several scales \cite{bovolo2008multilevel}. Robust CVA (RCVA) improves on potential co-registration errors in the CVA framework by replacing a point in the difference image with the difference to a neighboring pixel if the difference to this neighbor is smaller \cite{thonfeld2016robust}. Object CVA (OCVA) computes histograms of object sizes in an image and incorporates this information into a CVA framework \cite{li2016change}. Image differencing methods have also been successfully combined with morphological operations which allow transitioning from the pixel to the object level \cite{falco2016unsupervised}. 

Although neighboring pixels are somewhat included in the change analysis of a pixel in these extensions of CVA, the spatial extent of incorporated information is small compared to the effective window of sequential convolutional operations in neural networks. Neighborhood in this context is defined not only as the immediate neighbors to a pixel of interest, but also its larger spatial context up to a distance measure. The distant neighborhood of a pixel may help to identify changes because it is also affected by local trends in the image but unaffected by the change itself. For example, consider an explosion of a building between pre and post-image such as in the Beirut dataset used below. Analyzing the distant neighborhood allows to separate the actual change (destroyed building) from local trends such as dust and dirt remains on surrounding buildings stirred by the explosion. However, the use of distant neighborhood context has only found limited application in change detection thus far.
This is particularly surprising since applications of image differencing in other domains such as astronomy emphasize the importance of the relation of a pixel to its neighborhood \cite{wang2016causal}. 

Wang et al. (2016) present the causal pixel model (CPM) for the study of multi-temporal Kepler data which is used to spot transiting exoplanets in front of distant stars observed by the space telescope. The method is also more generally known as half-sibling regression (HSR) \cite{scholkopf2016modeling} Their task is conceptually related to a CD problem in remote sensing since it is also centered around spotting changes in multi-temporal reflection intensities which should be unrelated to the acquisition conditions. In their case, these deviations hint towards a transient object in front of a distant star rather than a change on the ground but the fundamental principle is similar. They solve this task by modeling pixels as a function of their distant neighbors. With this model, it is possible to obtain a prediction for pixels in subsequent time steps based on their distant neighbors and compare the prediction to the actual value of the pixel. The size of the difference between predictions of pixels and their actual values is interpreted as the strength of the change signal.

HSR is related to the application of local binary patterns \cite{heikkila2009description} for change detection in more traditional image recognition problems. Bilodeau et al. (2013) \cite{bilodeau2013change} design a method based on local binary similarity patterns (LBSP) to separate the image background from changes in multitemporal images. In their method, a binary similarity measure is computed between a pixel of interest and its closest neighbors within an image. If the binary similarity pattern updates notably between images, this is considered to be a change signal. A version of LBSPs has also been applied to change detection in remote sensing where multi-temporal images are split into overlapping blocks and LBSPs of these blocks are compared across time \cite{gupta2018change}. Similarly, the graph structure of image patches across time has been used for homogenous and heterogenous change detection \cite{sun2020patch}. The shared principle between LBSP and HSR is the approach to compare a pixel to its neighborhood and inspect how this relationship changes over time to discover potential changes. However, HSR relies mostly on distant neighborhood information rather than a small set of close neighbors and models this relationship explicitly to obtain a prediction for subsequent time periods. 

One key property of HSR is the fact that it is by design comparably robust to registration errors and varying acquisition conditions for a given sensor \cite{wang2016causal}. This is because variations in the acquisition conditions can also affect distant neighbors in an image whereas actual changes at the pixel level should be independent of distant context. Changing acquisition conditions and registration errors, however, are two of the primary sources of false positives in change detection \cite{bruzzone2010conceptual}. Since HSR deals comparably well with these issues it may work especially well for change detection in remote sensing time-series data. In this paper, we therefore apply HSR for change detection in remote sensing. When we know from astronomy that distant spatial context can improve resilience against varying acquisition conditions, this might be particularly helpful in remote sensing CD. 

We modify HSR in two major ways to apply it as SiROC for change detection in remote sensing: At first, we design an ensemble version of HSR based on mutually exclusive neighborhoods. Second, we make use of morphological operations to transition from pixel-level changes to object-level changes. SiROC is tested for urban CD in medium-resolution images on the Onera Satellite Change Detection Dataset (OSCD) and high-resolution images from the Beirut Harbor Explosion of 2020 (BHED). Outside of the urban context, we test SiROC on the Barrax Agriculture dataset and the Lamar Alpine dataset. Our main contributions are three-fold:

\begin{enumerate}
    \item We introduce SiROC, a robust method for unsupervised change detection in optical remote sensing that combines ideas from HSR with ensembles over mutually exclusive neighborhoods and morphological operations. 
    \item SiROC achieves competitive performance for medium- as well as high-resolution unsupervised change detection with optical images.
    \item SiROC also returns a built-in, well-calibrated uncertainty score with its change segmentation which makes it well suited for applications in conjunction with deep learning such as pseudo-labeling in self- or unsupervised learning.
\end{enumerate}

\begin{figure*}[!htbp]
\centering
\subfloat[]{\includegraphics[width=2 in]{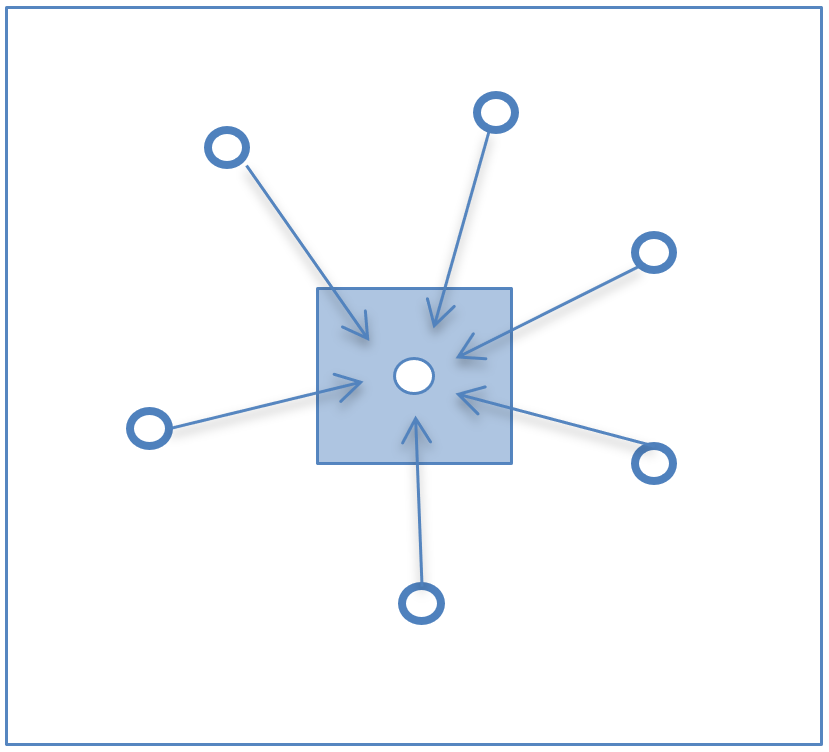}%
\label{HSR_fit}}
\hfil
\subfloat[]{\includegraphics[width=2 in]{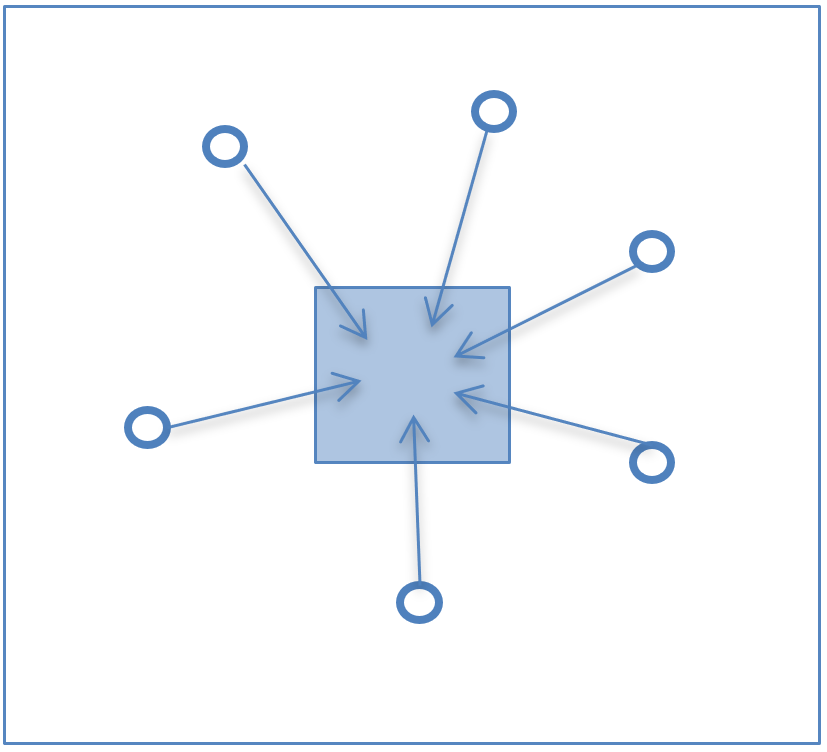}%
\label{HSR_predict}}
\hfil
\subfloat[]{\includegraphics[width=1.97 in]{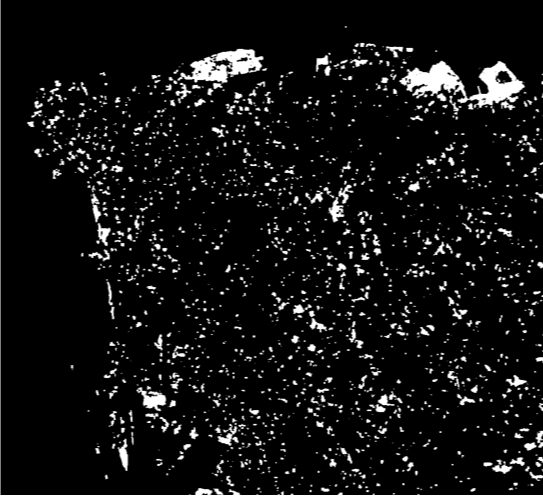}%
\label{HSR_difference}}
\caption{Half-Sibling Regression (HSR): Figure \ref{HSR_fit}: A set of neighbors is fit to a pixel of interest as a linear combination at time $t$). Figure \ref{HSR_predict}: At $t+1$, the pixel values of the neighbors are used together with the coefficients obtained at time $t$ to predict the pixel of interest in $t+1$. Figure \ref{HSR_difference}: The predicted pixel values are compared with the actual pixel values at $t+1$ to obtain a change signal.}
\label{fig:HSR}
\end{figure*}

\section{Method}
\subsection{Half-Sibling Regression (HSR) Image Differencing}
The foundation of our method is Half-Sibling Regression which was originally developed for time-series analysis of Kepler data in astronomy \cite{wang2016causal,wang2017pixel,Gebhard_2020,scholkopf2016modeling}. Figure \ref{fig:HSR} displays the intuition of HSR and how it is applied to obtain signals of changing pixels across time in three steps. 

At first, HSR models the pixel value of a star at time $t$ as a linear combination of the pixel values of many other stars from the distant neighborhood of the pixel (Figure \ref{HSR_fit}). The result of this first step is a linear coefficient for every included neighborhood pixel. In the second step (Figure \ref{HSR_predict}), predictions for the pixel at time $t+1$ are obtained with HSR based on the neighboring pixels at $t+1$ and the respective linear coefficients from step 1. If steps 1 and 2 are executed for the whole image, there is a prediction for any pixel at $t+1$ as well as its actual value. The predicted image at $t+1$ is then subtracted from the actual image value to obtain a change signal for every pixel (Figure \ref{HSR_difference}). Intuitively, one expects a change in the pixels where the predictions based on the pixel's relation to its neighborhood in the past divert from the actual realization of the pixel. In the following, we elaborate on the formal definition of the three steps described. We restrict our description of HSR to the case of two time periods for simplicity since this is how we apply the method to remote sensing as well.

\textit{Step 1} Let $I_{x,y,t}$ be a pixel in a single-channel, two-dimensional image time-series ($I$) at time $t$ with coordinates $(x,y)$. HSR models the pixel $I_{x,y,t}$ as a linear combination of a set of distant neighbors $\mathrm{N}$ from $I$. The neighborhood set has the points $I_{i,j,t}$ as elements such that $(i,j) \in \mathrm{N_{x,y}}$: 

\begin{equation}
    \label{eq: Linear Eq.}
    I_{x,y,t} = \sum_{(i,j)\in \mathrm{N_{x,y}}} \beta_{i,j,x,y} I_{i,j,t} + \epsilon_{x,y,t}
\end{equation}

Neighbors are chosen from the distant neighborhood because they might be subject to the same noise as the pixel of interest when they are selected to close to it. Wang et al. (2016) \cite{wang2016causal} require that an eligible neighbor has a distance of at least 20 pixels from the pixel of interest to be considered. This ensures that the pixel of interest and the chosen neighbors have practically no overlap in stellar illumination. The number of neighboring pixels considered is generally large and Wang et al. (2016) \cite{wang2016causal} select 4000 neighboring pixels in their original proposal of HSR to model one pixel of interest for Kepler data. Given the high temporal density of observations for each pixel (every 30 minutes) in Wang et al. (2016) \cite{wang2016causal}, this is still solvable because the number of observed time periods exceeds the number of neighboring pixels used. 

However, in the bitemporal case, where only one period is used for fitting, there are many potential combinations of $\beta$ which solve equation (\ref{eq: Linear Eq.}). We derive $\beta$ as the closed form solution of the least squares problem. It is a function of the pixel of interest $I_{x,y}$, the respective neighbor $I_{i,j}$ and the quadratic sum of all neighbors $I_{i',j'}$:

\begin{equation}
    \label{eq: new beta.}
    \beta_{i,j,x,y} =\frac{I_{i,j,t}}{\sum_{(i',j')\in \mathrm{N_{x,y}}}  I_{i',j',t}^{2}}I_{x,y,t}
\end{equation}

\textit{Step 2}: With the coefficients obtained in step 1, $I_{x,y,t+1}$ can be predicted as 

\begin{equation}
    \label{eq: Prediction}
    \hat{I}_{x,y,t+1} = {\sum_{(i,j)\in \mathrm{N_{x,y}}}}\beta_{i,j,x,y} I_{i,j,t+1}
\end{equation}

With the expression for $\beta$ from equation (\ref{eq: new beta.}), equation (\ref{eq: Prediction}), can be rearranged to: 

\begin{equation}
    \label{eq: Prediction Sum}
    \hat{I}_{x,y,t+1} =  \frac{\sum_{(i,j)\in \mathrm{N_{x,y}}} I_{i,j,t+1}I_{i,j,t}}{\sum_{(i,j)\in \mathrm{N_{x,y}}}{I_{i,j,t}^{2}}}I_{x,y,t} \equiv g_{t+1} I_{x,y,t}
\end{equation}

where $g_{t+1}$ resembles a growth rate of the sum of pixel values in the selected neighbors from $t$ to $t+1$. In essence, the assumption is that if the pixel values around $I_{x,y,t}$ increase by a factor $g_{t+1}$ and no changes occurred at this location, $I_{x,y,t+1}$ should be close to $I_{x,y,t}g_{t+1}$. We can circumvent the explicit calculation of beta and directly obtain $\hat{I}_{x,y,t+1}$ based on equation (\ref{eq: Prediction Sum}) which is computationally efficient. 

\textit{Step 3}: The difference between $I_{x,y,t+1}$ and $\hat{I}_{x,y,t+1}$ is taken as the change signal for pixel $I_{x,y}$ between $t$ and $t+1$. 

\begin{equation}
    \label{eq: Residual.}
    I_{x,y,t+1} = \hat{I}_{x,y,t+1} + \epsilon_{x,y,t+1}
\end{equation}

After obtaining  $I_{x,y,t+1}$ for all $(x,y) \in I_{t+1}$, the residual is given as the difference of the image matrices: 
\begin{equation}
    \label{eq: Residual.}
    \mathbf{\epsilon_{t+1}} = \mathbf{\hat{I}_{t+1}} - \mathbf{I_{t+1}}
\end{equation} 

Note that this is slightly different from the standard application of image differencing in change vector analysis in multi-temporal remote sensing. We do not directly take the difference of the image vectors at $t$ and $t+1$. Instead, we predict how the image would have looked like in $t+1$ if the local neighborhood relations persisted. Then, we use this predicted image as input for image differencing with the actual image in $t+1$. The extension of HSR to images with several channels is straightforward as one can directly sum the absolute values of $\epsilon$ for each channel to incorporate HSR information from all channels. Let $\epsilon_{t+1,c}$ be the residual of channel c of a multispectral image with $C$ channels. Then, the aggregated chance signal can be computed as:

\begin{equation}
    \label{eq: Aggregation}
    \mathbf{\epsilon_{t+1}} := 
    \sum_{c=1}^{C}\mathbf{|\epsilon_{t+1,c}|}
\end{equation} 

 \begin{algorithm}
 \caption{: SiROC}
 \begin{algorithmic}[1]
 \renewcommand{\algorithmicrequire}{\textbf{Input:}}
 \renewcommand{\algorithmicensure}{\textbf{Output:}}

 \REQUIRE $I_t$, $I_{t+1}$, s, n\_max,e\_start  
 \ENSURE  Binary Change Segmentation\\

  \STATE e = e\_start, n = e\_start+s
  \STATE Uncertainty\_CM = zeros\_like($I_t$)
  \WHILE {n $<$  n\_max}
  \FOR {(channel in channels)}
  \FOR {(pixel in $I_{t}$)}
  \STATE Apply HSR(n,e) to get $\mathbf{\hat{I}_{t+1}}$
  \ENDFOR
  \STATE Channel\_Difference\_Image = $\mathbf{\hat{I}_{t+1}}$ - $\mathbf{I_{t+1}}$
  \ENDFOR
  \STATE Diff\_Image = Sum($|$Channel\_Difference\_Images$|$) 
  \STATE Binary\_CM = Otsu\_Thresholding(Diff\_Image)
  \STATE Binary\_CM\_Object = Morph\_Profile(Binary\_CM)
  \STATE Uncertainty\_CM = Uncertainty\_CM + Binary\_CM\_Object
  \STATE n = n + s
  \STATE e = e + s 
  \ENDWHILE
  \STATE Final\_Segmentation = Threshold(Uncertainty\_CM)
 \end{algorithmic} 
 
 \end{algorithm}

\subsection{HSR for Earth Observation Data (SiROC)}
We improve and adapt the standard HSR Image Differencing model to apply it effectively for CD in remote sensing as Sibling Regression for Optical Change detection (SiROC). Algorithm \ref{SiROC} outlines SiROC in pseudocode. In summary, there are two major differences between SiROC and HSR Image Differencing. At first, we redesign the notion of included and excluded pixels in the neighborhood selection to create an ensemble version of HSR over mutually exclusive neighborhoods. This does not only improve performance but also allows us to obtain an uncertainty along with the prediction. The rationale of splitting neighborhoods by distance is to inspect trends at different distances separately instead of pooling the trends together. For example, two trends at different spatial scales might offset each other when pooling them although both may be a signal for change. Second, we combine HSR with morphological profiles to move from pixel-level to object-level changes since changes in remote sensing typically occur at the object level.

\textit{Ensembling:} The starting point for SiROC is applying HSR to  $\mathbf{I_t}$ to obtain  $\mathbf{\hat{I}_{t+1}}$ based on a set of neighboring pixels. We use all pixels that have a distance of at least $e$ but at most $n$ rows or columns from the pixel of interest. Graphically, this corresponds to all points in a square with width $2n$ and $I_{x,y,t}$ in its center that are not in the smaller square with width $2e$ around $I_{x,y,t}$. Formally, a pixel $I_{x',y',t}$ is included in the set of neighbors for $I_{x,y,t}$ if 
\begin{equation}
e < max\{|x'-x|,|y'-y|\} \leq  n
\end{equation}

With $\mathbf{\hat{I}_{t+1}}$, a channel-level difference image is obtained by taking the difference $\mathbf{\hat{I}_{t+1}}$ - $\mathbf{I_{t+1}}$. The absolute value of the change signal is summed across the channels. We apply Otsu-thresholding \cite{otsu1979threshold} to the resulting difference image which has been successfully used for thresholding difference images in CD before \cite{thonfeld2016robust}. Further, the evaluation of competing methods is also based on this thresholding approach. This allows for comparing relevant methods in a fixed setting. Nevertheless, Otsu-thresholding is a design choice here with a variety of alternatives that can also be used in conjunction with SiROC including the T-point method \cite{coudray2010robust}, the Rosin method \cite{rosin2001unimodal} or the expectation-maximization algorithm \cite{bazi2007image,zanetti2015rayleigh}.

The result of the thresholding step is a binary segmentation of the difference image on the pixel level. However, in remote sensing applications, changes such as the construction of roads or buildings tend to occur at the object level. This is why object-based methods often tend to be superior for these applications \cite{song2020uncertainty}. We rely on morphological profiles which are an established tool to bridge the gap between pixel-level change segmentations and the object level \cite{dalla2008unsupervised}.

\textit{Morphological Profile:}
A morphological profile is the sequential application of morphological opening and closing to an image \cite{dalla2008unsupervised}. We employ morphological opening and closing at one spatial filter size $p$. Intuitively, morphological closing helps to fill in missed pixels in detected change objects as changed. On the other hand, morphological opening removes spurious false positives when there are no other changes around them. After obtaining an object-level change segmentation for a given neighborhood size $n$ and exclusion window $e$, we repeat the procedure and use new neighbors that are further away than the current set. Both, $n$ and $e$ increase by the same added factor $s$. In the next iteration, the previous neighborhood window becomes the exclusion window and a new binary segmentation based on more distant neighbors is obtained. This procedure is repeated until a maximum neighborhood size is reached. The number of models F is given as $F=((n\_max - n\_min) // s)+1$. Every model in the ensemble classifies each pixel either as change (1) or no-change (0) which can be interpreted as a voting mechanism among models. Voting mechanisms across spatial scales \cite{liu2017multiscale} or different bands \cite{saha2020unsupervised} are a common aggregation mechanism in change detection. The number of votes per pixel ranges between 0 and F. 

\textit{Majority Voting:}
The matrix of votes per pixel can be visualized as a heatmap of agreement between different sets of neighbors if a change occurred. This also directly transports a measure of uncertainty embedded in SiROC. If a pixel has no or the maximum number of votes, the agreement is high and the method is confident in its prediction. If the number of votes is split the model shows low confidence in its prediction for this point.
We threshold these votes with a pre-defined voting share $0 \leq v \leq 1$ that is required to classify a pixel as changed. $v$ is the sensitivity of our model towards change. The choice of $v$ contains a trade-off between objectives. With a higher $v$, the number of false negatives rises but false positives decline (and vice versa).  
Since all models are equally weighted in the voting process, the importance of a single neighbor is decreasing in its distance to the pixel of interest. This is because the number of neighbors used per model is increasing in n. The underlying assumption is that pixels closer to the point of interest carry more information about its potential change. This assumption is domain-specific to Earth observation and stands in contrast to the idea of HSR in astronomy where there is no weighting based on distance. The application of the voting threshold is the last step of SiROC to obtain the final change segmentation. The voting matrix is normalized by the number of models before the percentage threshold is applied.

To summarize, SiROC has the following hyperparameters:
\begin{enumerate}
    \item Maximum neighborhood size: n\_max 
    \item Initial exclusion window: e\_start 
    \item Step size of ensemble: s 
    \item Filter Size of Morphological Operations: p 
    \item Voting Threshold: $0 \leq v \leq 1$ 
\end{enumerate}

The initial size of the neighborhood window n\_start is given as e\_start + s. 

\section{Experiments and Results}\label{sec:Results}
Section \ref{subsec:datasets} describes the datasets used to assess the performance of SiROC and competing methods. The competing methods used as a benchmark and the evaluation criteria are described in more detail in Section \ref{subsec:methods}. The results on OSCD, BHED, the Agriculture Dataset and the  Alpine Dataset are presented in depth in Sections \ref{subsec:oscd}, \ref{subsec:beirut} \ref{subsec:agriculture}, and \ref{subsec:alpine} respectively.
 
\subsection{Description of Datasets}\label{subsec:datasets}
\textit{Onera Change Detection Dataset (OSCD)}: OSCD is a benchmark for bitemporal urban CD based on multispectral Sentinel-2 images  \cite{daudt2018urban}. It contains manual annotations of binary changes for 24 cities across the globe where 14 are used for training and 10 for testing. The labels focus on urban changes such as newly constructed buildings and natural changes such as sea-level rise or differences in vegetation are not annotated. The two images per city are selected to be cloud-free and are generally taken about 1-3 years apart. While there are 13 bands available in Sentinel-2 images, we restrict our focus to the RGB channels here. Although SiROC is able to handle channels outside of the visible spectrum as well our experiments show that the inclusion of the NIR band does not add value in the urban applications considered here. This may be different in vegetation monitoring where NIR bands tend to be more insightful. Spatial bands beyond RGB and NIR do not have a spatial resolution of 10m and are therefore excluded as well.

\textit{Beirut Harbor Explosion Dataset (BHED)}:
On 4\textsuperscript{th} August 2020, a devastating explosion of large amounts of ammonium nitrate occurred in the port of Beirut in Lebanon. It led to over 200 deaths and left more than 300,000 people homeless because of heavy damages to buildings in the city.\footnote{https://en.wikipedia.org/wiki/2020\_Beirut\_explosion}  
We collect a pair of cloud-free Planetscope images with 3m per pixel resolution on 1\textsuperscript{st} and 5\textsuperscript{st} August before and after the explosion. We combine these images with ground truth on destroyed buildings provided by the Center for Satellite Based Crisis Information (ZKI) of the German Aerospace Center.\footnote{https://activations.zki.dlr.de/en/activations/items/ACT148.html} The building destruction map is based on manual annotation of very-high resolution images and field reports on the ground. Note that the annotations contain building destruction rather than building damage. Therefore, partial damages to buildings that withstood the explosion are not included. With this dataset, we aim to test the applicability of SiROC not only in medium but also in higher resolution images in problems where fast and accurate annotations are essential.

\textit{Agriculture Dataset}:
To test SiROC also outside the urban domain, we include two other test datasets from Saha et al. (2019) \cite{saha2019unsupervised} as reference points. The first one, the Agricultural dataset, is a scene with bitemporal Sentinel-2 images from July 2015 over Barrax, Spain with 600 $\times$ 600 pixels in size. Between the two images is a time period of 10 days between which agricultural field activity changed notably. The reference map was manually annotated by the authors of Saha et al. (2019) \cite{saha2019unsupervised}.   

\textit{Alpine Dataset}:
The second dataset consists of pre and post Sentinel-2 images of a fire in an alpine region close to Trento, Italy in spring 2019. A variety of other seasonal vegetation trends like ice and snow complicate this dataset. The scene has a size of 350 $\times$ 350 pixels with ground truth annotated manually by the authors of Saha et al. (2019) \cite{saha2019unsupervised}. 

\subsection{Competing Methods and Criteria}\label{subsec:methods}
We compare our results to a variety of state-of-the-art unsupervised methods for optical CD in remote sensing. Since SiROC needs no training and does not rely on pre-trained neural networks its primary group of comparison consists of other image differencing-based methods. This makes SiROC fast compared to deep learning methods with comparable speed to traditional methods. We include several frameworks that improve on classical CVA. RCVA \cite{thonfeld2016robust} incorporates close neighborhood information to make CVA more robust against misregistration. PCVA \cite{bovolo2008multilevel} uses change vector analysis of multi-level parcels to improve on CVA. DCVA is based on deep feature extraction with a deep neural network pretrained on imagenet \cite{saha2019unsupervised}. While DCVA was originally developed for high-resolution images, it is a resolution-agnostic framework relying on deep feature extraction from RGB channels. We also include a version of this method which we call DCVAMR specifically adjusted for medium-resolution, multispectral Sentinel-2 imagery \cite{saha2020unsupervised} for the OSCD dataset. For BHED, we include DCVA, RCVA and PCVA as baselines as DCVAMR is not capable of handling Planetscope input channels. The most recent advancement in unsupervised CD for high-resolution imagery is Saha et al. (2020) \cite{saha2020deepjoint} who employ self-supervised pre-training on remote sensing images in combination with a DCVA framework. We call this refined version of DCVA `SSDCVA' and include it as the primary baseline besides general DCVA for BHED.  

In line with previous evaluations on OSCD \cite{daudt2018urban,saha2020unsupervised}, we analyze the performance of SiROC against the state-of-the-art in binary change segmentation based on specificity and sensitivity. Specificity is defined as the number of true positives (TP) over the sum of TP and false positives (FP): Specificity = $\frac{TN}{TN+FP}$. Sensitivity is the number of TP over the sum of TP and false negatives: Sensitivity = $\frac{TP}{TP+FN}$. This criterion is also known as recall. A method that is sensitive towards changes has a high sensitivity but a low specificity (and vice versa). A superior method should balance these objectives and evaluate better in both criteria. To further elaborate on the balance of change and no change class, we also report Precision = $\frac{TP}{TP+FP}$ and F1-Score = $\frac{2*Precision*Recall}{Precision+Recall}$.

\subsection{Results on OSCD}\label{subsec:oscd}
\textit{Parameters}: We tune the parameters of SiROC on the OSCD training set resulting in the following parameter specifications:

\begin{enumerate}
    \item Maximum neighborhood size: n\_max=200 
    \item Initial exclusion window: e\_start=0 
    \item Step size of ensemble: s=8
    \item Filter Size of Morphological Operations: p=5
\end{enumerate}

The maximum neighborhood size is 200 with stepsize 8. Contrary to the original idea of HSR in astronomy, we do not find it to be optimal to exclude direct neighbors of the pixel of interest from the analysis resulting in an initial exclusion window of zero. While neighboring pixels may be subject to the same kind of object-level change on the ground, they still can contribute important information if their weight is moderate. We find the best results with morphological opening and closing with a filter size of 5. We do not tune the voting threshold because this parameter does not influence the change signal directly but rather how the method balances false positives and false negatives. 

\textit{Quantitative Results}:
Table \ref{tab:Results_OSCD} reports specificity and sensitivity scores of SiROC and competing methods on the OSCD test set. Scores are averaged on the city level. SiROC with a voting threshold of $v=1/2$ achieves a specificity of 88.31\% with a sensitivity of 70.71\% and 24.80 \% precision as well as 36.72\% F1-Score. This is a high score in all four categories by a significant margin.  
The difference to DCVAMR is about 6-13 percentage points (p.p.) depending on the category. DCVA achieves a sensitivity that is slightly below but close to SiROC but lacks behind in specificity, sensitivity and F1-score by more than 10 p.p.
Compared to the best results of methods without deep learning-based feature extraction, SiROC gains about 12 p.p. in specificity, 7 p.p. in sensitivity 12 p.p. in precision and 15 p.p. in F1 on RCVA. 

\begin{table}
\centering
\caption{Quantitative Results OSCD Test Set}
\begin{threeparttable}
\footnotesize
\setlength{\tabcolsep}{\tabcolsep}
\begin{tabular}{lcccc}
\toprule
{} & Specificity & Sensitivity & Precision & F1 \\
\midrule
SiROC  & \textbf{88.31\%} & \textbf{70.71}\% & \textbf{24.80\%} & \textbf{36.72\%} \\
DCVAMR
& 78.38\% & 64.63\% & 14.01\% & 23.03\% \\
DCVA & 76.96\% & 69.02\% & 14.03\% & 23.33\% \\
PCVA  & 75.61\% & 47.00\% & 9.50\%  & 15.81\%\\
RCVA  & 76.96\% & 64.08\%  & 13.16\% & 21.84\% \\

\midrule
\multicolumn{5}{c}{Ablation Scores} \\
\midrule
No MP  & 80.64\% & 69.88\% &  16.44\% & 26.62\% \\
HSR  & 79.45\% & 70.24\% & 15.70\% & 25.66\% \\
\bottomrule
\end{tabular}\textbf{}
\end{threeparttable}
\label{tab:Results_OSCD}
\end{table}
To understand the origin of the performance difference to the previous state-of-the-art in more detail, we provide two ablation scores of SiROC. At first, we remove the morphological operations in SiROC. While morphological profiles help to transition to an object-level change mask, SiROC still exceeds previous unsupervised performance without them. No MP performance improves by 2 p.p. on specificity and 5 p.p. sensitivity vs. DCVAMR and by 4 p.p. specificity and 1 p.p. on sensitivity vs. DCVA. The resulting F1 score is 3-4 p.p. higher than deep learning-based methods and 5-10 p.p. higher than traditional methods here. To evaluate the effectiveness of ensembling, we also provide a score for a vanilla HSR with the same neighborhood size and no exclusion window. The Vanilla HSR performs slightly better but in the range of DCVA and DCVAMR with a specificity of 79.45\% and a sensitivity of 70.24\%. The F1 score is about 1 p.p. lower without ensemble voting. 

Therefore, the majority voting mechanism is an effective tool to extract a more granular signal from the general HSR predictions. Further, the use of wide spatial context pioneered in astronomy is advantageous for change detection in remote sensing as well. In summary of Table \ref{tab:Results_OSCD}, SiROC sets a new state-of-the-art of unsupervised CD in medium-resolution images on OSCD. Even without morphological filters, the method still notably outperforms previous scores which points to a strong signal for change information in the original HSR method and the effectiveness of the majority voting mechanism. Combined with ensembling over different neighborhoods and morphological profiles, this exceeds previous quantitative results on the OSCD dataset.  

\begin{figure*}[!htbp]
\centering
\subfloat[SiROC Confidence]{\includegraphics[width=1.5 in]{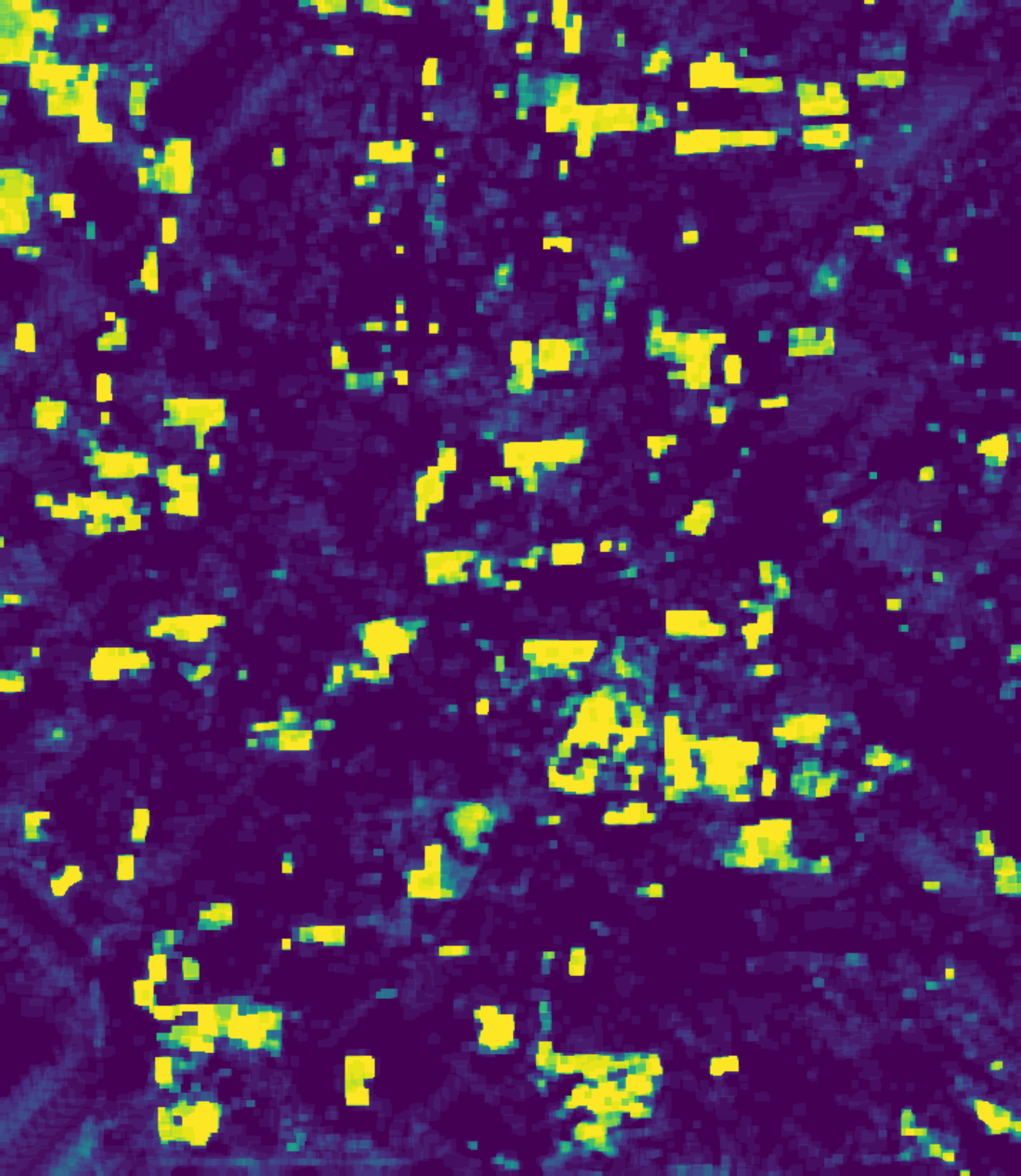}
\label{Vegas Heatmap}}
\hfil
\subfloat[SiROC (Ours)]{\includegraphics[width=1.5 in]{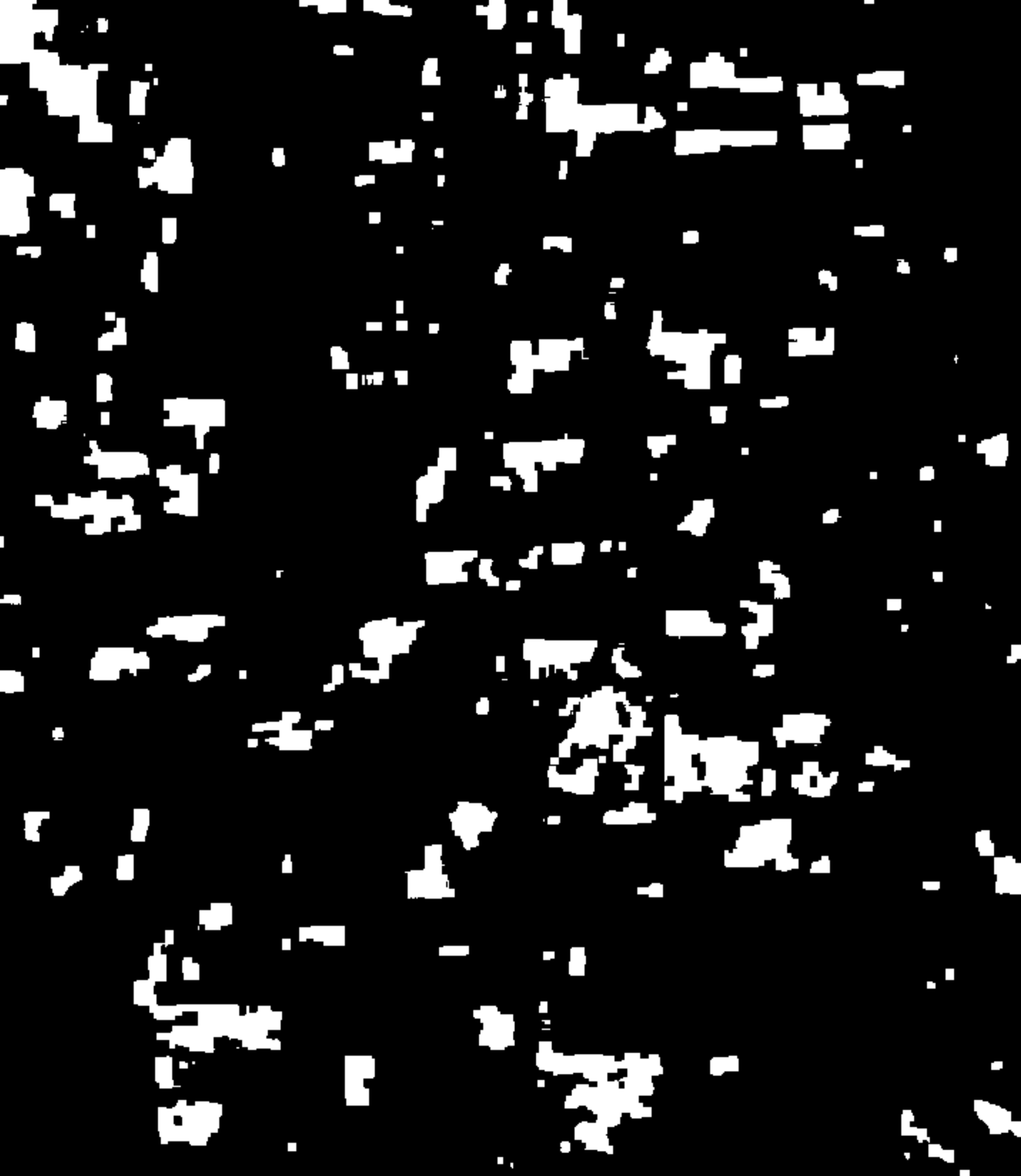}%
\label{Vegas SiROC}}
\hfil
\subfloat[SiROC (No MP)]{\includegraphics[width=1.5 in]{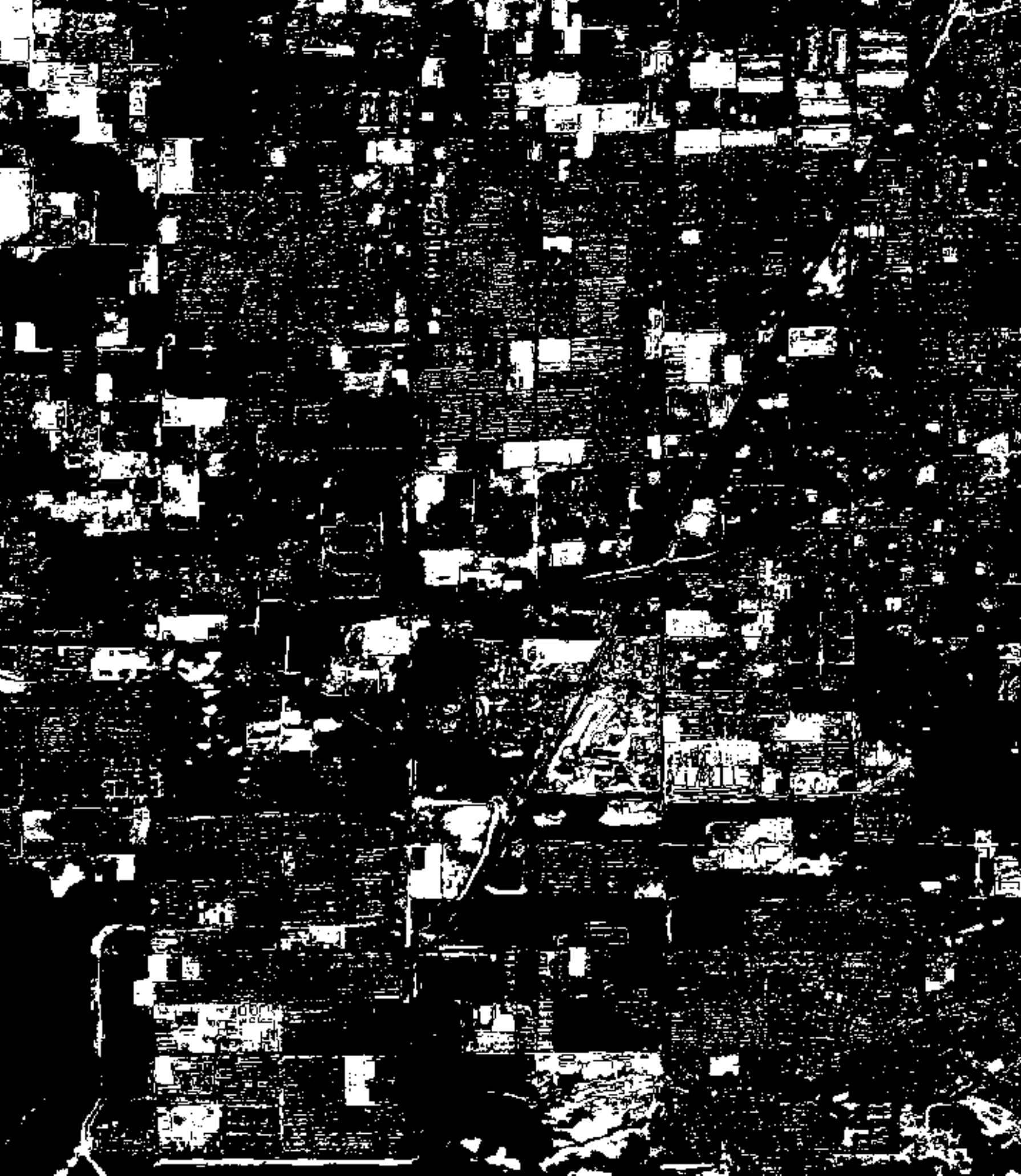}%
\label{Vegas SiROC No MP}}
\hfil
\subfloat[DCVAMR]{\includegraphics[width=1.5 in]{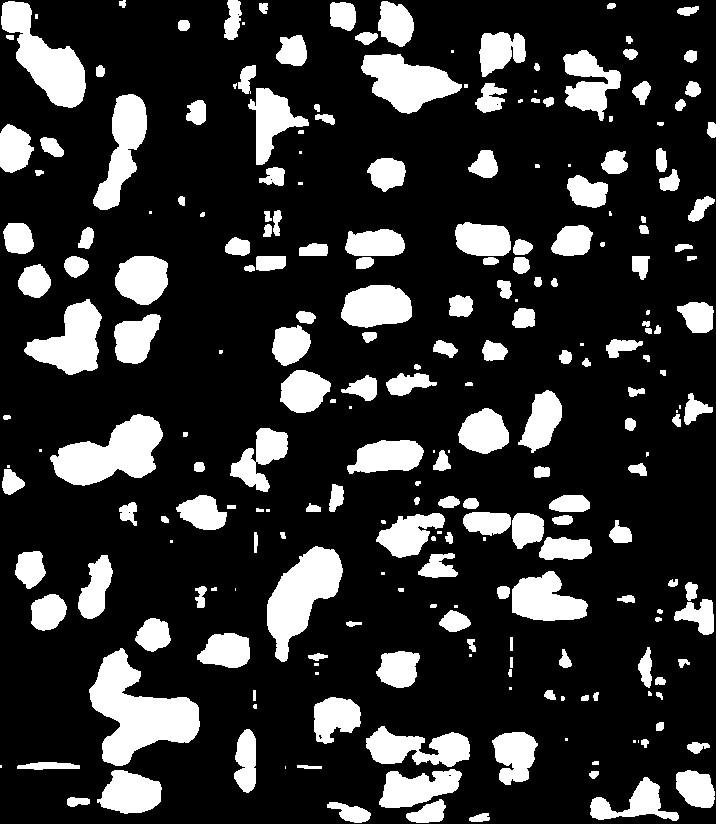}%
\label{Vegas DTL}}
\hfil
\subfloat[DCVA]{\includegraphics[width=1.5 in]{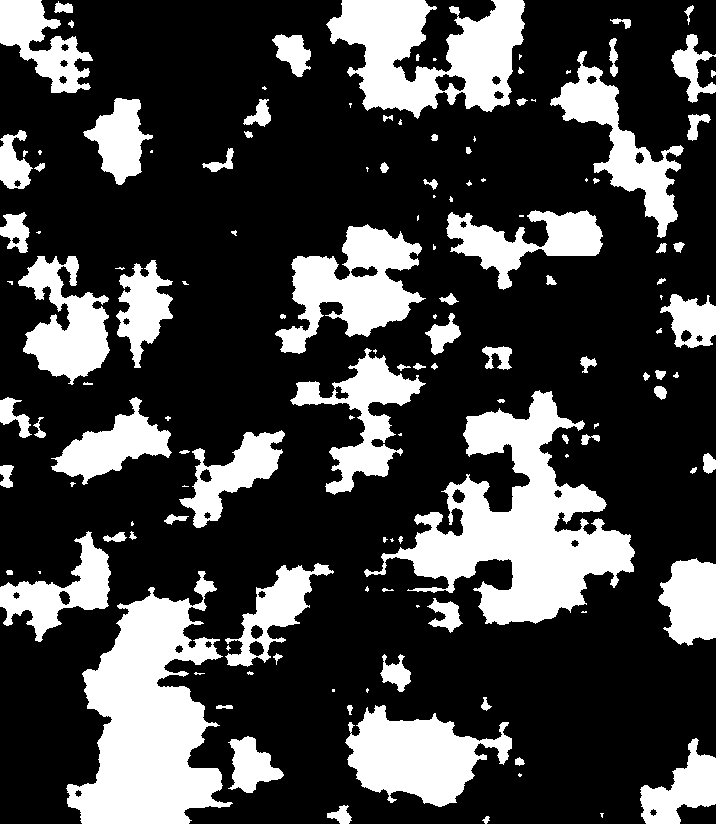}%
\label{Vegas DCVA}}
\hfil
\subfloat[PCVA]{\includegraphics[width=1.5 in]{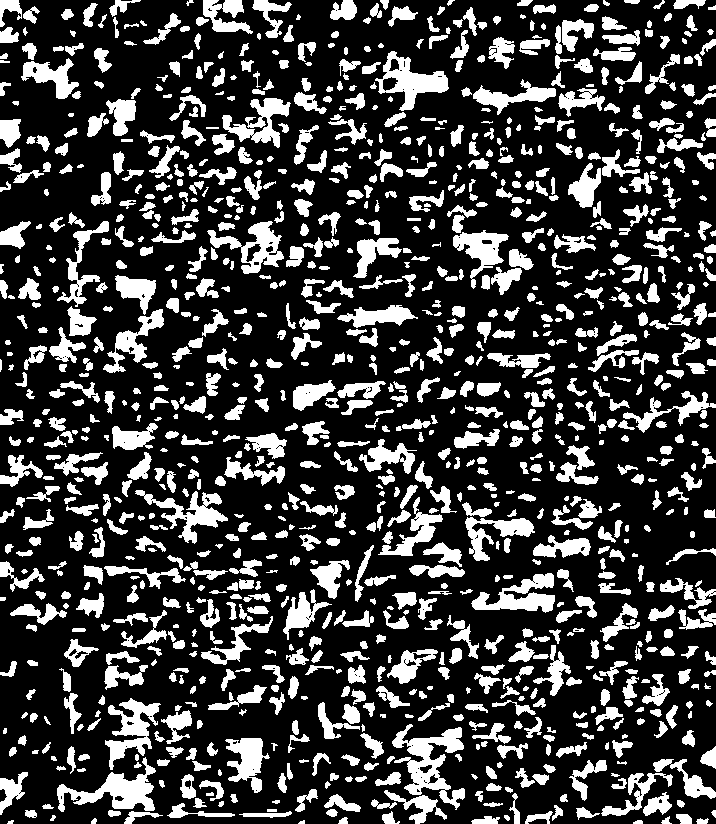}%
\label{Vegas PCVA}}
\hfil
\subfloat[RCVA]{\includegraphics[width=1.5 in]{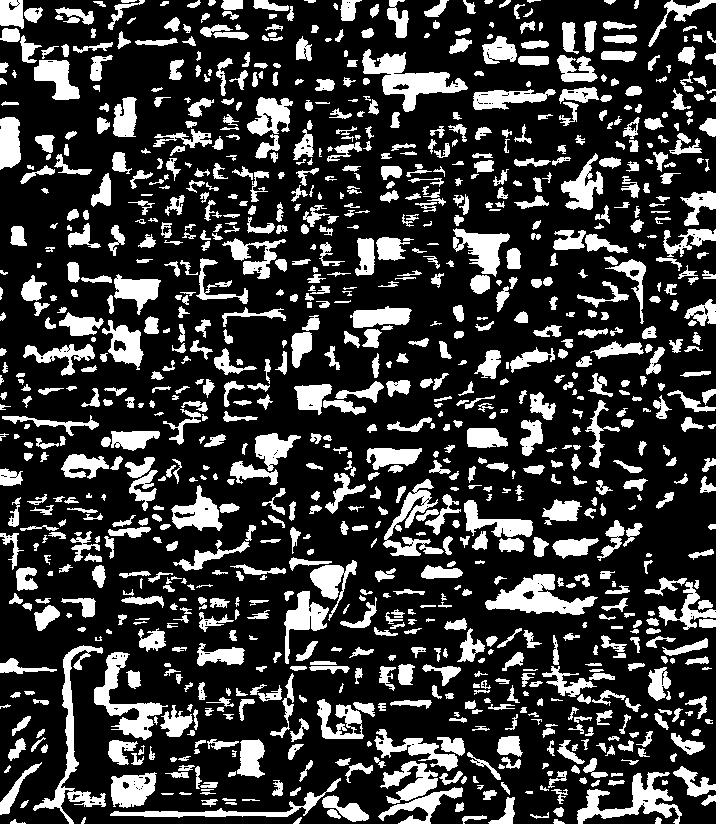}%
\label{Vegas RCVA}}
\hfil
\subfloat[Ground Truth]{\includegraphics[width=1.5 in]{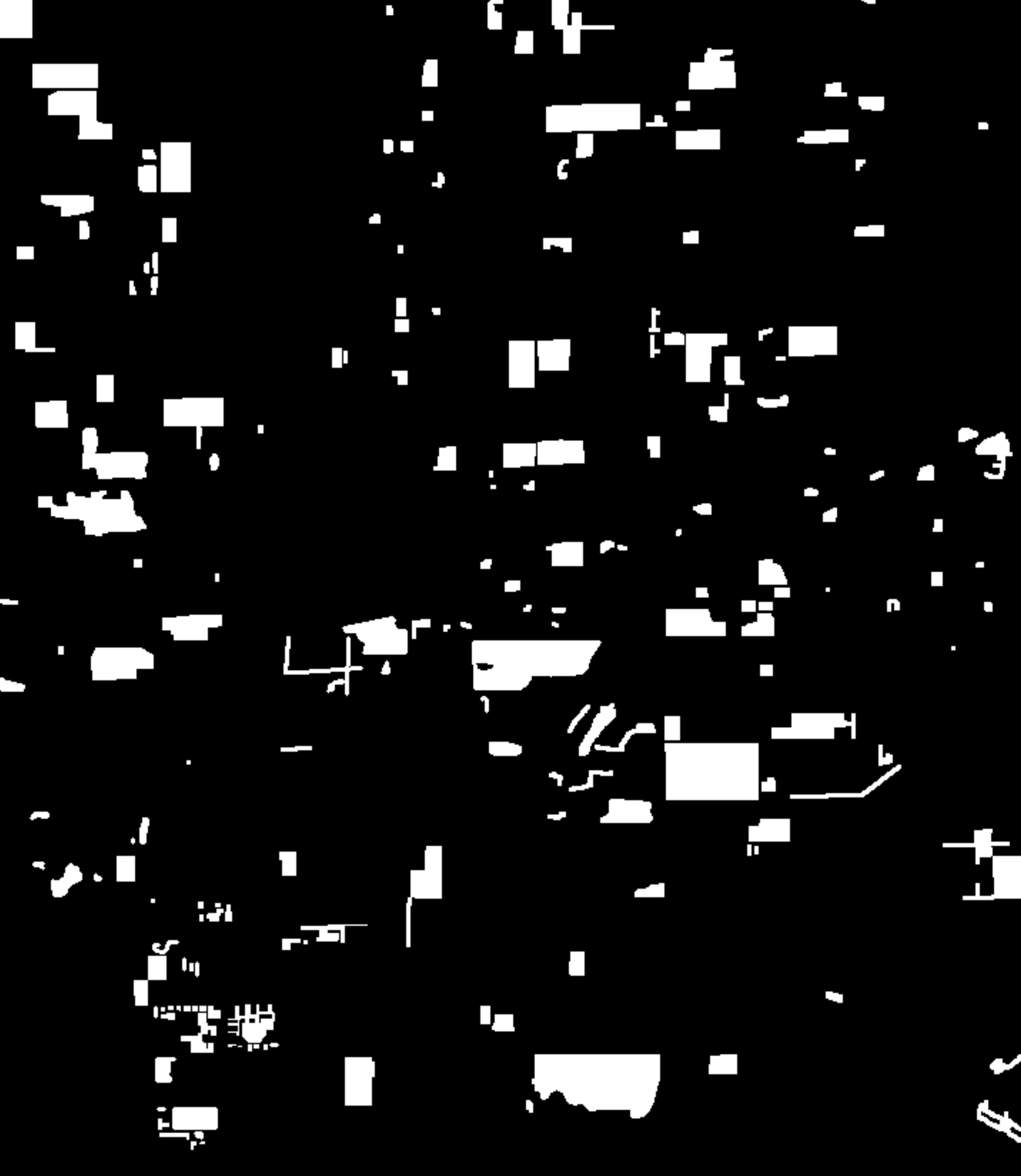}%
\label{Vegas GT}}
\caption{Qualitative Comparison OSCD - Las Vegas. Figure \ref{fig:Las Vegas} visualizes the number of change votes per pixel in SiROC (\ref{Vegas Heatmap}) and the corresponding binary predictions after (\ref{Vegas SiROC}) and before morphological operations (\ref{Vegas SiROC No MP}). Competing models are in \ref{Vegas DTL}-\ref{Vegas RCVA} and the ground truth in \ref{Vegas GT}. SiROC predicts change regions and shapes of the ground truth well while competing methods struggle either with identifying the shapes (\ref{Vegas DTL} \& \ref{Vegas DCVA}) or the areas of change (\ref{Vegas PCVA} \& \ref{Vegas RCVA}) for the Las Vegas Pair.}
\label{fig:Las Vegas}
\end{figure*}

\textit{Qualitative Results}:
The edge of SiROC compared to other unsupervised methods in the quality of change annotations for medium-resolution imagery is also visible when inspecting the predictions for specific scenes. Figure \ref{fig:Las Vegas} displays exemplary change masks for Las Vegas. For SiROC, a threshold of $v=1/2$ was used to obtain the images with a specificity of 95.28\%, sensitivity of 78.75\% and precision of 58.14\% for this scene. Figure \ref{Vegas Heatmap} visualizes the confidence of SiROC in the change propensity of a pixel as a heatmap from dark purple (0\% votes) to yellow (100\% votes). When comparing this to the ground truth on the bottom right, one can see that change confidence is strongly associated with the occurrence of a change. Figure \ref{Vegas SiROC} shows the binary change map after applying the threshold to the uncertainty map. Not only does SiROC pick up on the changed areas in the image, but it also fits the shapes of changing buildings fairly well. The visual similarities between Figure \ref{Vegas SiROC} and \ref{Vegas GT} are striking, especially compared to the other segmentations of competing methods. Also before applying the morphological operations, SiROC identifies the areas of interest in the image well although the predicted mask is naturally slightly more spurious. The morphological operations help to remove these spurious changes but the change signal in the predictions is in line with the ground truth (\ref{Vegas SiROC No MP}). DCVAMR is generally able to discover the changing regions of an image but struggles to identify the shapes of changing objects and rather fits round blobs (Figure \ref{Vegas DTL}). DCVA tends to discover large changes and overestimate their size whereas smaller changes go undetected (Figure \ref{Vegas DCVA}). This might be related to the fact that DCVA was originally designed for high-resolution optical imagery in which building changes are larger in terms of pixel size. This is in line with the fact that DCVAMR, which is explicitly adjusted for Sentinel 2, tends to fit the size of changes better even though it also struggles with change shapes. PCVA and RCVA seem to extract building footprints rather than building changes here which leads to overcrowding of the segmentation mask. 

\begin{figure*}[!htbp]
\centering
\subfloat[SiROC Confidence]{\includegraphics[width=1.5 in]{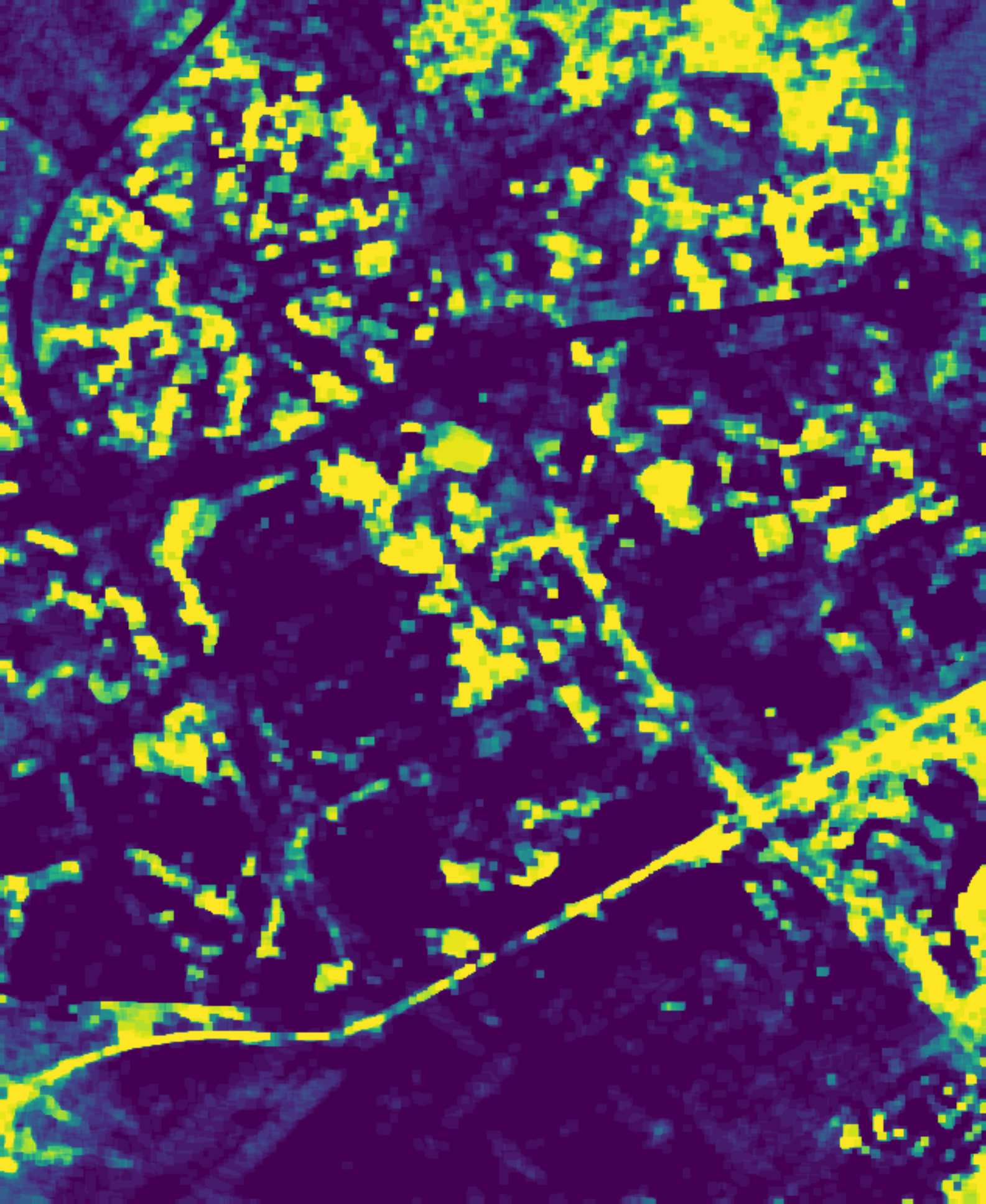}%
\label{Dubai Heatmap}}
\hfil
\subfloat[SiROC (Ours)]{\includegraphics[width=1.5 in]{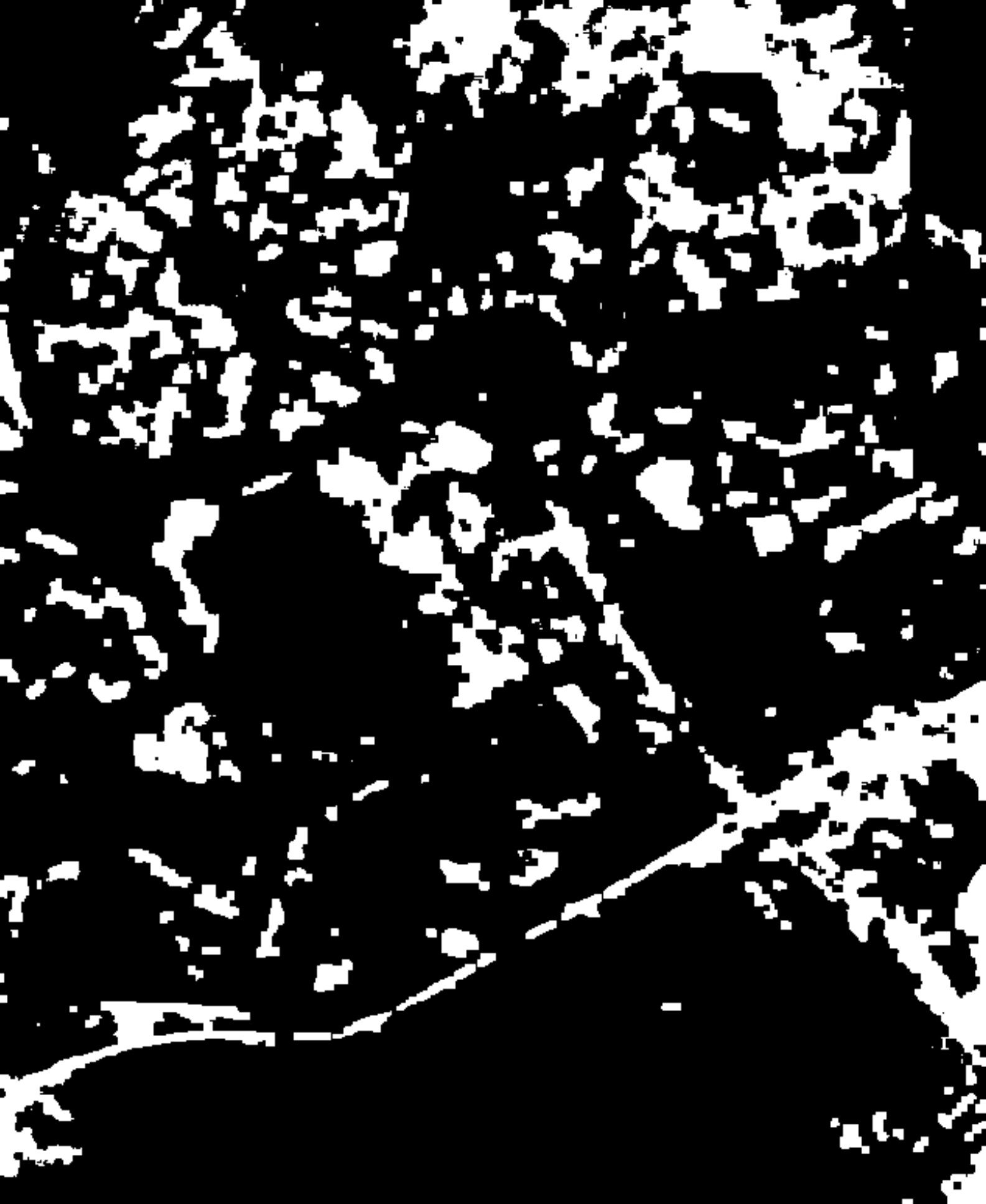}%
\label{Dubai SiROC}}
\hfil
\subfloat[SiROC (No MP)]{\includegraphics[width=1.5 in]{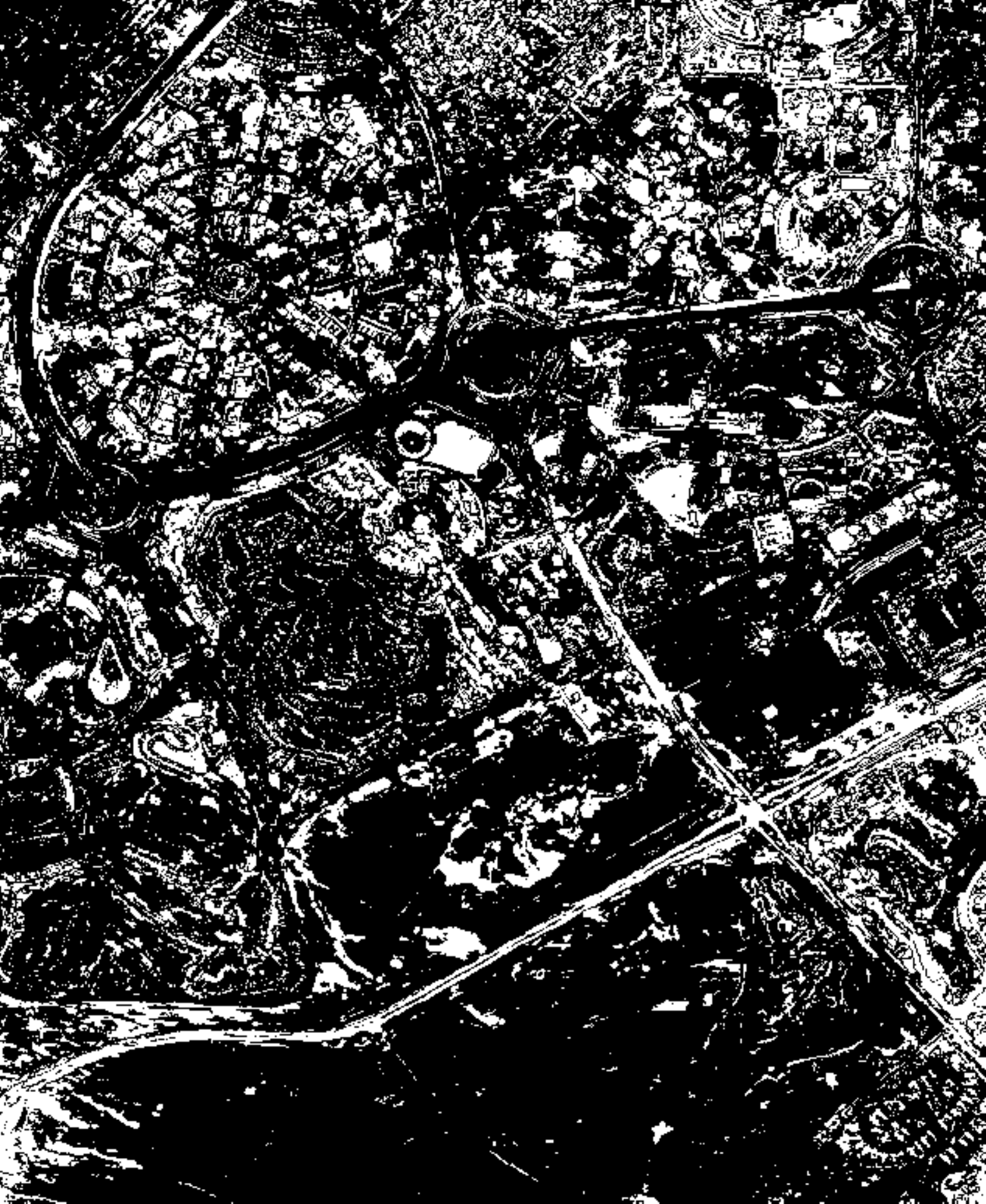}%
\label{Dubai SiROC No MP}}
\hfil
\subfloat[DCVAMR]{\includegraphics[width=1.5 in]{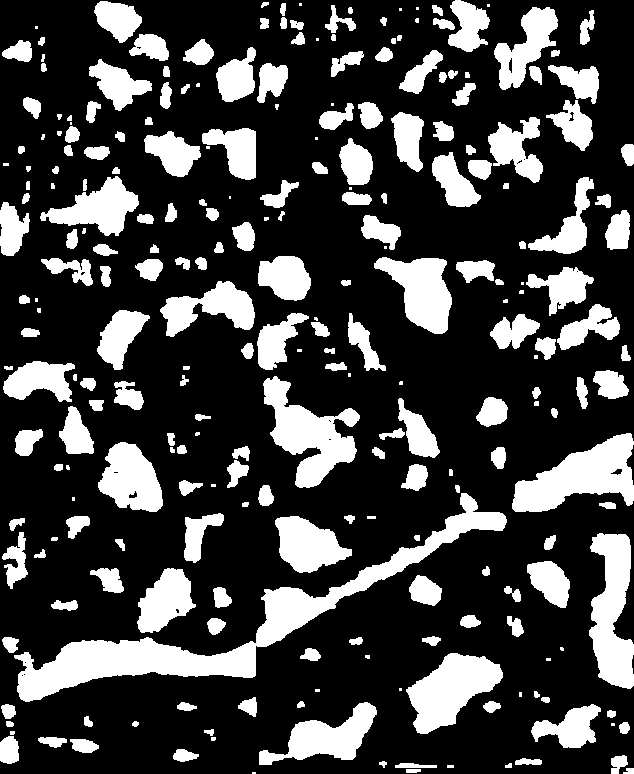}%
\label{Dubai DTL}}
\hfil
\subfloat[DCVA]{\includegraphics[width=1.5 in]{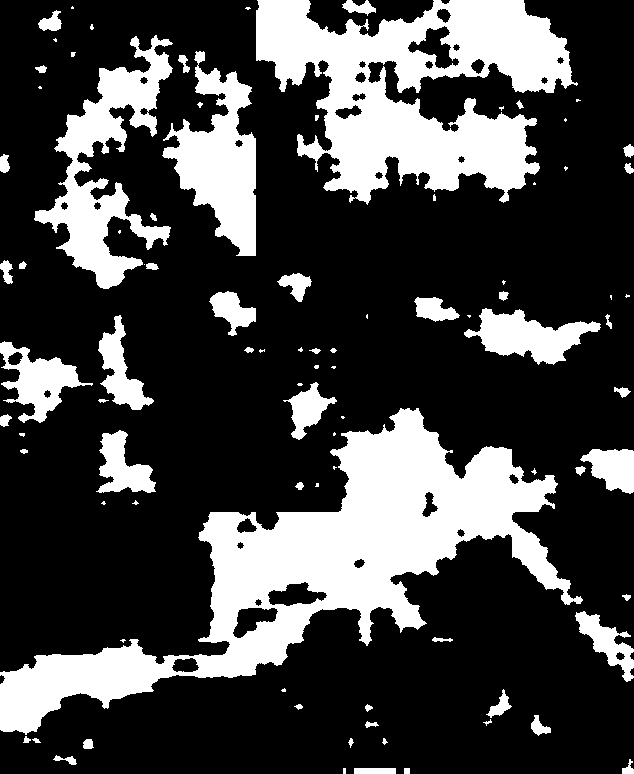}%
\label{Dubai DCVA}}
\hfil
\subfloat[PCVA]{\includegraphics[width=1.5 in]{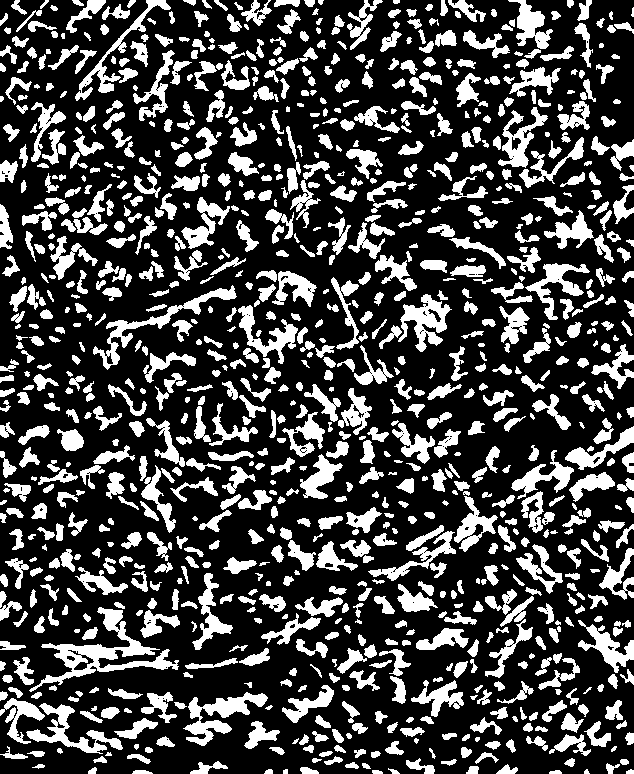}%
\label{Dubai PCVA}}
\hfil
\subfloat[RCVA]{\includegraphics[width=1.5 in]{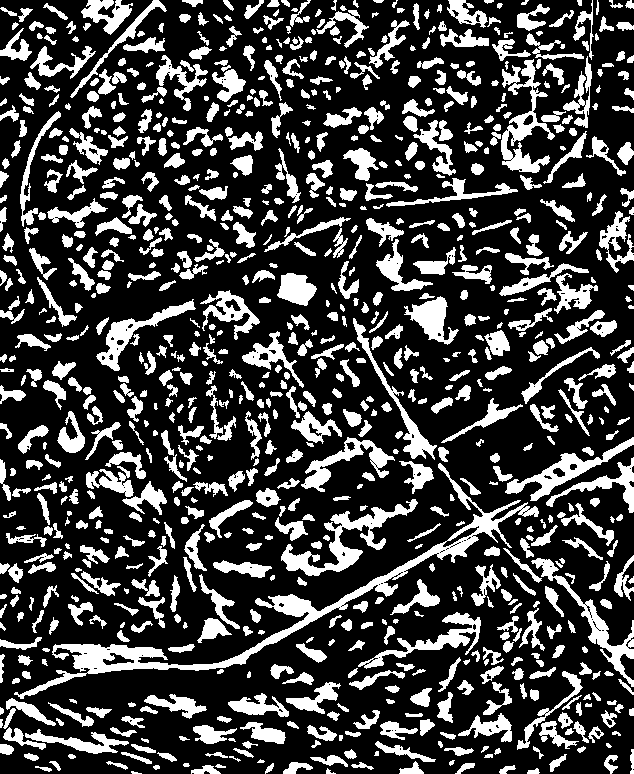}%
\label{Dubai RCVA}}
\hfil
\subfloat[Ground Truth]{\includegraphics[width=1.5 in]{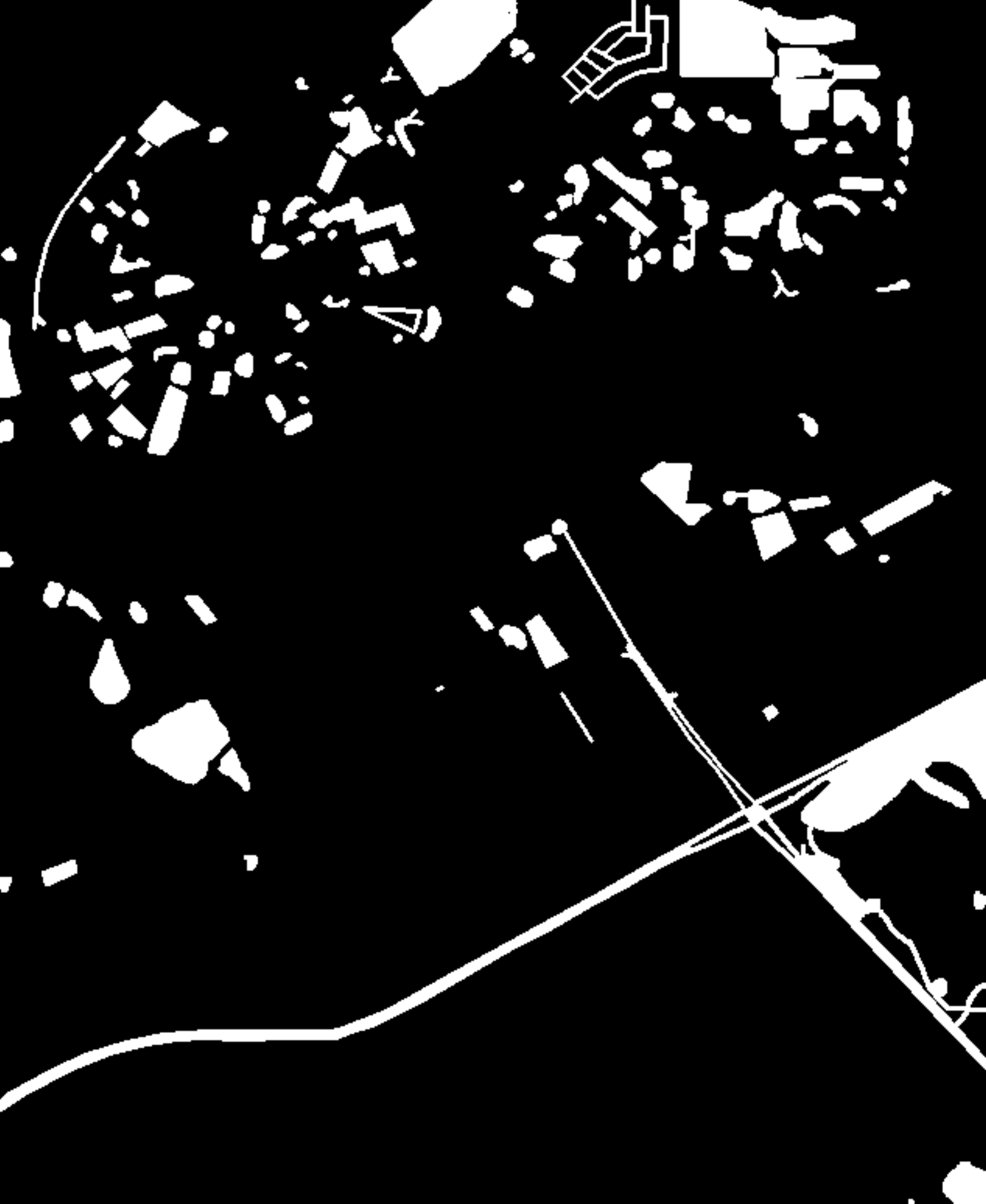}%
\label{fig_second_case}}
\caption{Qualitative Comparison OSCD - Dubai. The structure is identical to Figure \ref{fig:Las Vegas} but predictions and ground truth are presented for Dubai. Also for this scene, SiROC predicts changing areas and their shape comparably well even. In contrast, competing methods miss the shapes of changing areas such as the street in the lower part of the image or struggle to detect relevant regions.}
\label{fig:Beirut}
\end{figure*}

A similar picture emerges when inspecting results for Dubai, which is a slightly more complex scene since the shapes of changes differ widely. SiROC detects changing regions again well but seems to struggle with the shape of changes in the upper part of the image. The newly constructed road is identified well. Consequently, the quantitative scores on this scene are slightly lower compared to the Las Vegas Scene with 86.87 \% specificity, 76.61\% sensitivity and 39.14\% precision. The struggles of the competing methods are similar to the Las Vegas Scene: DCVAMR fits round shapes to any kind of change (Figure \ref{Dubai DTL}), DCVA overestimates the size of large changes (Figure \ref{Dubai DCVA}), and PCVA extracts a spurious change map that rather looks like building footprints (Figure  \ref{Dubai PCVA}). Therefore, the quality inspection of visual results confirms that SiROC obtains superior results on OSCD.

\begin{figure*}[!htbp]
\centering
\subfloat[Las Vegas]{\includegraphics[width=3 in]{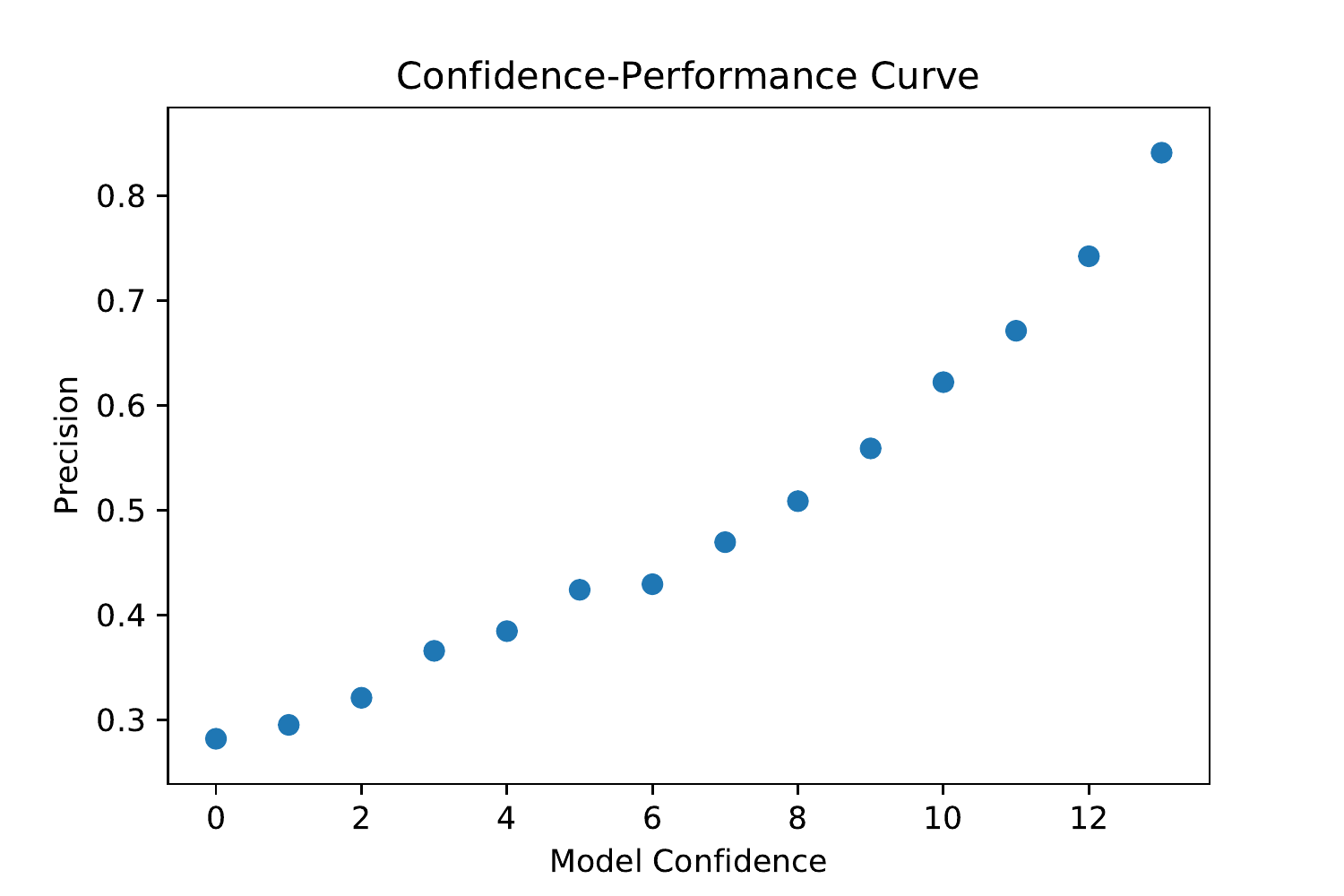}%
\label{Confidence Vegas}}
\hfil
\subfloat[Dubai]{\includegraphics[width=3 in]{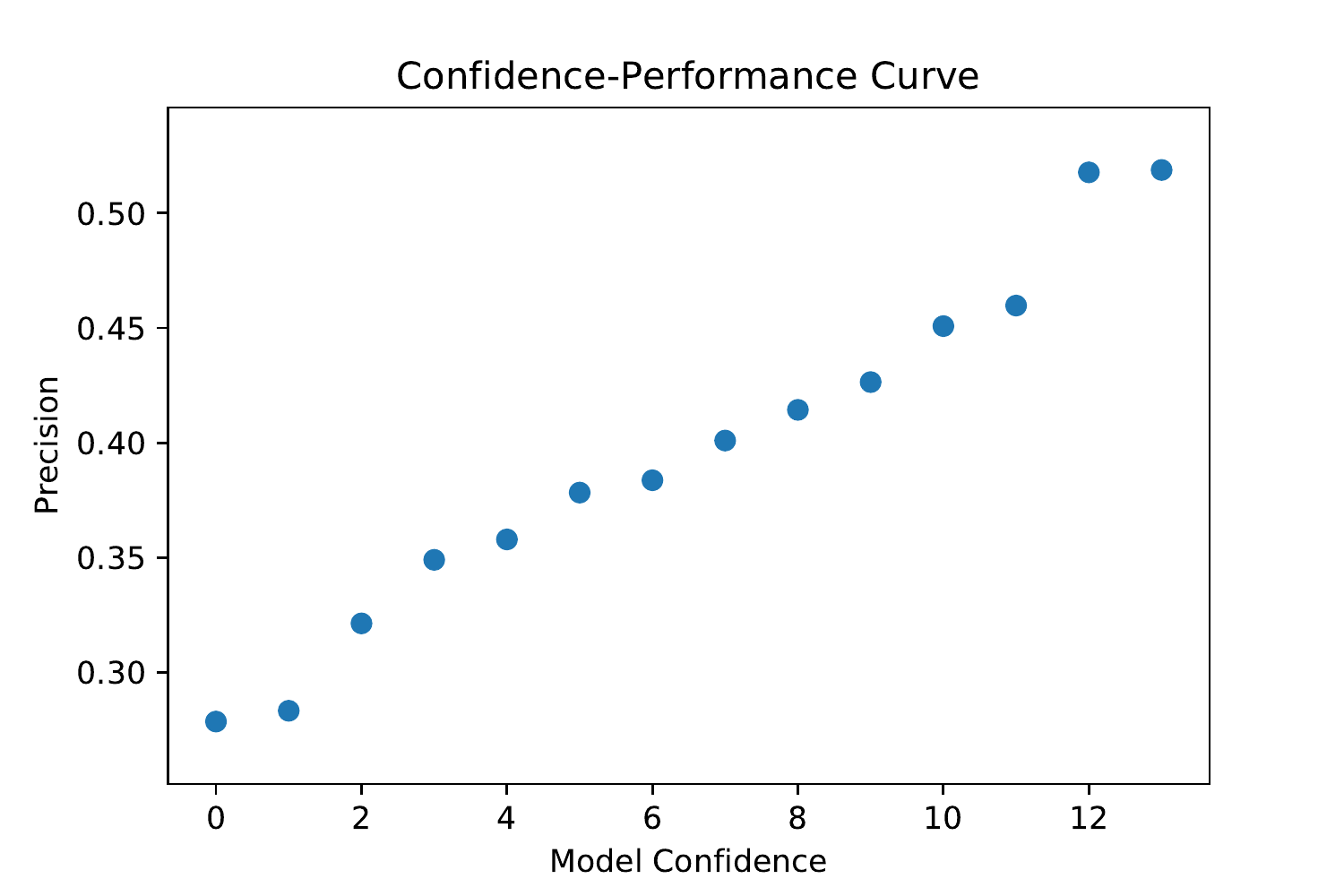}%
\label{Confidence Dubai}}
\hfil
\subfloat[Chongping]{\includegraphics[width=3 in]{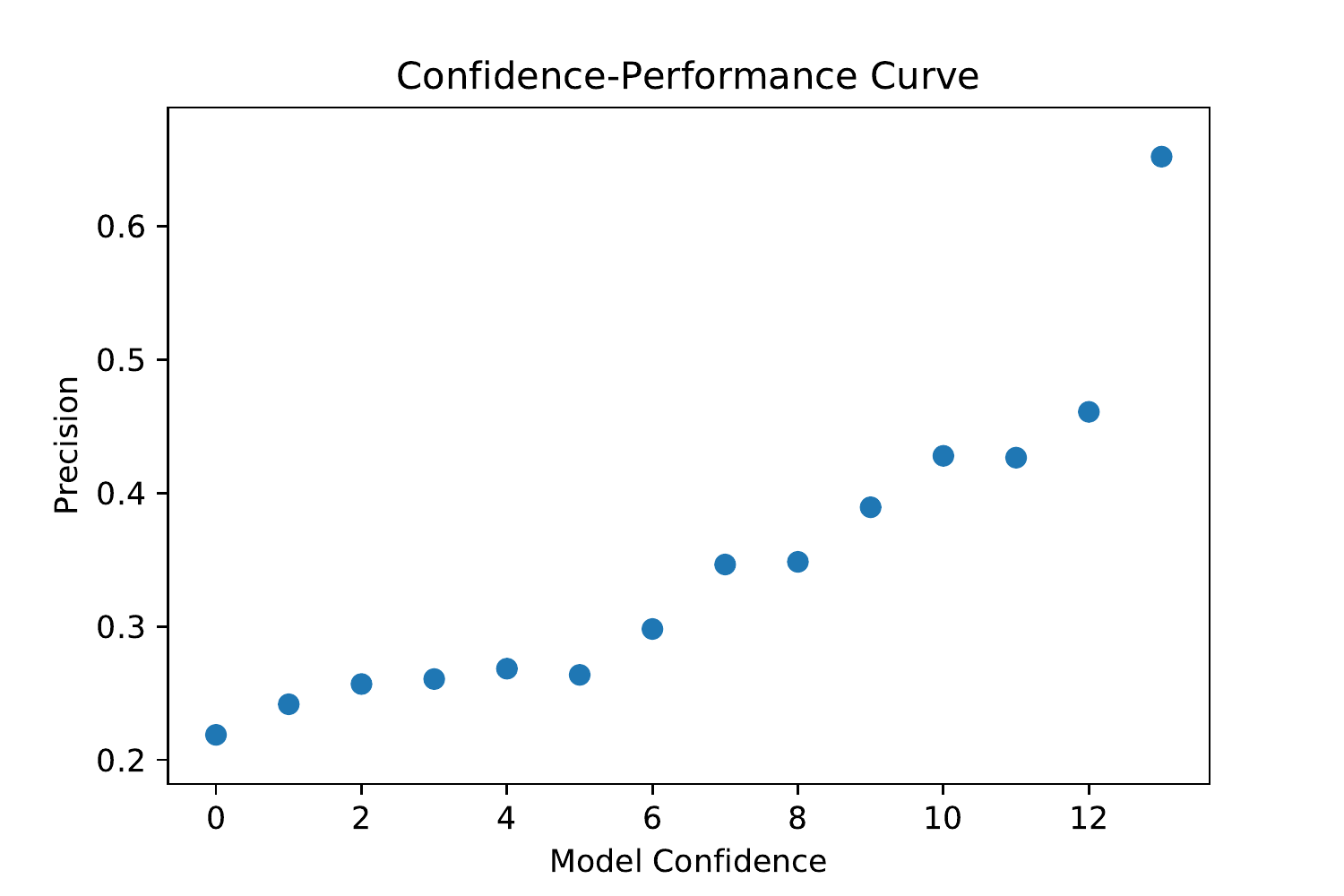}%
\label{Confidence Chongping}}
\hfil
\subfloat[Montpellier]{\includegraphics[width=3 in]{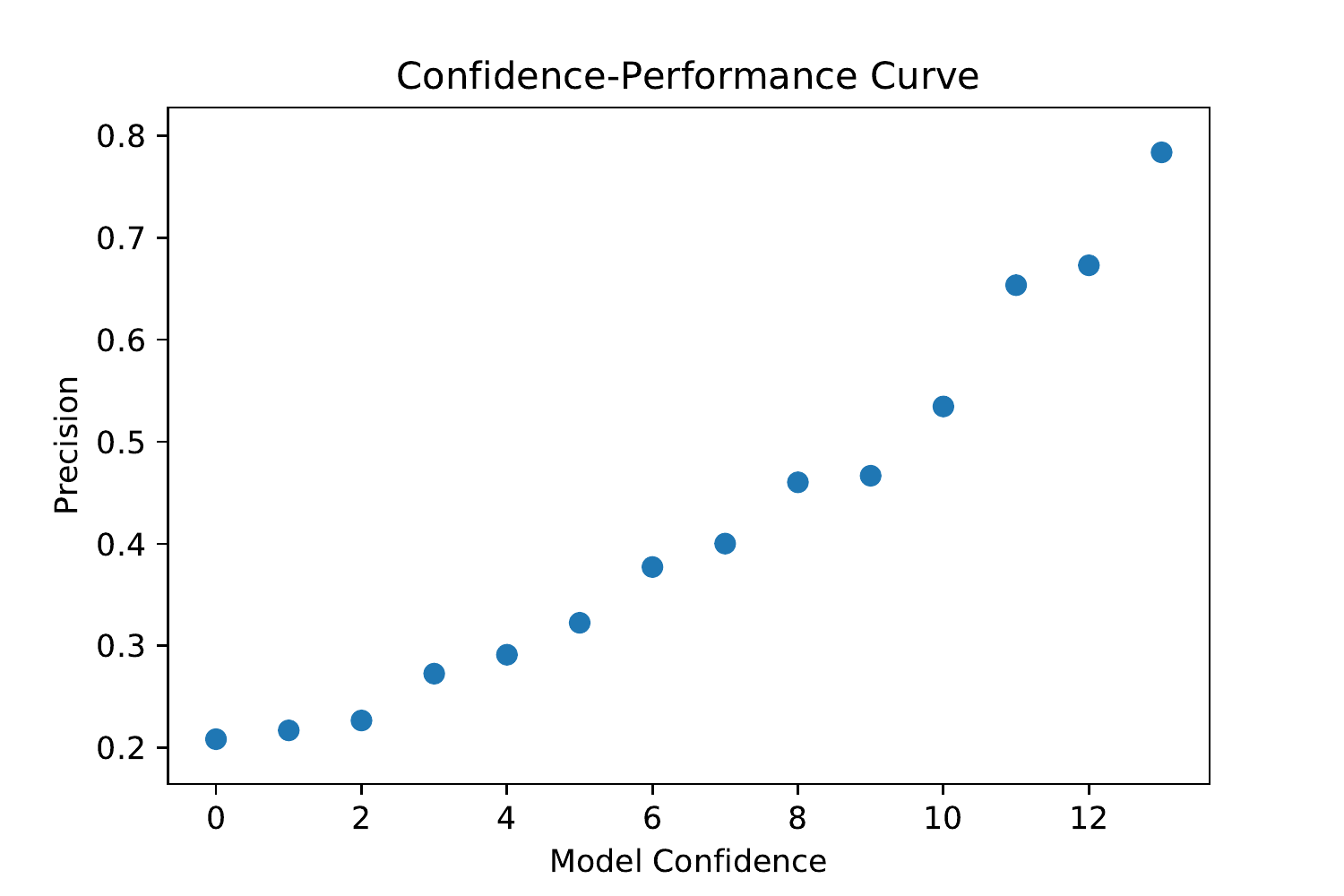}%
\label{Confidence Montpellier}}
\caption{Confidence-Performance Plots on four cities of the OSCD Dataset. On the x-axis, point in the image are sorted into buckets by SiROC confidence. For each of these buckets, the performance is estimated separately. Since performance is generally non-decreasing in confidence, the uncertainty measure is considered well calibrated.}
\label{fig:Uncertainty}
\end{figure*}

\textit{Uncertainty Estimation}:
To properly analyze if the confidence of SiROC also corresponds to well-calibrated uncertainties, we test this with calibration curves. 
For this, we split pixels into subsets based on the SiROC confidence and analyze the respective performance for a level of confidence. If the performance of SiROC is in principle increasing with the confidence, the uncertainty levels in fact correspond to the certainty of the prediction the model has. Figure \ref{fig:Uncertainty} plots these confidence-performance curves for four cities in the OSCD test set. For all four cities, we see that model precision is non-decreasing in the confidence of the SiROC. Most of the time, the prediction increases notably in the confidence which means that SiROC not only performs well for this task but also returns well-calibrated uncertainties as part of its prediction.

\begin{table}
\centering
\caption{Sensitivity to Hyperparameters (OSCD Training Set)}
\begin{threeparttable}
\footnotesize
\setlength{\tabcolsep}{\tabcolsep}
\begin{tabular}{lcccc}
\toprule
{} & Specificity & Sensitivity & Range & \# Evaluations \\
\midrule
Selected & 89.48\% & 67.90\% &  \\
\midrule
N\_max  & 87.96\% & 70.53\% & [30,250] & 20 \\
e\_start  & 89.37\% & 68.59\% & [0,150] & 20 \\
s  & 88.89\% & 68.06\% & [1,5] & 5\\
p  & 91.81\% & 59.06\% & [2,5] & 4\\
Joint  & 90.34\% & 60.12\% & All above & 75 \\

\bottomrule
\end{tabular}\textbf{}
\end{threeparttable}
\label{tab:Sensitivity_OSCD}
\begin{justify}
Average Specificity and Sensitivity on the OSCD training set when varying SiROC hyperparameters. The average scores are compared to the selected optimal selection of parameters and their training performance in the first line. For single parameters (rows 2-5), we vary only the mentioned parameter on the grid given in the column range and leave the others at the selected optimum. The column Evaluations gives the number of runs that were executed to obtain the average scores. Finally, in the last row "Joint" we vary all parameters on the given intervals in the rows above simultaneously. 
\end{justify}
\end{table}

\textit{Sensitivity to Hyperparamenters}: 
To allow effective use of SiROC in practice, we offer a sensitivity analysis of the hyperparameter choice on OSCD along with recommendations for this choice in other applications. This sensitivity analysis is executed on the training set to avoid multiple evaluations on the test set. The results are shown in Table \ref{tab:Sensitivity_OSCD}. While the performance of the method naturally varies with the choice of hyperparameters, SiROC looks fairly robust against its hyperparameter choices. The first row gives the training set performance based on the selected parameters described in section C as a comparison point. Varying only the maximum neighborhood N\_max, the number of rows excluded e\_start and the stepsize s at the selected parameter specification influences the training performance marginally, at most. For all three parameters, the average specificity decreases while average sensitivity increases slightly. These three parameters essentially navigate how to group and prioritize neighborhoods. Excluding close context (e\_start), including more distant context (N\_max), and aggregating neighborhoods into larger groups (s) therefore does not seem to matter notably in practice to achieve good performance. The performance is slightly more sensitive towards the size of the morphological profile (p) where average specificity increases marginally and sensitivity drops by 9 p.p. if this is varied leaving other parameters untouched. Similarly, when varying all four parameters simultaneously in 75 random draws, performance drops with a difference of about 8 p.p. in sensitivity with similar specificity. In terms of magnitude, this performance drop is only a fraction of the difference between SiROC and its closest competitors on the OSCD test set. This implies that SiROC would likely outperform competing methods on this dataset for a variety of hyperparameter choices. 

For potential other applications of SiROC in the future we suggest using the obtained parameter combination initially. This provides a starting point for further analysis in different contexts. Since the performance seems to be comparably susceptible to the size of the morphological profile, this parameter may deserve special attention during tuning. In the following, results on the remaining three datasets are obtained with this parameter combination which was the result of tuning on OSCD. Even though this may not necessarily give the best possible performance, we aim to validate that SiROC achieves convincing results in other applications without finetuning on single scenes.

\subsection{Results on BHED}\label{subsec:beirut}

\begin{table}
\centering
\caption{Quantitative Results Beirut Explosion}
\begin{threeparttable}
\footnotesize
\setlength{\tabcolsep}{\tabcolsep}
\begin{tabular}{lcccc}
\toprule
{} & Specificity & Sensitivity & Precision & F1 \\
\midrule
SiROC & \textbf{92.01\%} & \textbf{83.38\%} & \textbf{19.89\%} & \textbf{32.12\%} \\
DCVA
& 91.87\% & 79.85\% & 11.37\% & 19.93\% \\
SSDCVA
& 88.25\% & 81.08\%  & 8.80\% & 15.95\%\\
PCVA  & 88.61\% & 58.56\%  & 6.74\% & 12.10\% \\
RCVA  & 86.56\% & 66.71\% & 6.52\% & 11.89\% \\

\midrule
\multicolumn{5}{c}{Ablation Scores} \\
\midrule
SiROC (p=10) & \textbf{92.34\%} & \textbf{91.89\%} & \textbf{22.20\%} & \textbf{35.76\%} \\
no MP  & 88.02\% & 79.67\%  & 13.65\% & 23.30\% \\
HSR  & 86.65\% & 71.63\%  & 11.31\% & 19.54\%\\

\bottomrule
\end{tabular}\textbf{}
\end{threeparttable}
\label{tab:Results_Beirut}
\end{table}

\begin{figure*}[!htbp]
\centering
\subfloat[Pre Image]{\includegraphics[width=2 in]{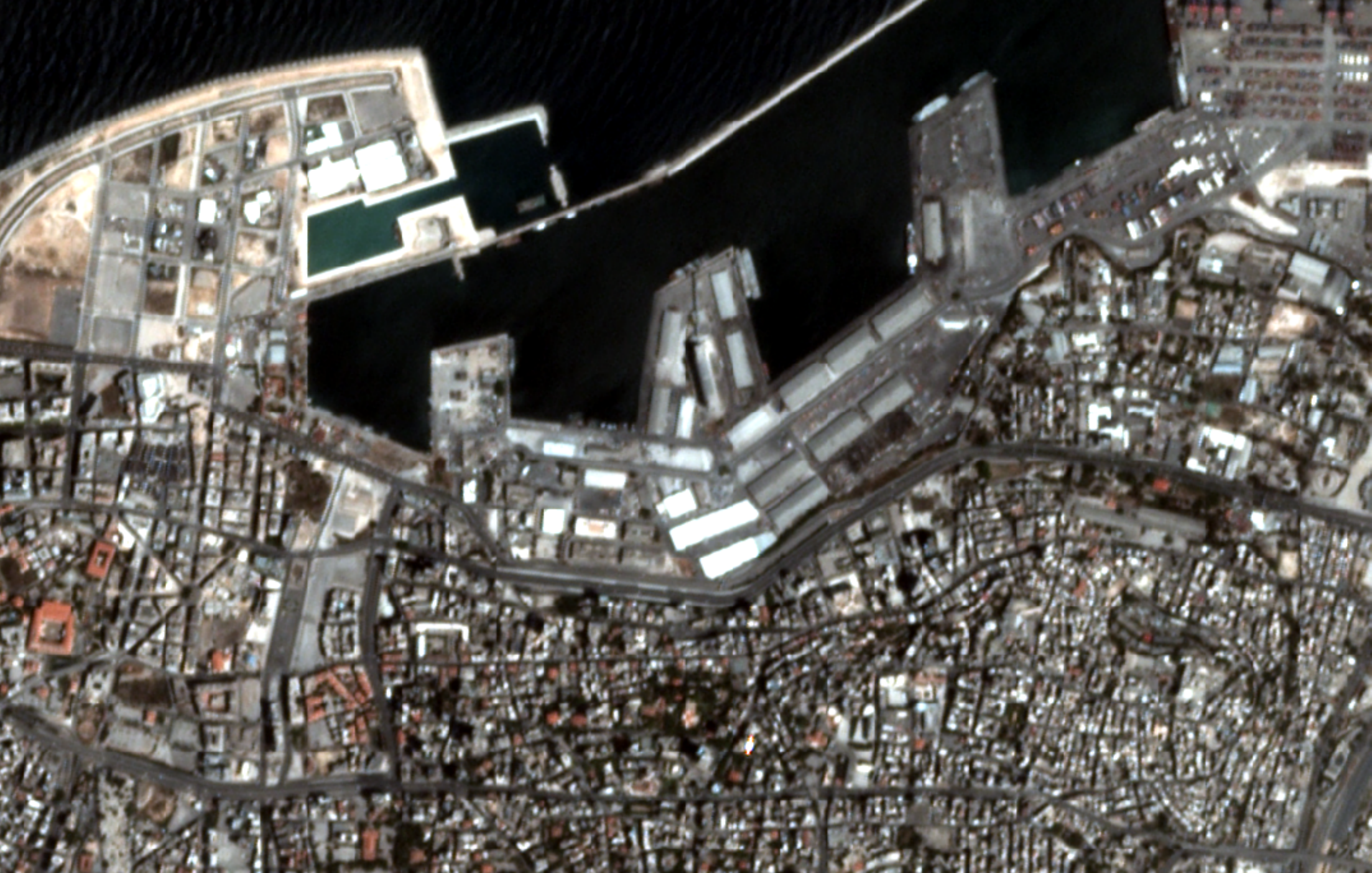}%
\label{Beirut Pre}}
\hfil
\subfloat[Post Image]{\includegraphics[width=2 in]{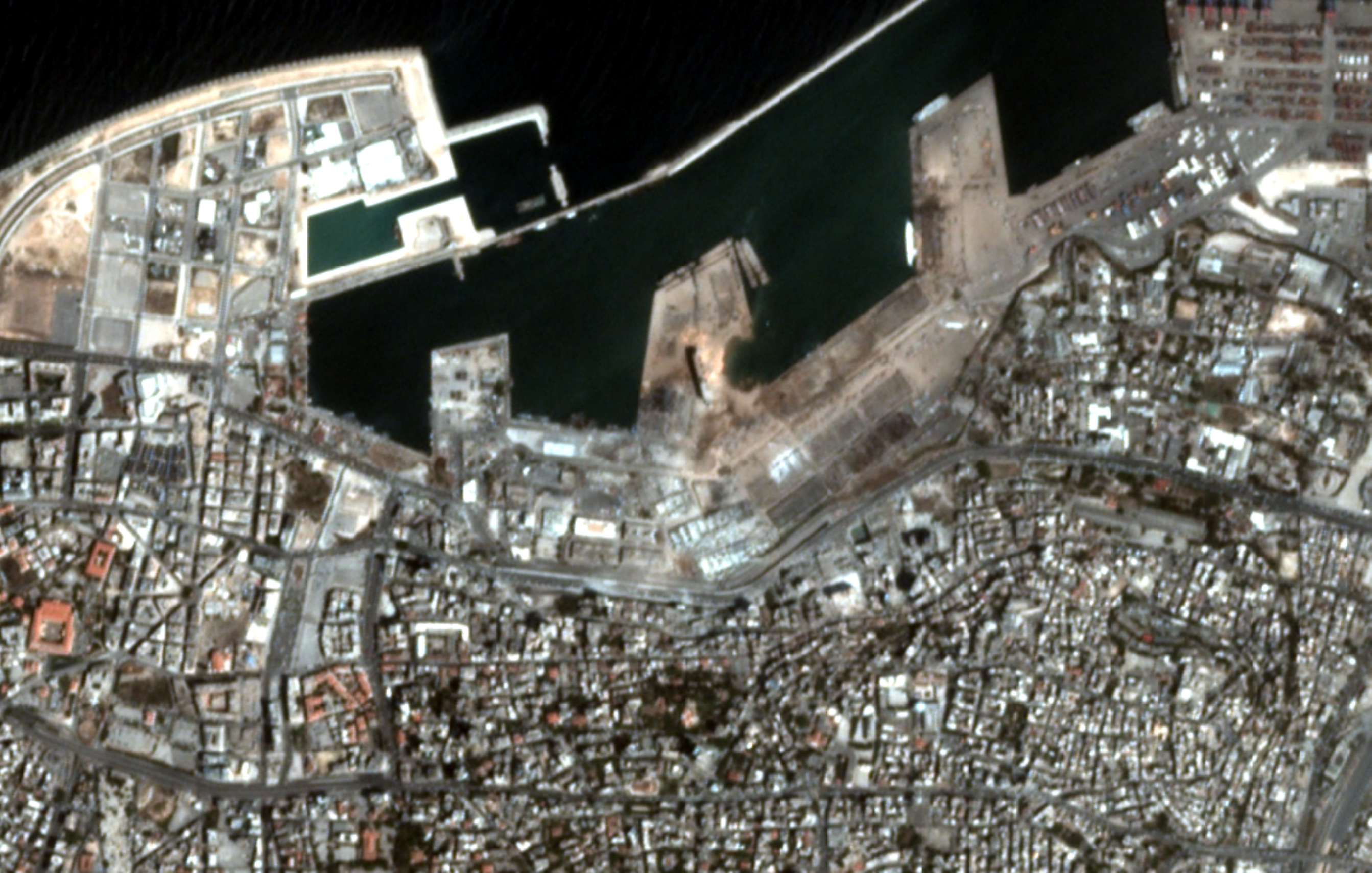}%
\label{Beirut Post}}
\hfil
\subfloat[Ground Truth]{\includegraphics[width=2 in]{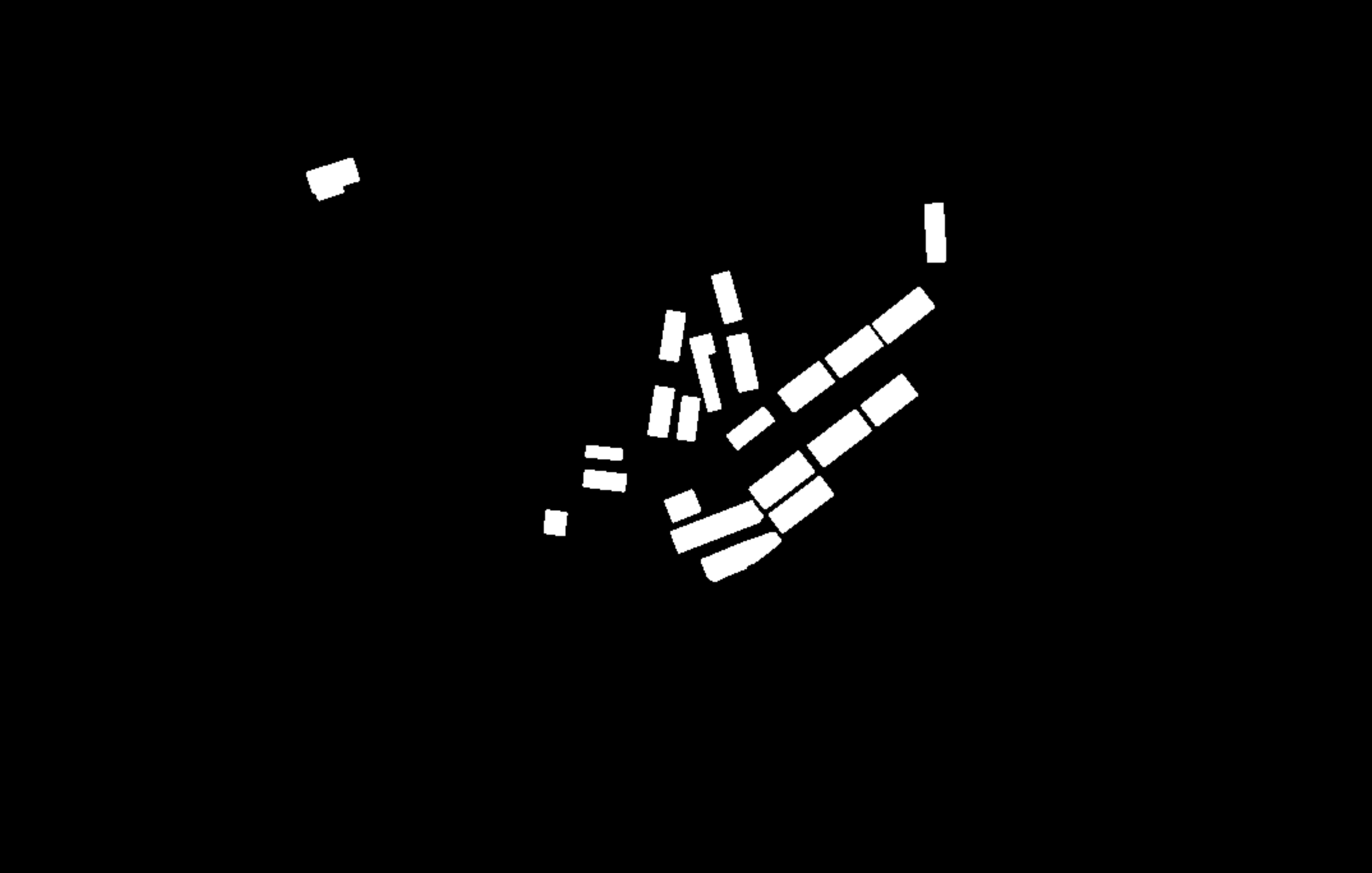}%
\label{Beirut Labels}}
\hfil
\subfloat[SiROC (Ours)]{\includegraphics[width=2 in]{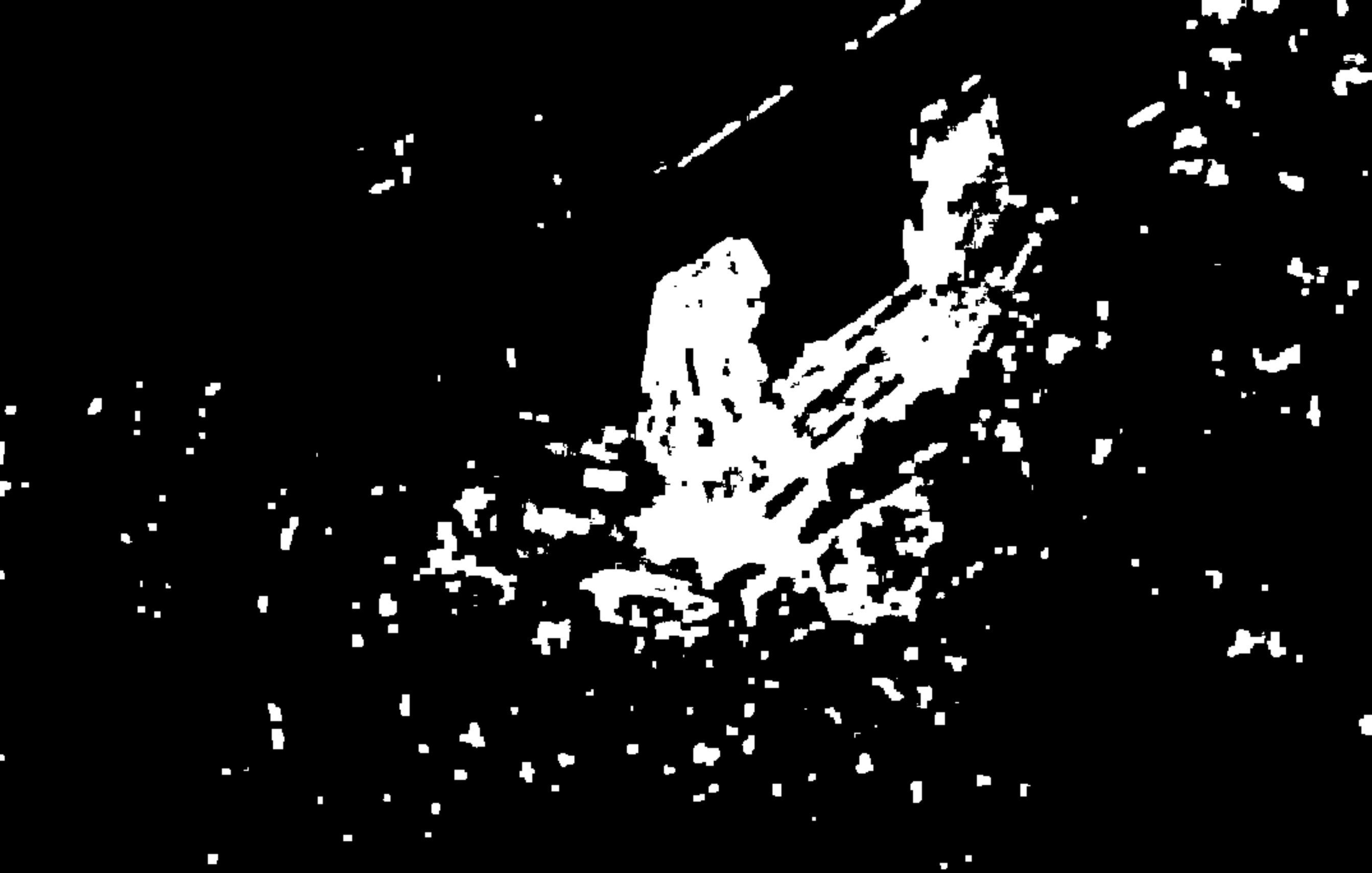}%
\label{Beirut Sirco 5}}
\hfil
\subfloat[SiROC (p=10)]{\includegraphics[width=2 in]{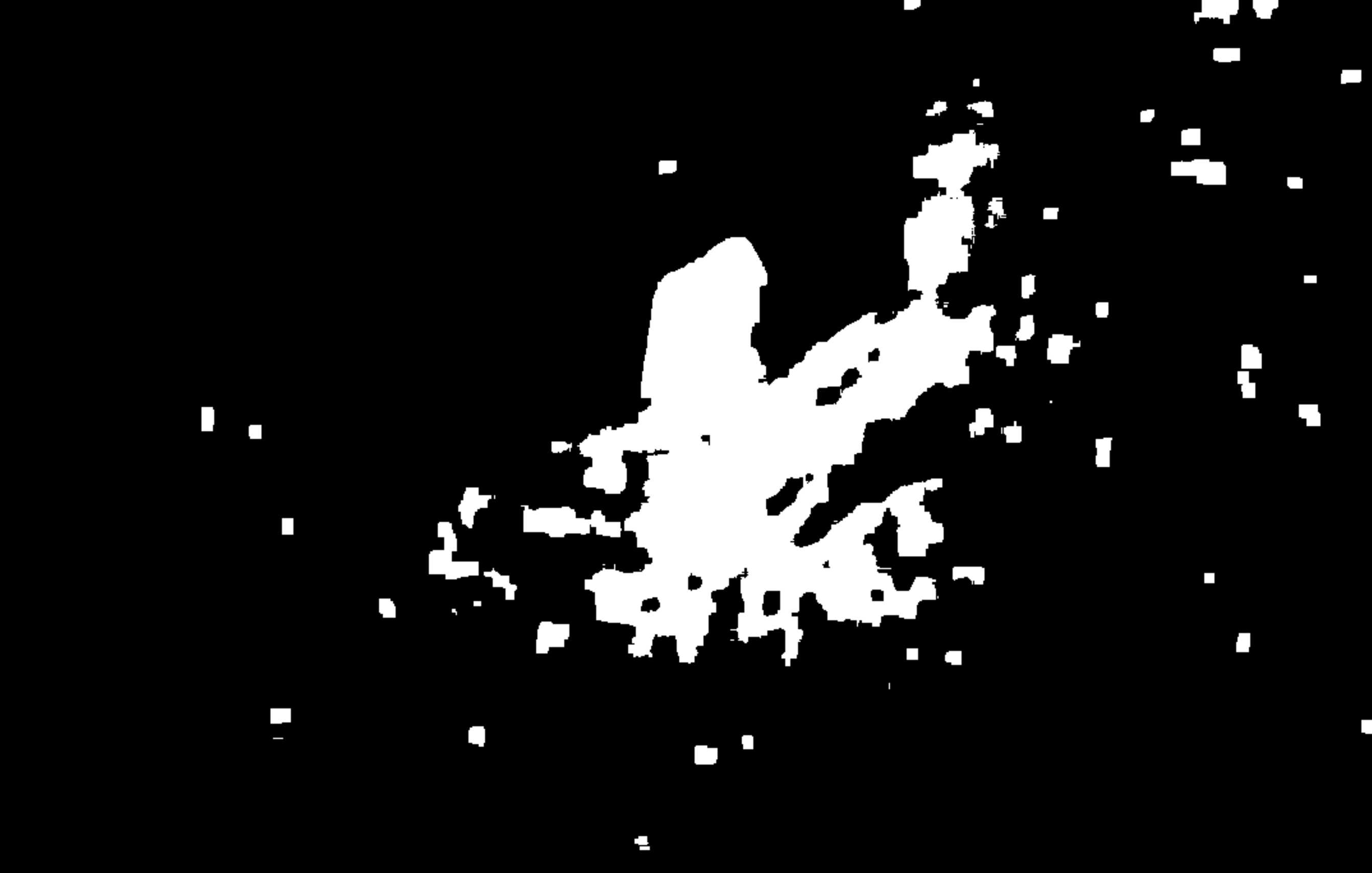}%
\label{Beirut SiROC 10}}
\hfil
\subfloat[SiROC (No MP)]{\includegraphics[width=2 in]{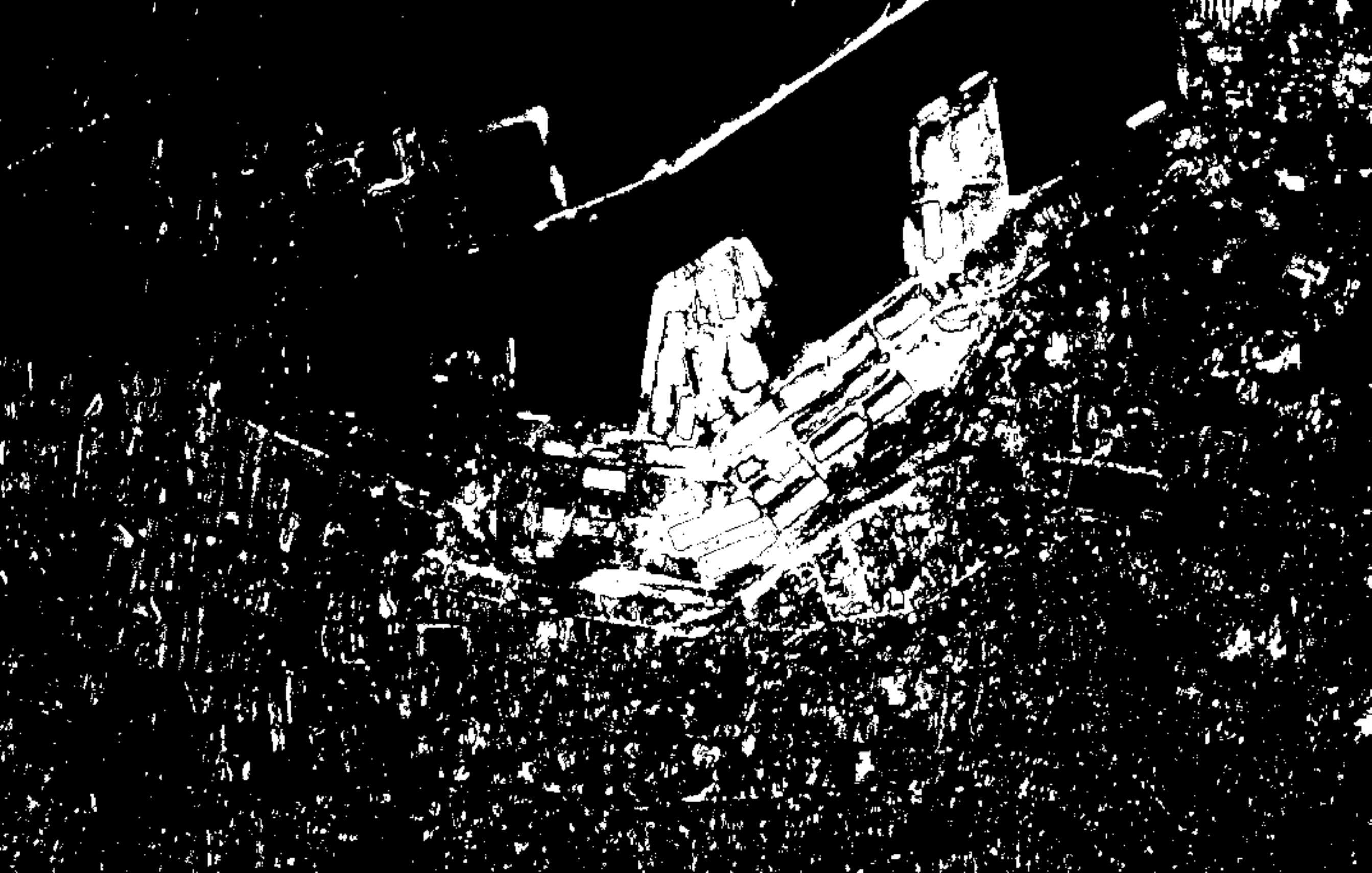}%
\label{Beirut SiROC No MP}}
\hfil
\subfloat[SSDCVA]{\includegraphics[width=2 in]{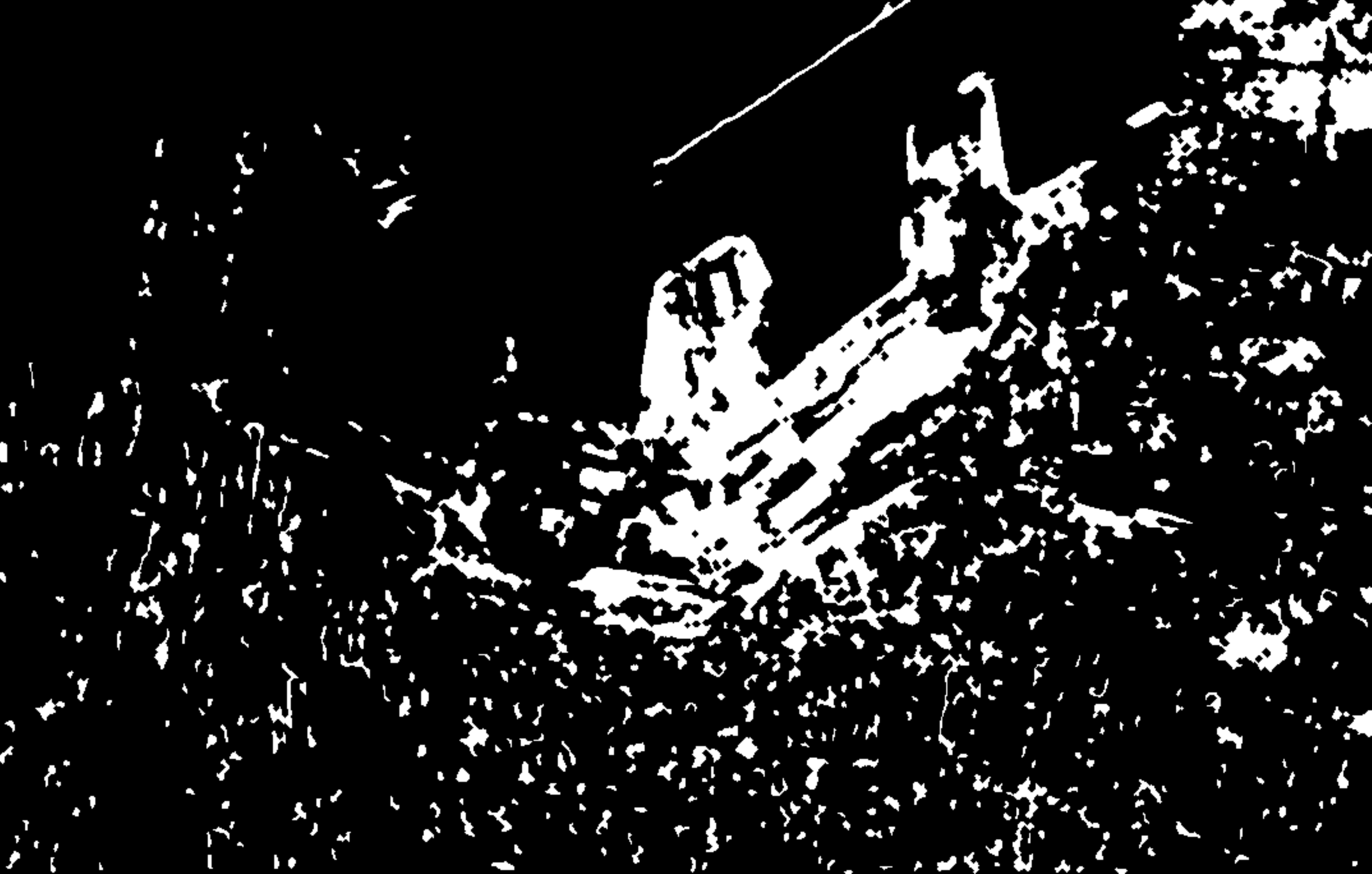}%
\label{Beirut SSDCVA}}
\hfil
\subfloat[DCVA]{\includegraphics[width=2 in]{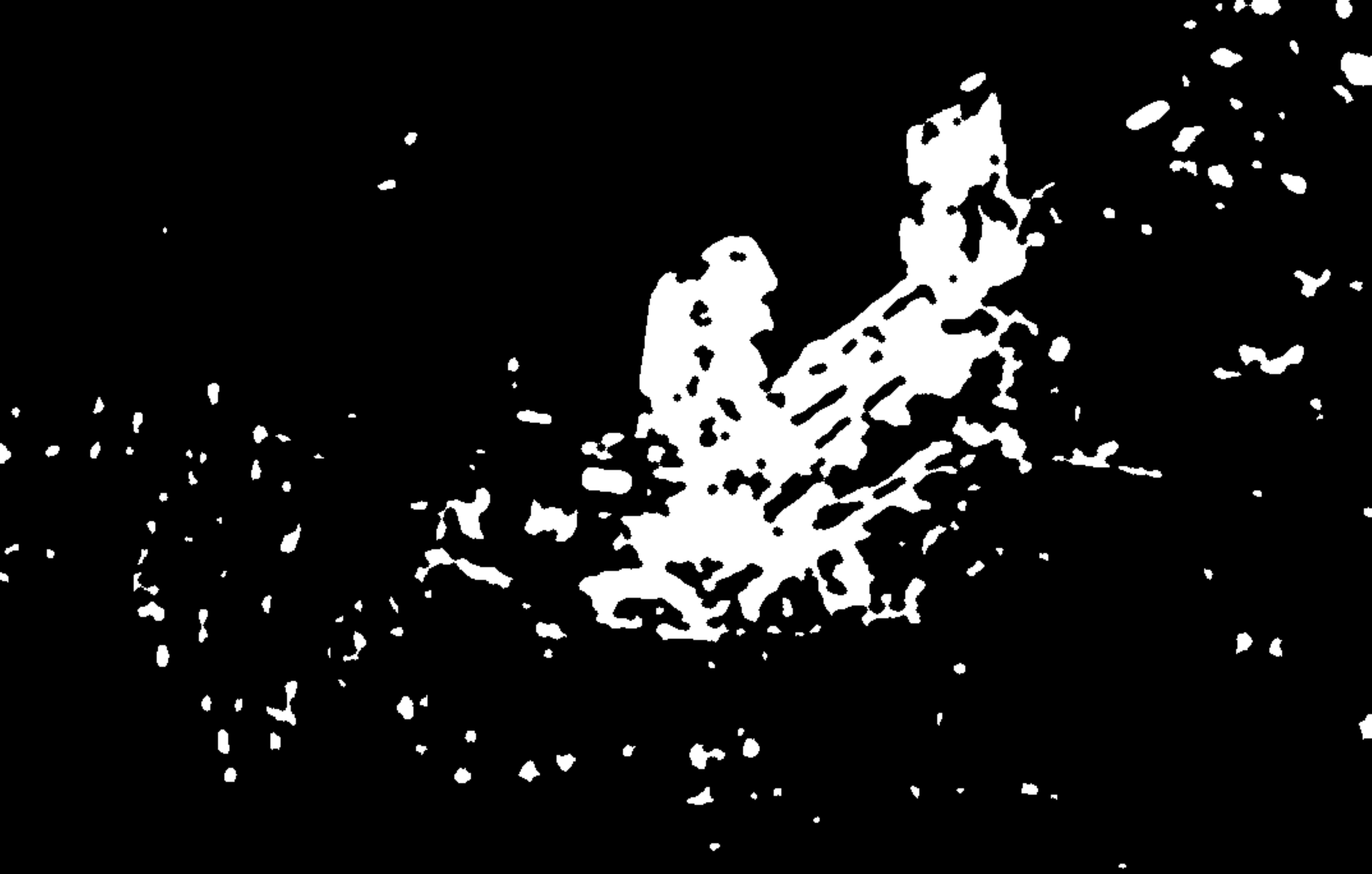}%
\label{Beirut DCVA}}
\hfil
\subfloat[RCVA]{\includegraphics[width=2 in]{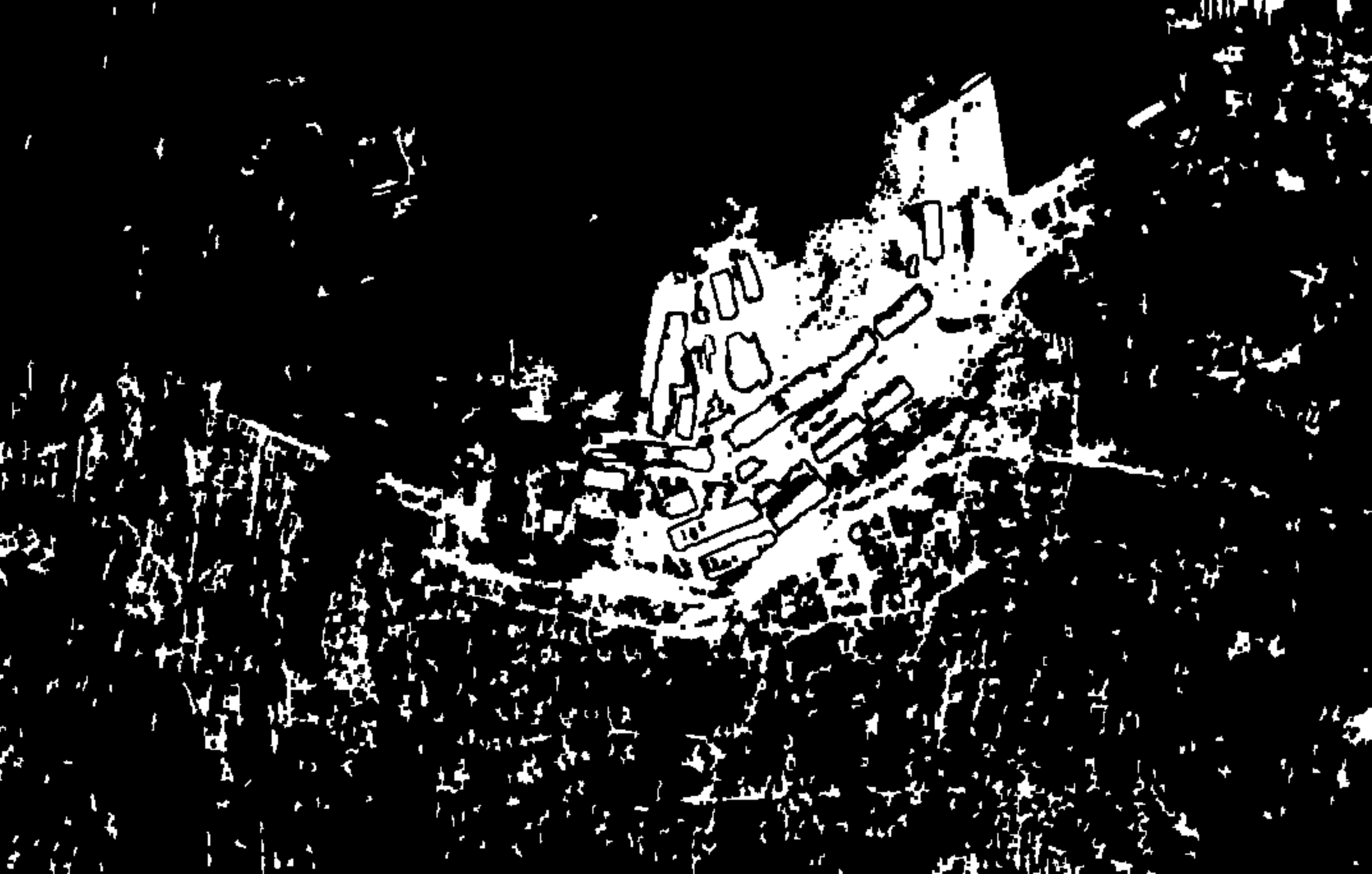}%
\label{Beirut RCVA}}
\hfil
\caption{Qualitative Comparison Beirut Explosion. Figure \ref{fig:Beirut} shows the Planetscope image pair (\ref{Beirut Pre} \& \ref{Beirut Post}), change ground truth (\ref{Beirut Labels}) and model predictions (\ref{Beirut Sirco 5} - \ref{Beirut RCVA}) for the Beirut Explosion scene. SiROC with main parameters (\ref{Beirut Sirco 5}) identifies the area of destroyed buildings around the epicenter correctly with few false positives although the shapes are lacking some granularity. Increasing the size of morphological operations improves accuracy but tends to fit one large blob with missing building shapes \ref{Beirut SiROC 10}. Excluding morphological operations increases false positives although the main changes in the center are still idenfitied well (\ref{Beirut SiROC No MP}). Competing methods struggle not only with the shape of change but also detect a number of false positives far away from the explosion (\ref{Beirut SSDCVA}-\ref{Beirut RCVA}).}
\label{fig:Beirut}
\end{figure*}

\textit{Quantitative Results}:
Table \ref{tab:Results_Beirut} displays specificity, sensitivity, precision and F1 scores on the scene. Generally, scores on BHED are higher than on OSCD since the changes are centered around the same area and have similar shapes. SiROC with default parameters achieves a specificity of 92.01\% and a sensitivity of 83.38\%. DCVA achieves a similar specificity with 91.87\% but falls short in terms of sensitivity by about 4 p.p. with a score of 79.85\%. SSDCVA places slightly below DCVA with a specificity of 88.25\% and a sensitivity of 81.08\%. SiROC beats SSDCVA by about 3 p.p. on sensitivity and about 2 p.p. on sensitivity. PCVA and RCVA clearly fall behind SiROC and also DCVA-based methods. F1 score and precision results confirm the previous impressions with a gap of 12-20 p.p. in F1 and 9-13 p.p. precision respectively. When we adjust the scale of morphological operations to 10, SiROC performs significantly better which suggests that there may be notable tuning potential for higher resolution inputs. Still, the baseline parameters perform well on this scene. Therefore, SiROC demonstrates its usefulness beyond medium-resolution images and can also be used in conjunction with high-resolution images for CD. 

The other ablation scores again point towards the most important steps within SiROC to achieve this performance. Without morphological profiles, the scores of SiROC drop about 4-9 p.p. in all four categories. Still, it achieves slightly superior precision and F1 scores but falls short to DCVA with a difference of about 3 p.p. in specificity and similar sensitivity. This is a notable difference to medium-resolution imagery on OSCD where the exclusion of morphological filters decreased the performance of SiROC but it was still superior to DCVA-based methods. This is not necessarily surprising since deep learning-based methods tend to relatively improve their CD performance compared to traditional methods with increasing spatial resolution. Without majority voting over different neighborhoods, the Vanilla HSR version performs better but in the range of RCVA and PCVA. Again, it is the combination of HSR, ensembling over different neighborhoods and transitioning to the object level with morphological operations that all contribute significantly to the overall performance of SiROC. 

\textit{Qualitative Results}:
Figure \ref{fig:Beirut} shows visual comparisons of the discussed methods on BHED. The first row of images presents the pre-explosion image (Figure \ref{Beirut Pre}), the post image (\ref{Beirut Post}) and the ground truth (\ref{Beirut Labels}). The heart of the explosion in the port can be found in the middle of the image with almost the entirety of buildings completely destroyed around it. Panel \ref{Beirut Sirco 5} presents the binary SiROC segmentation with baseline parameters obtained on OSCD. While SiROC is missing some granularity in its segmentation of destroyed building footprints, the changing areas are well identified with few false positives outside of the port. For a larger morphological filter size (p), the main area is identified more densely with better quantitative results but the shapes of buildings vanish (\ref{Beirut SiROC 10}). Without morphological operations, the core change is still well-segmented although the number of false positives in the outer regions of the image increases (\ref{Beirut SiROC No MP}).
SSDCVA shows similar tendencies to summarize the port area as one large change with a number of spurious false positives (\ref{Beirut SSDCVA}). DCVA shows fewer salt and pepper noise than SSDCVA here and generally segments the exploded buildings similar to SiROC, however, with a slightly more perforated shape (\ref{Beirut DCVA}). The segmentation by RCVA is not really competitive here since the maps are spurious and changes are not well identified (\ref{Beirut RCVA}). Results for PCVA are similar to RCVA and hence omitted.

\subsection{Results on Agriculture Dataset}\label{subsec:agriculture}
\textit{Quantitative Results}: Table \ref{tab:Results_Agriculture} displays the results of SiROC and competing methods on the agriculture scene. SiROC is applied to the dataset with the parameters obtained on Onera without further adjustment. Hence, the results we provide are a validation exercise in the different context of non-visible parts of the spectrum without parameter finetuning. 

To be consistent with previous evaluations on this dataset \cite{saha2019unsupervised}, we compare SiROC with PCVA and RCVA based on vegetation (VEG) and near-infrared (NIR) channels of Sentinel-2 as inputs. The score for DCVAMR is based on the full Sentinel-2 input images as the method was deliberately designed to incorporate all channels.

While SiROC achieves the top score in terms of specificity and precision with 90.81\% and 74.23\% respectively, it falls short of DCVAMR on sensitivity (88.70\% vs 94.26\%) and F1 score (80.85\% vs 81.47\%). DCVAMR seems to lean slightly more towards the change class whereas SiROC rather classifies a pixel as no change in unclear cases. SiROC is superior to PCVA and comparable to RCVA in performance for both, VEG and NIR channels as inputs.

The ablation scores underline that morphological profiles still help although the effects are smaller than in urban applications with an average difference in about 1-2 percentage points in all four criteria. Further, excluding the majority voting mechanism does not hurt performance but actually improves it slightly here. The vanilla HSR performs slightly worse but in the range of RCVA and better than PCVA on its own. Smaller benefits of including majority voting and morphological profiles could be linked to the fact that parameters for these operations were tuned in an urban RGB context. Although already quite effective, the accuracy of SiROC could likely be further improved with parameter finetuning.

\begin{figure*}[!htbp]
\centering
\subfloat[Pre Image RGB]{\includegraphics[width=1.5 in]{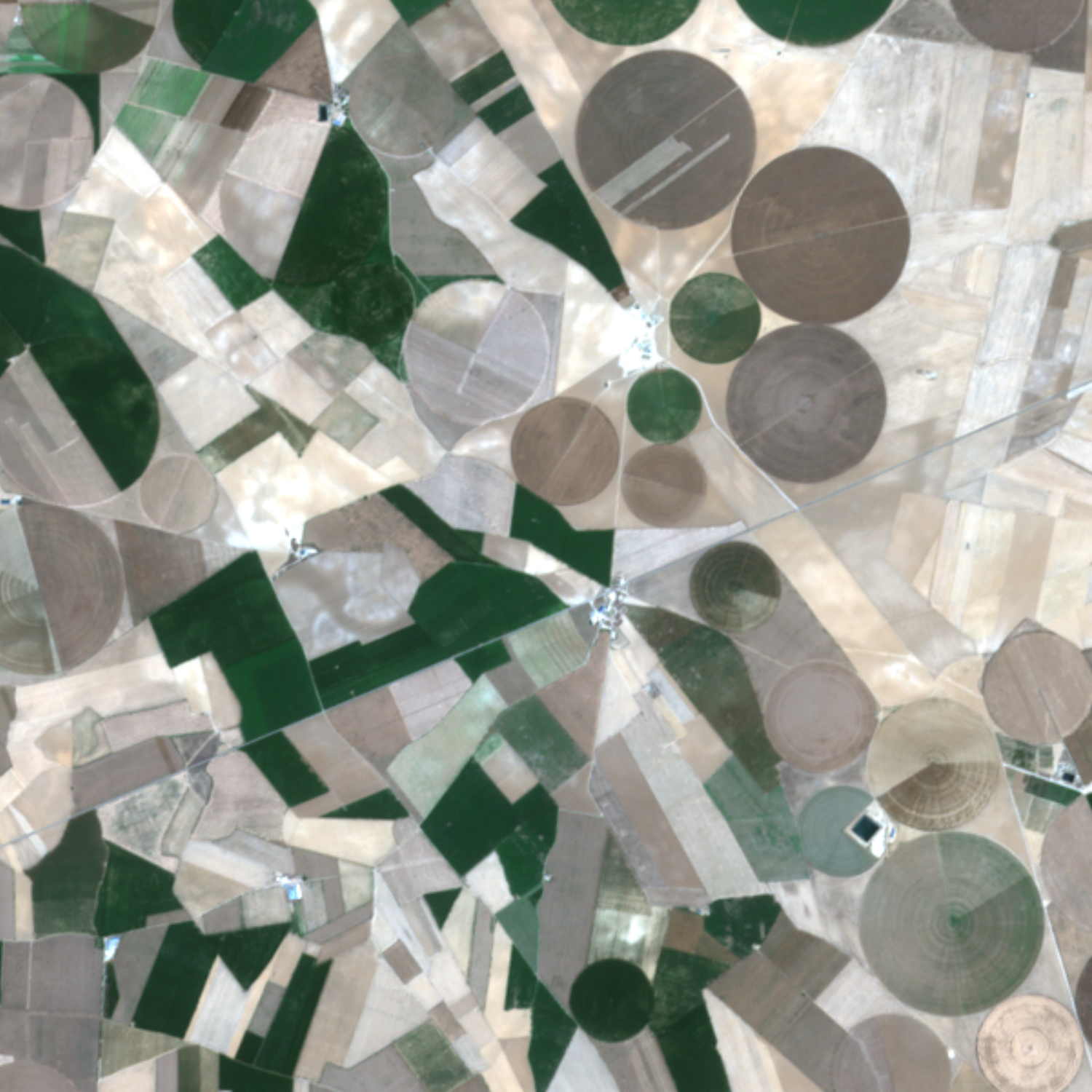}%
\label{Barrax Pre}}
\hfil
\subfloat[Post Image RGB]{\includegraphics[width=1.5 in]{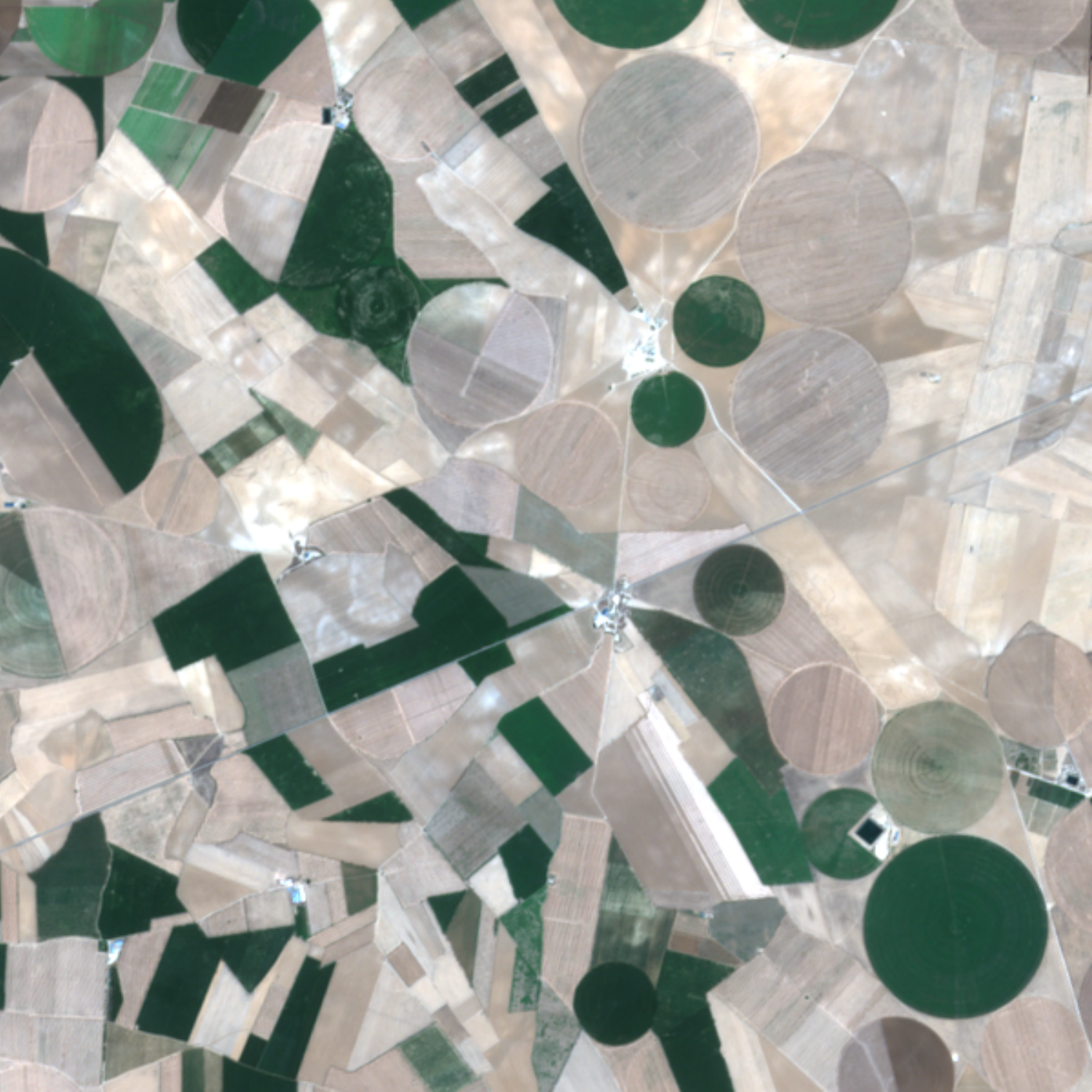}%
\label{Barrax Post}}
\hfil
\subfloat[Ground Truth]{\includegraphics[width=1.5 in]{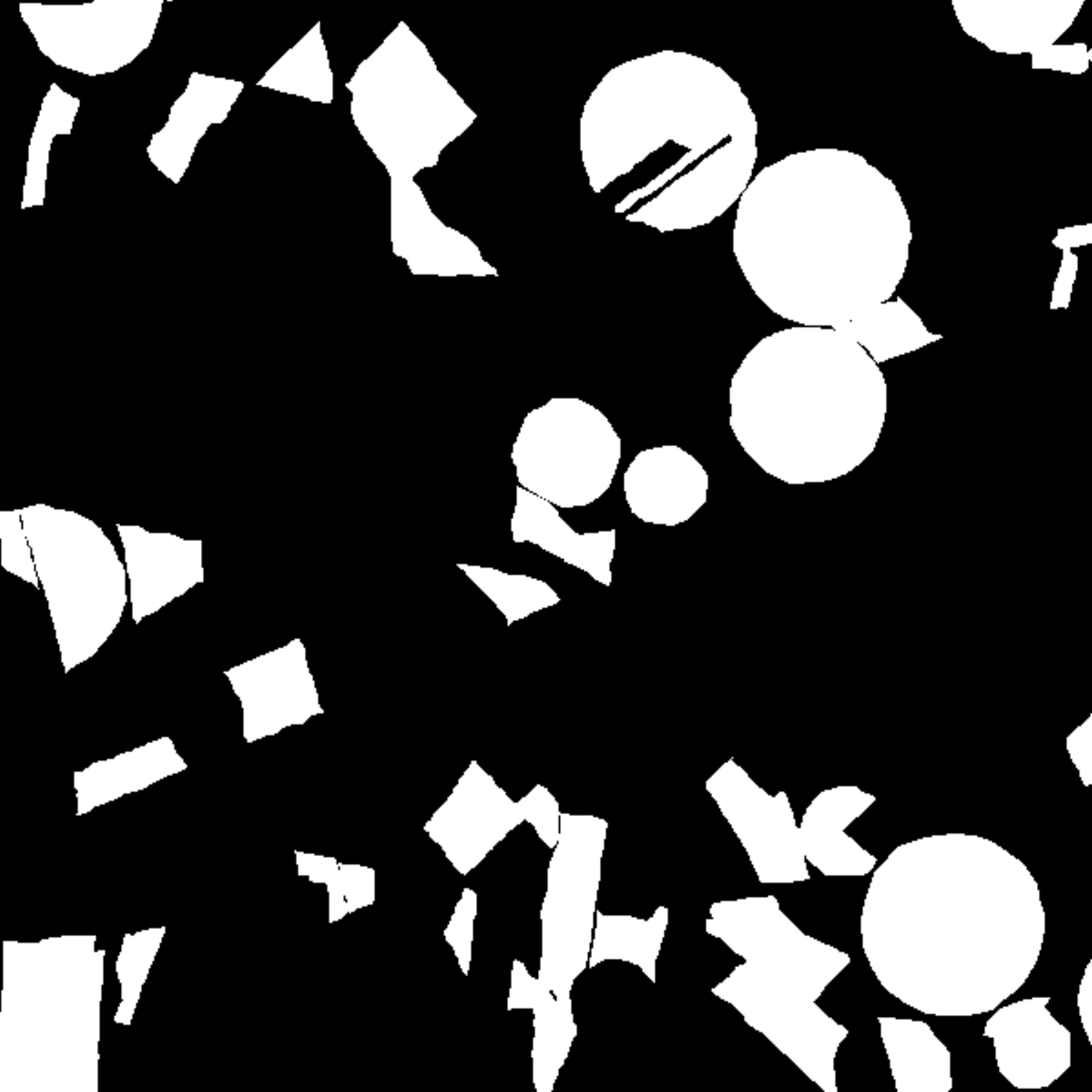}%
\label{Barrax Labels}}
\hfil
\subfloat[SiROC (NIR)]{\includegraphics[width=1.5 in]{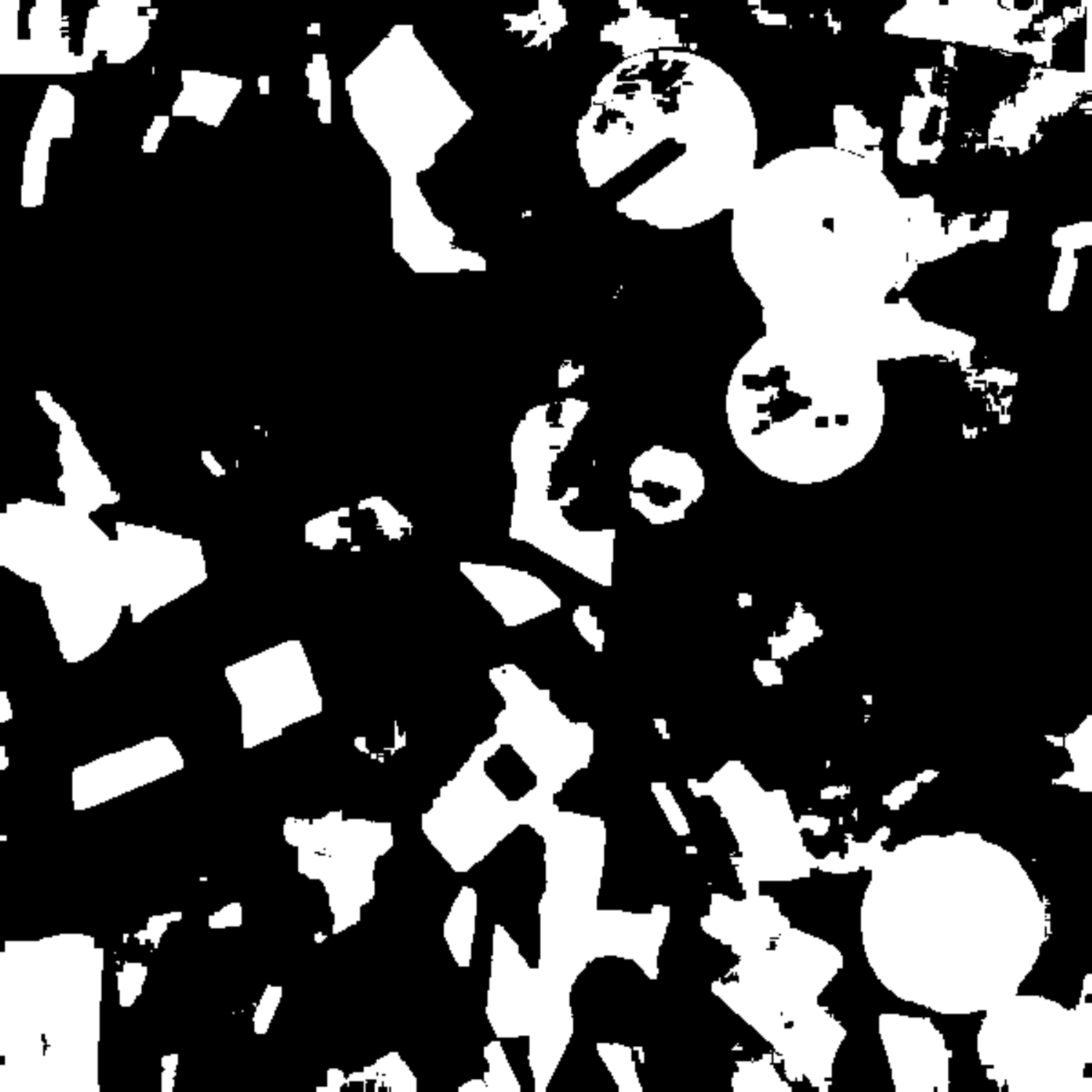}%
\label{SiROC Barrax CM NIR}}
\hfil
\subfloat[SiROC No MP (NIR)]{\includegraphics[width=1.5 in]{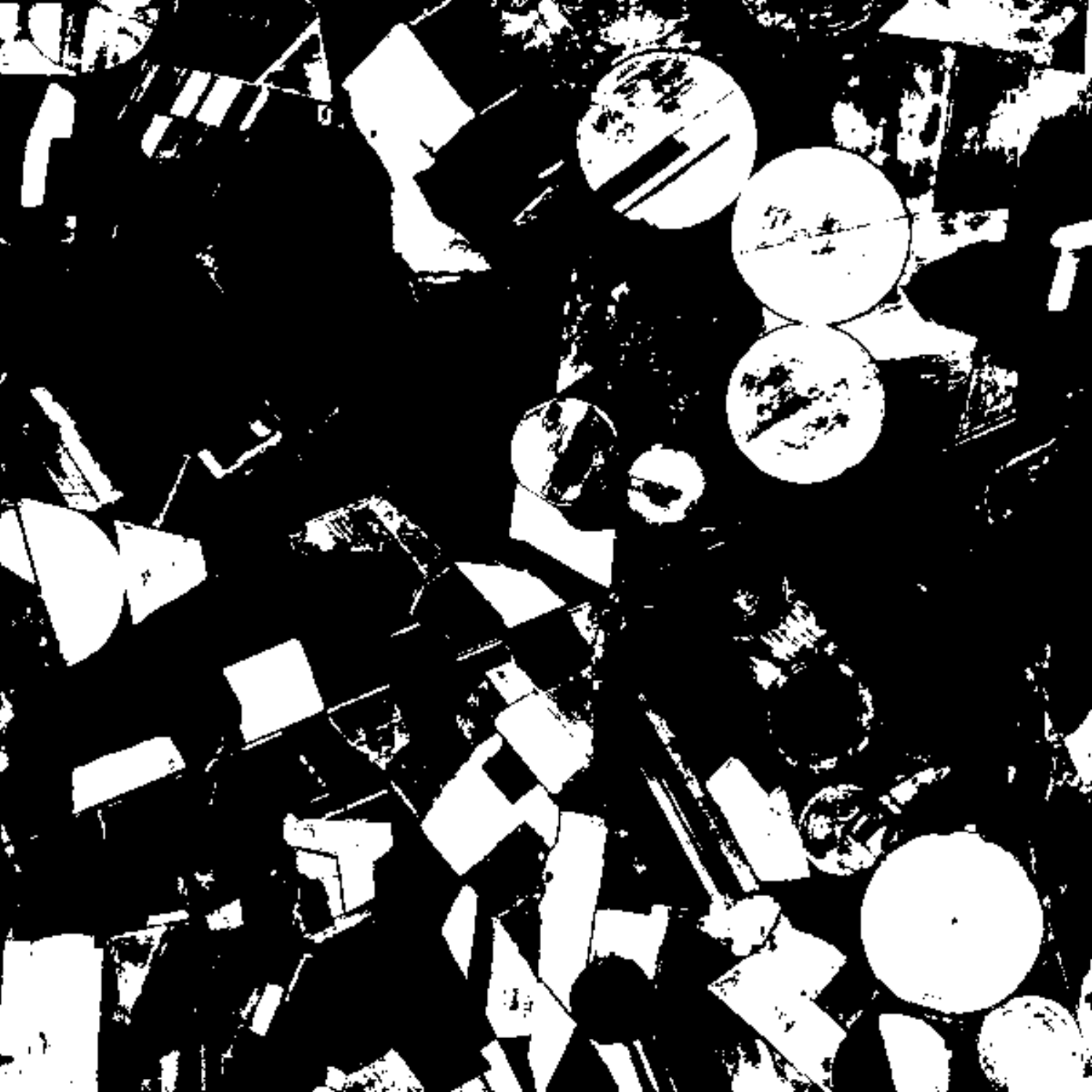}%
\label{SiROC Barrax CM No MP }}
\hfil
\subfloat[DCVAMR]{\includegraphics[width=1.5 in]{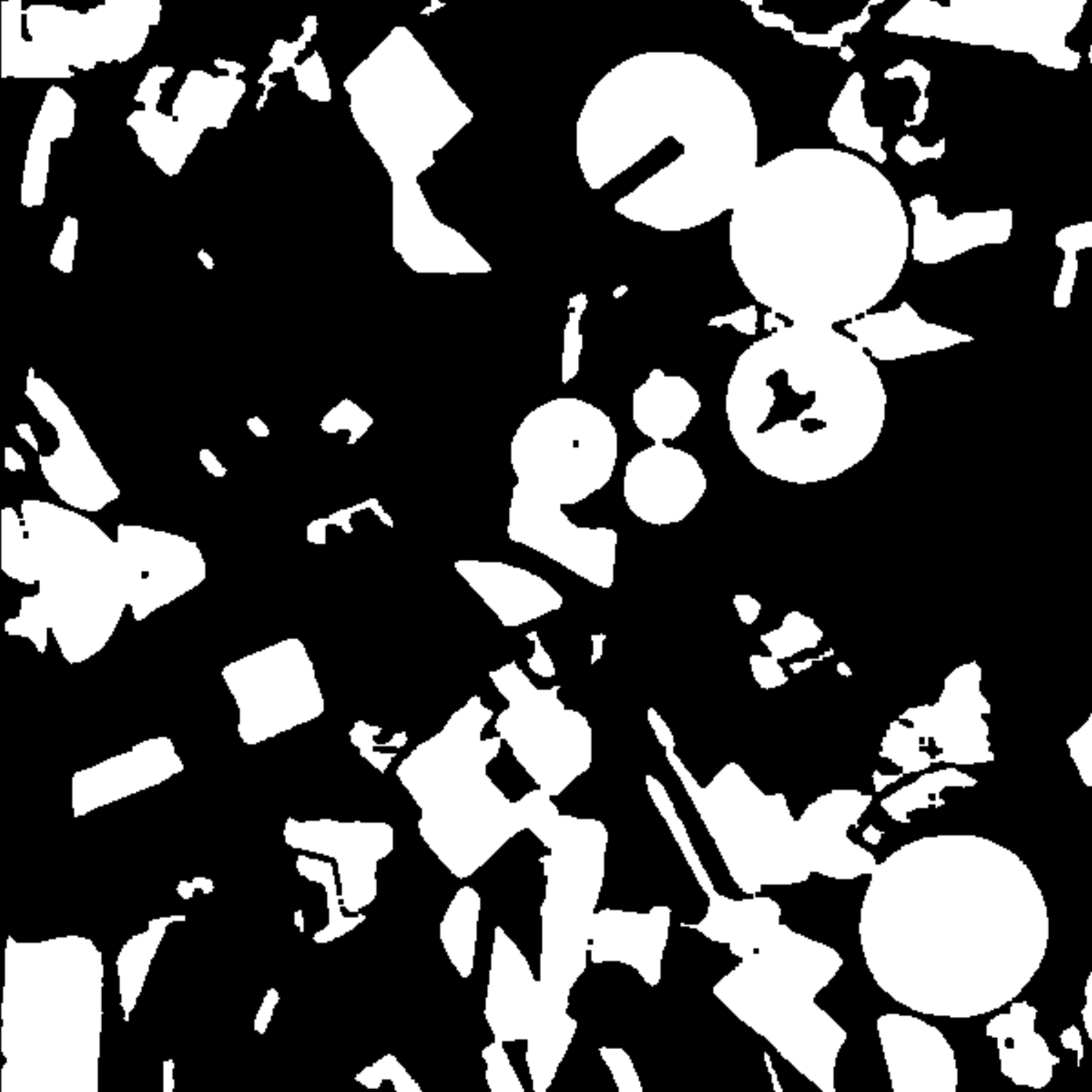}%
\label{DCVA Barrax}}
\hfil
\subfloat[PCVA (NIR)]{\includegraphics[width=1.5 in]{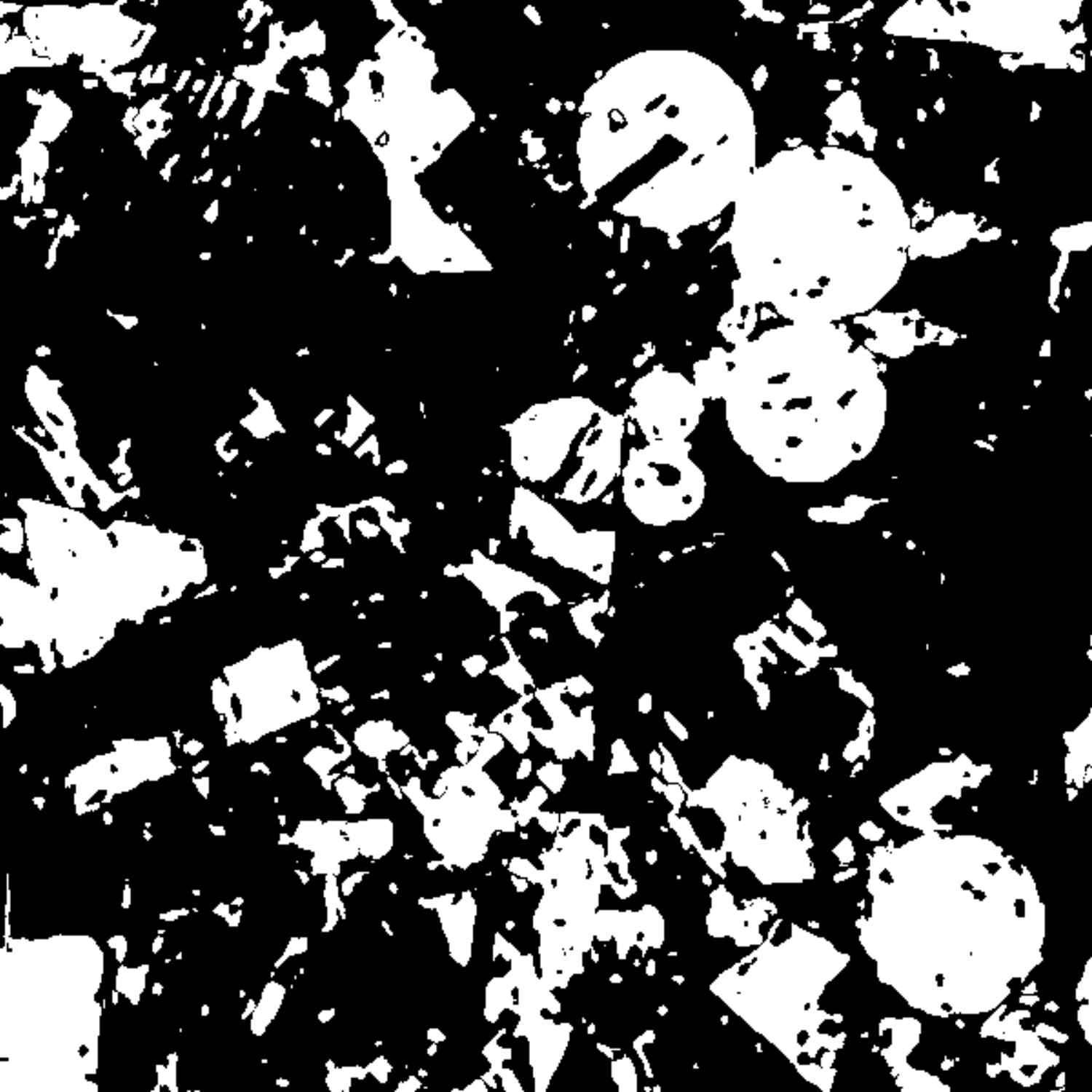}%
\label{PCVA NIR Barrax}}
\hfil
\subfloat[RCVA (NIR)]{\includegraphics[width=1.5 in]{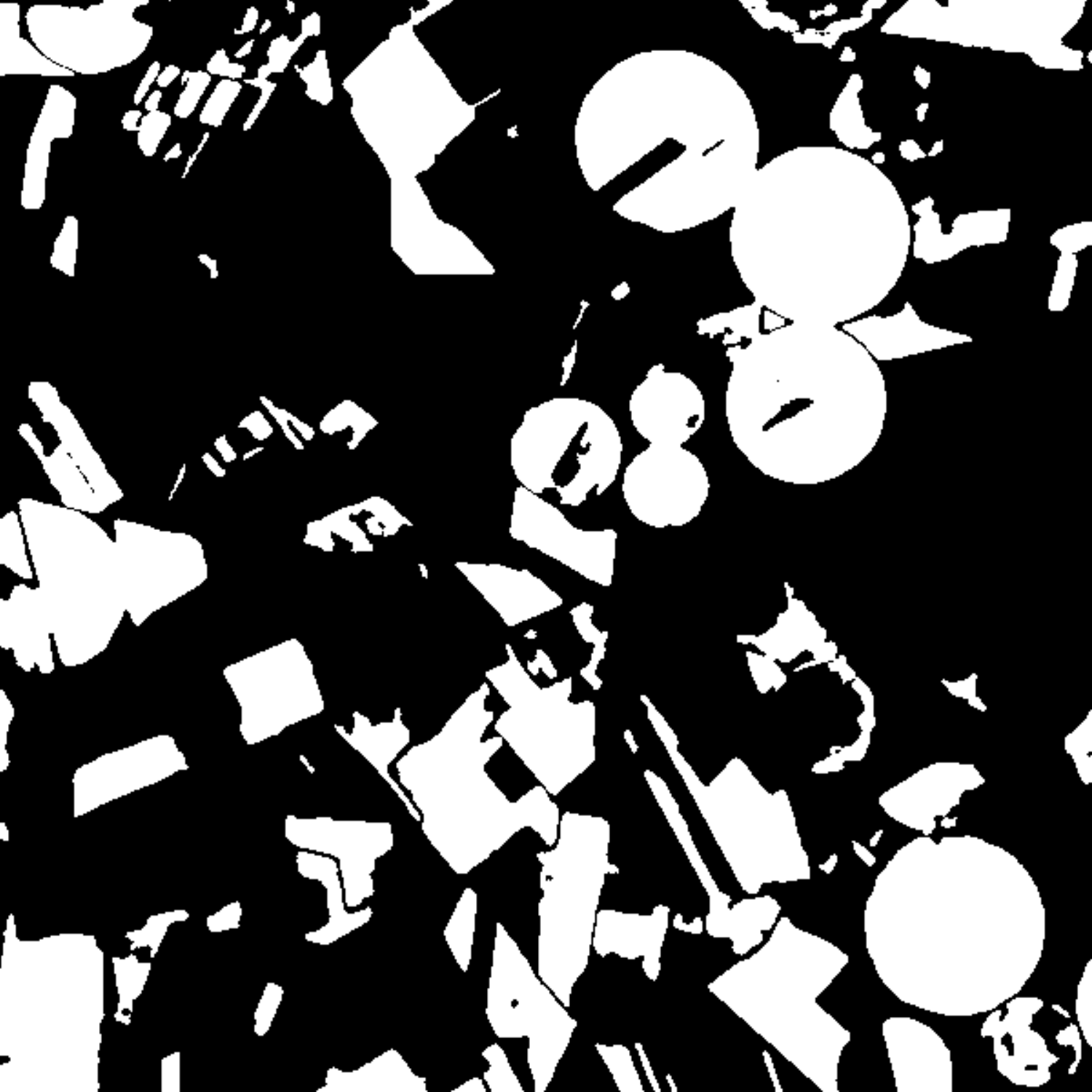}%
\label{RCVA NIR Barrax}}
\caption{Qualitative Results Agriculture Dataset. Pre (\ref{Barrax Pre}) and post RGB (\ref{Barrax Post}) image with changing agricultural fields. The ground truth (\ref{Barrax Labels}) shows high similarity to SiROC (\ref{SiROC Barrax CM NIR}) but also to DCVA (\ref{DCVA Barrax}) and RCVA (\ref{RCVA NIR Barrax}) while PCVA (\ref{PCVA NIR Barrax}) has some false positive areas and spurious change predictions. Change segmentations without morphological profiles \ref{SiROC Barrax CM No MP } for SiROC still works well which is line with the quantitative results of Table \ref{tab:Results_Agriculture}.}
\label{fig:Agriculture}
\end{figure*}

\textit{Qualitative Results}: Figure \ref{fig:Agriculture} presents pre and post RGB images (\ref{Barrax Pre}-\ref{Barrax Post}), the ground truth \ref{Barrax Labels}, and change predictions (\ref{SiROC Barrax CM NIR}-\ref{RCVA NIR Barrax}). The visual impression of change predictions confirms the quantitative results. Predictions are fairly accurate on this scene which suggests a comparably easy task relative to the more complex OSCD scenes. SiROC segments changing regions well and struggles with the varying field shapes only in rare instances. Similarly, the results of DCVAMR and RCVA are also fairly accurate with a slightly higher tendency to predict the change class. In comparison, the mask by PCVA produces some false positive regions. Overall, SiROC shows similar performance to highly effective methods also in the agriculture domain.

\begin{table}
\centering
\caption{Quantitative Results Agriculture Dataset}

\begin{threeparttable}
\footnotesize
\setlength{\tabcolsep}{\tabcolsep}
\begin{tabular}{lcccc}
\toprule
{} & Specificity & Sensitivity & Precision & F1 \\
\midrule
SiROC (VEG) & 90.69\% & 86.38\% & 73.53\% & 79.44\% \\
SiROC (NIR) & \textbf{90.81\%} & 88.70\% & \textbf{74.28\%} & 80.85\% \\
DCVAMR & 88.88\% & \textbf{94.26\%} & 71.73\% & \textbf{81.47\%} \\
PCVA (VEG)  & 88.83\% & 83.18\%  & 69.04\% & 75.45\% \\
PCVA (NIR) & 86.60\% & 84.56\% & 65.38\% & 73.74\% \\
RCVA (VEG)  & 88.91\% & 91.95\%  & 71.28\% & 80.31\% \\
RCVA (NIR) & 87.39\% & 92.36\% & 68.67\% & 78.77\% \\

\midrule
\multicolumn{5}{c}{Ablation Scores} \\
\midrule
No MP (VEG)  & 89.87\% & 84.66\%  & 71.44\% & 77.49\% \\
No MP (NIR)  & 89.70\% & 87.15\%  & 71.70\% & 78.67\%\\
HSR (VEG) & 89.96\% & 87.71\%  & 72.35\% & 79.29\%\\
HSR (NIR) & 89.29\% & 90.64\%  & 71.70\% & 80.06\% \\

\bottomrule
\end{tabular}\textbf{}
\end{threeparttable}
\label{tab:Results_Agriculture}
\end{table}

\subsection{Results on Alpine Dataset}
 \label{subsec:alpine}
\begin{table}
\centering
\caption{Quantitative Results Alpine Dataset}
\begin{threeparttable}
\footnotesize
\setlength{\tabcolsep}{\tabcolsep}
\begin{tabular}{lcccc}
\toprule
{} & Specificity & Sensitivity & Precision & F1 \\
\midrule
SiROC (NIR) & 98.92\% & 75.71\% & 52.28\% & 61.85\% \\
SiROC (SWIR) & \textbf{99.28\%} & 59.51\% & 56.10\% & 57.76\% \\
DCVAMR
& 99.06\% & \textbf{94.99\%} & \textbf{61.23\%} & \textbf{74.46\%} \\
PCVA (NIR)  & 98.95\% & 46.99\%  & 41.04\% & 43.82\% \\
PCVA (SWIR)  & 95.48\% & 35.80\%  & 10.98\% & 16.80\% \\
RCVA (NIR) & 99.22\% & 63.99\% & 56.20\% & 59.84\% \\
RCVA (SWIR) & 86.56\% & 66.71\% & 6.52\% & 11.89\% \\

\midrule
\multicolumn{5}{c}{Ablation Scores} \\
\midrule
No MP (NIR) & 97.81\% & 65.58\% & 31.74\% & 42.78\% \\
No MP (SWIR) & 97.39\% & 55.51\% & 24.87\% & 34.36\% \\
HSR (NIR) & 95.32\% & 82.10\% & 21.45\% & 34.01\% \\
HSR (SWIR) & 96.32\% & 66.12\% & 21.87\% & 32.87\% \\

\bottomrule
\end{tabular}\textbf{}
\end{threeparttable}
\label{tab:Results_Alpine}
\end{table}
\textit{Quantitative Results}:
Results for the Alpine dataset can be found in Table \ref{tab:Results_Alpine}. Even though SiROC reaches the highest specificity, it does not quite pass the overall performance of DCVAMR from \cite{saha2019unsupervised} on this dataset. Nevertheless, SiROC ranks highly also in sensitivity, precision and F1 score, particularly based on NIR inputs with total scores of 98.92\%, 75.71\% , 52.28\% and 61.85\%. RCVA with NIR inputs is comparable in performance but SiROC is the only method that makes effective use of SWIR inputs compared to PCVA and RCVA.

The ablation scores underline the effectiveness of morphological transformations with about a 20 pp. drop in F1 score compared to SiROC for both NIR and SWIR. Removing the ensembling leads to a notable drop in F1 scores, particularly with NIR inputs.

\begin{figure*}[!htbp]
\centering
\subfloat[False Color Composite (SWIR pre, SWIR post, SWIR pre)]{\includegraphics[width=1.5 in]{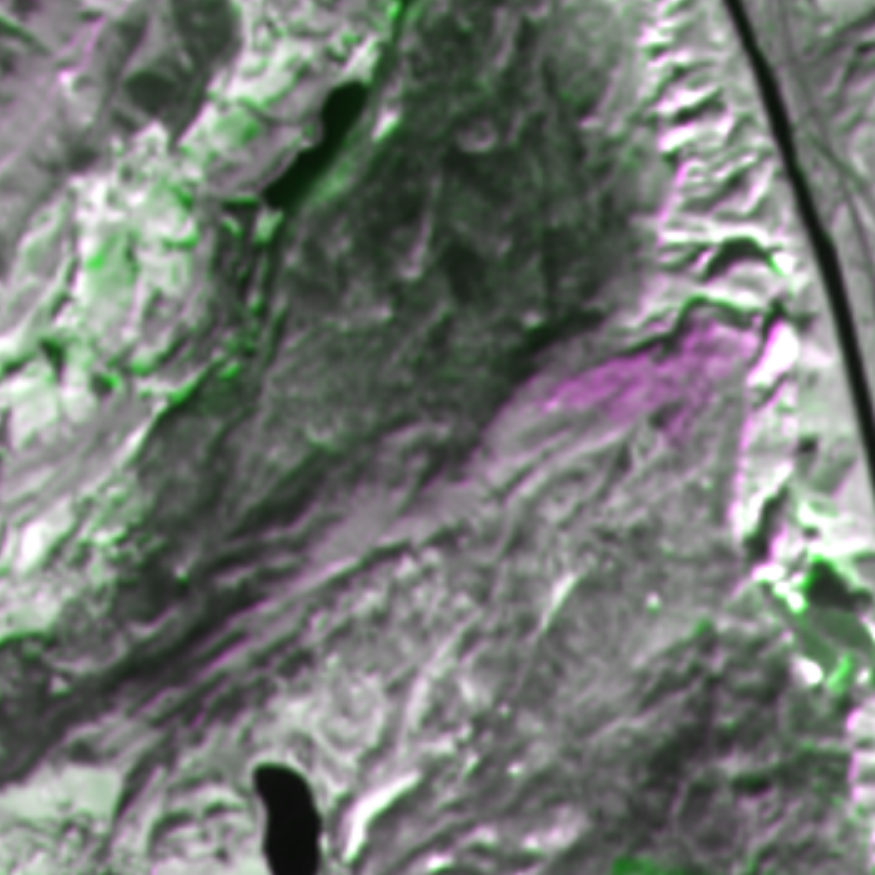}%
\label{Lamar FCC}}
\hfil
\subfloat[Ground Truth]{\includegraphics[width=1.5 in]{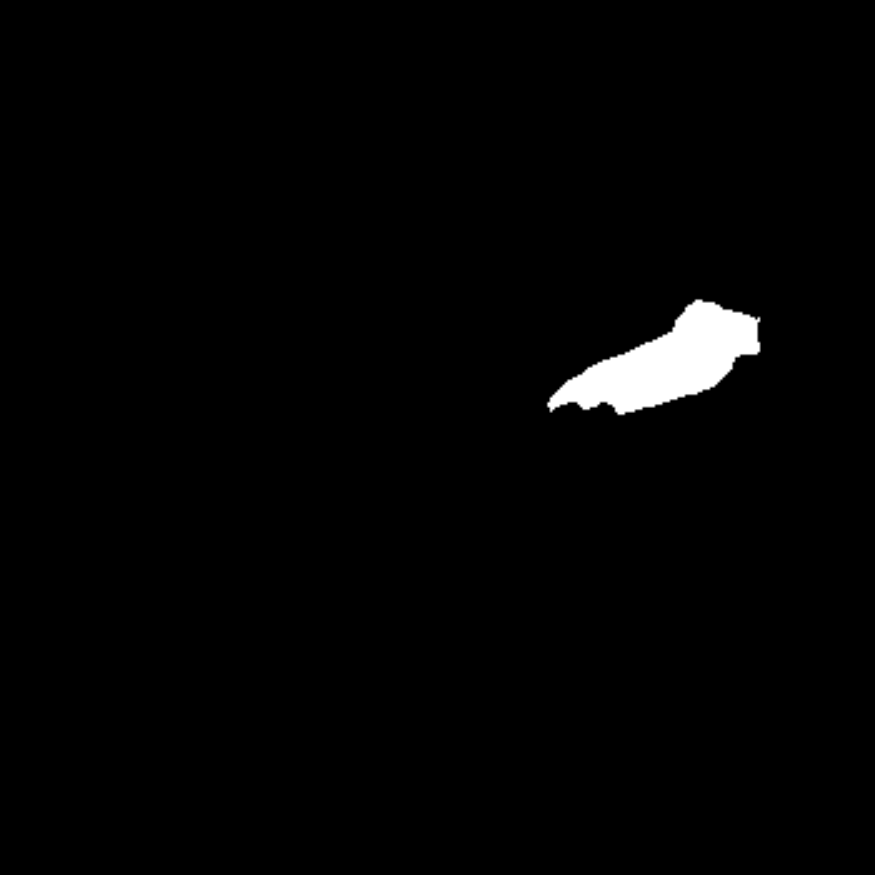}%
\label{Lamar Labels}}
\hfil
\subfloat[SiROC (NIR)]{\includegraphics[width=1.5 in]{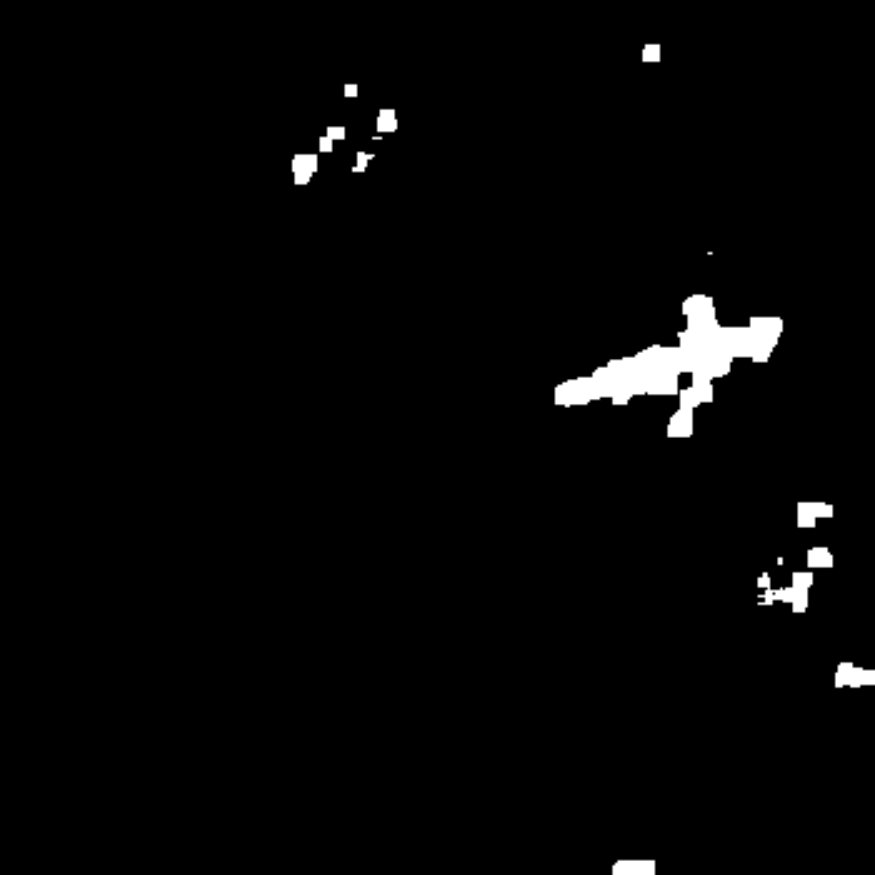}%
\label{SiROC Lamar CM NIR }}
\hfil
\subfloat[SiROC No MP (NIR)]{\includegraphics[width=1.5 in]{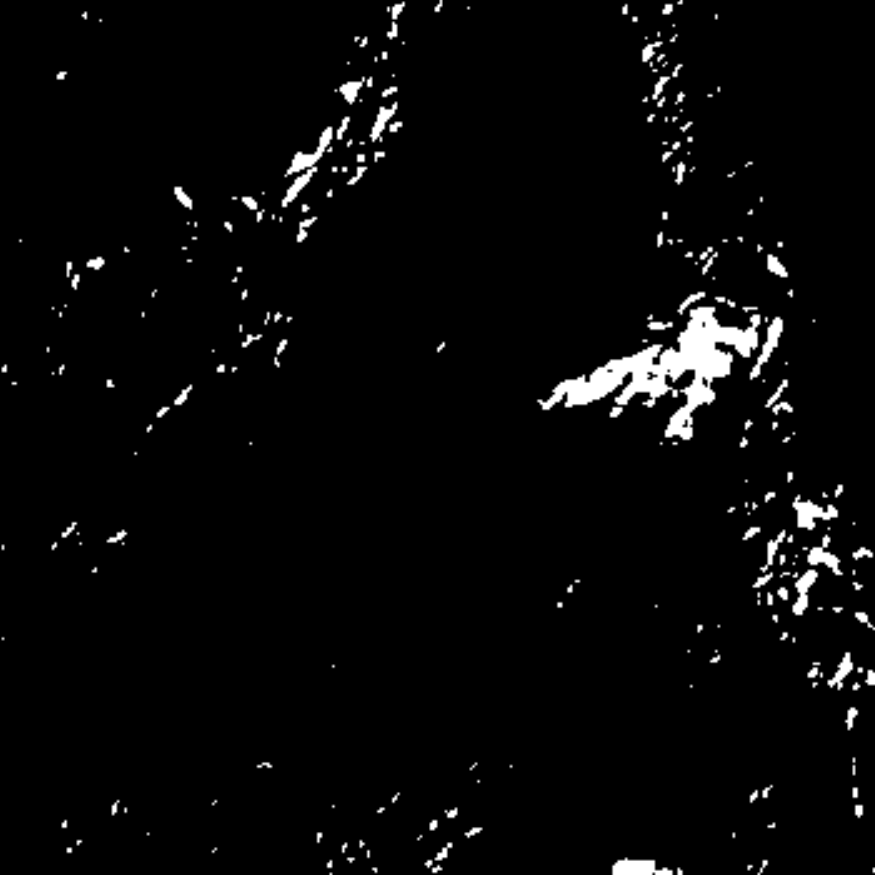}%
\label{SiROC Lamar Nir No MP}}
\hfil
\subfloat[SiROC (SWIR)]{\includegraphics[width=1.5 in]{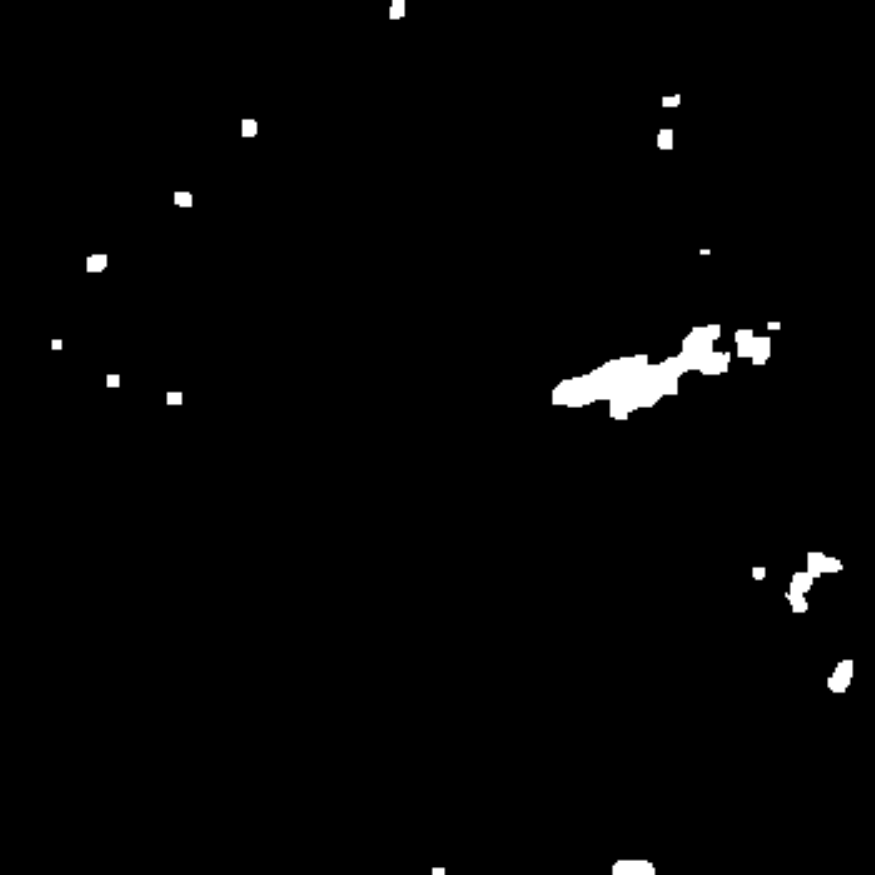}%
\label{SiROC Lamar CM swIR }}
\hfil
\subfloat[DCVAMR]{\includegraphics[width=1.5 in]{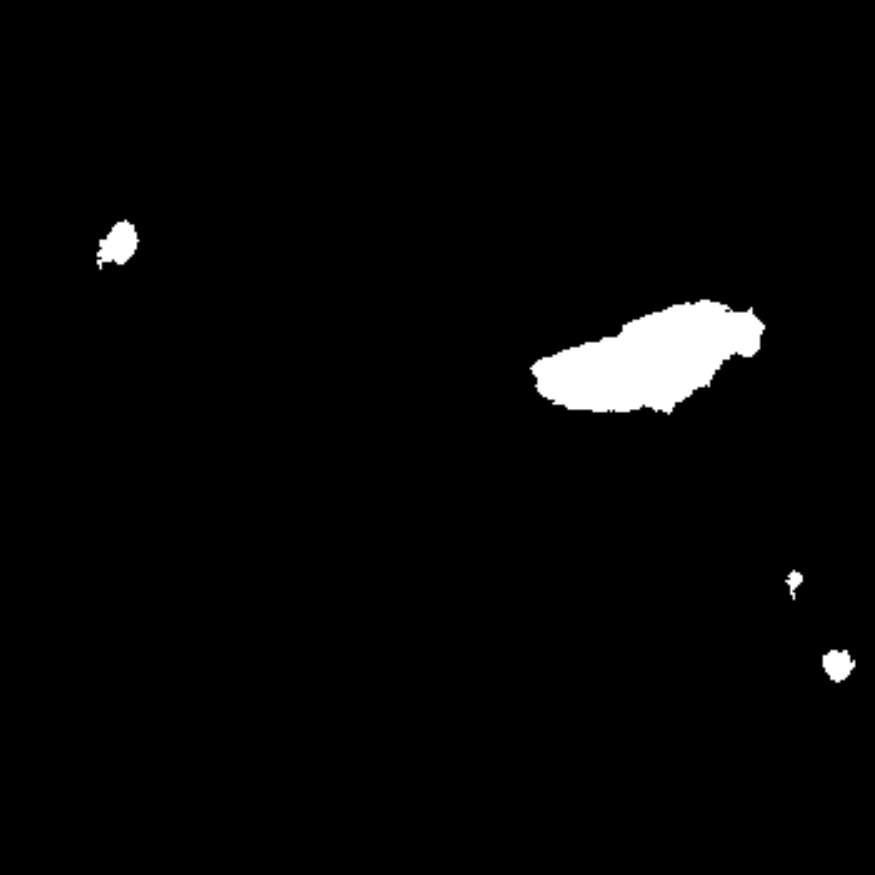}%
\label{Lamar DTL}}
\hfil
\subfloat[PCVA (NIR)]{\includegraphics[width=1.5 in]{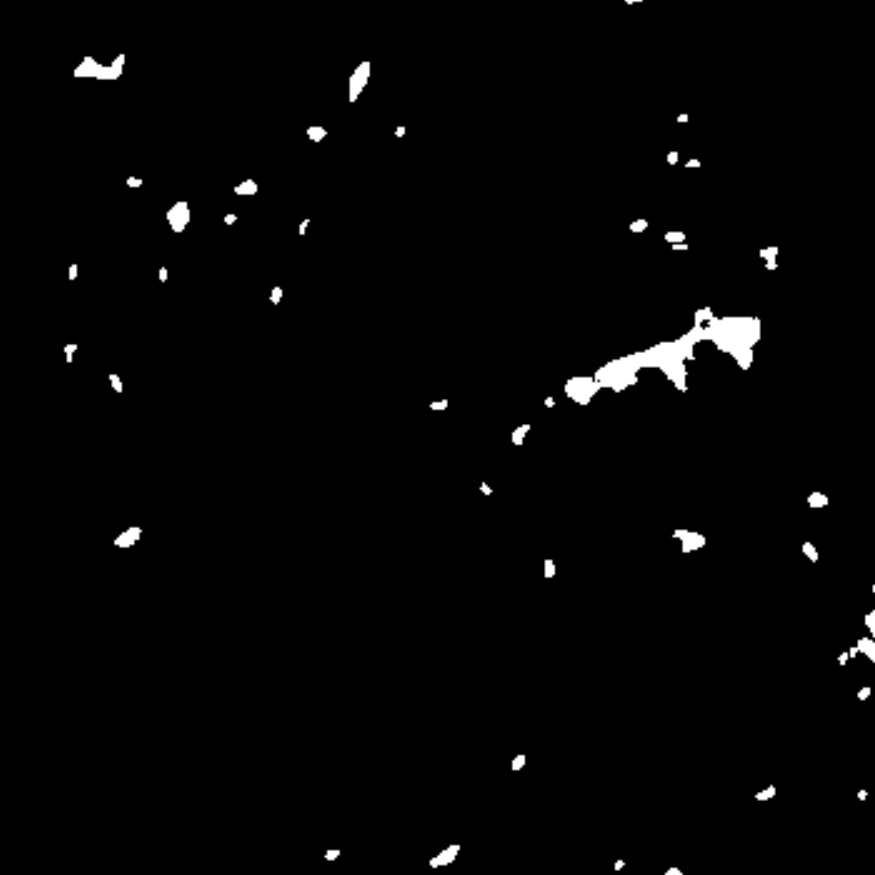}%
\label{Lamar PCVA}}
\hfil
\subfloat[RCVA (NIR)]{\includegraphics[width=1.5 in]{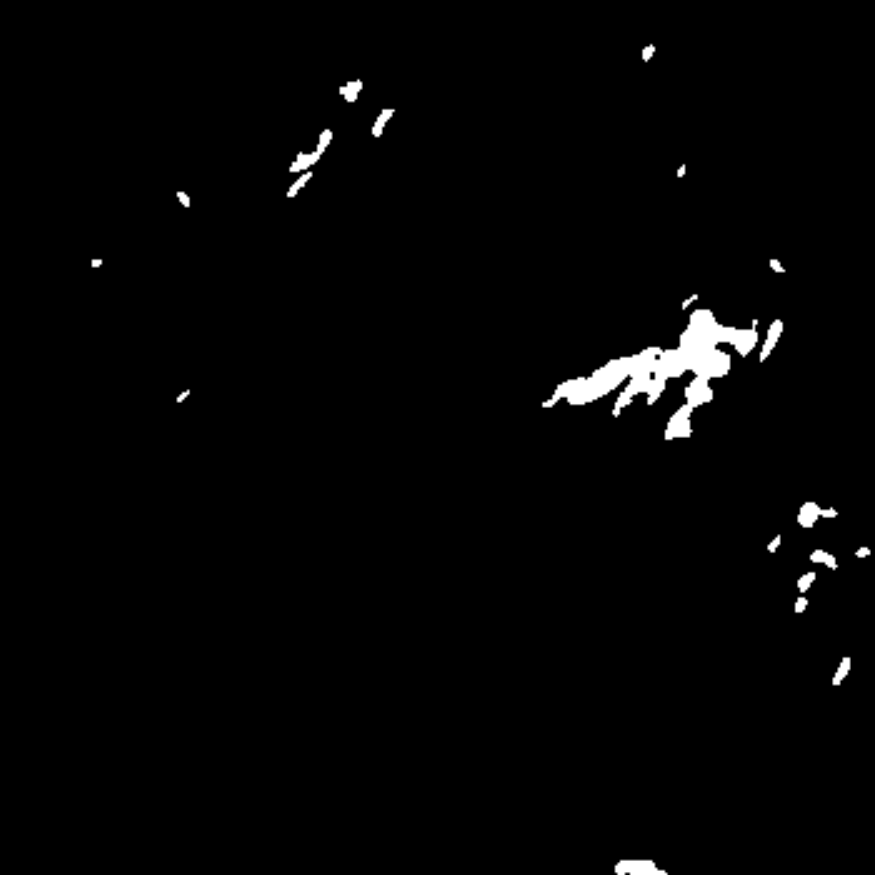}%
\label{Lamar RCVA}}
\caption{Qualitative Results Alpine Dataset. The area of the fire can be seen in purple in the false color composite in Panel \ref{Lamar FCC} with the reference map in \ref{Lamar Labels}. SiROC (NIR) (\ref{SiROC Lamar CM NIR }) and SiROC (SWIR) (\ref{SiROC Lamar CM swIR }) both identify the changing area well although the shape is better approximated with NIR inputs. Without morphological profiles SiROC picks up more false positives \ref{SiROC Lamar Nir No MP}. DCVAMR shows the most convincing results here (\ref{Lamar DTL}). RCVA is roughly comparable (\ref{Lamar RCVA}) to SiROC while PCVA falls behind slightly (\ref{Lamar PCVA}).}
\label{fig:Alpine}
\end{figure*}

\textit{Qualitative Results}:
Figure \ref{fig:Alpine} plots prediction masks for selected models for the Alpine dataset. In Panel \ref{Lamar FCC}, the false color composite shows the annotated area of change affected (\ref{Lamar Labels}) by a fire in purple on the right. SiROC identifies this well although it is tempted to also classify a small number of false positives as change. While it is hard to control for seasonality in a bitemporal setting, SiROC (NIR) (\ref{SiROC Lamar CM NIR }) still excludes most other vegetation updates which are not the result of actual change here. The morphological profiles help on this scene to exclude spurious predictions (\ref{SiROC Lamar Nir No MP}). Compared to SiROC (SWIR) (\ref{SiROC Lamar CM swIR }), SiROC (NIR) segments the changing area slightly better although the shape is identified more clearly by DCVAMR (\ref{Lamar DTL}). PCVA (NIR) (\ref{Lamar PCVA}) seems to struggle slightly more with the shape of the burned area whereas the results of RCVA (NIR) (\ref{Lamar RCVA}) look similar to the results of SiROC (NIR) which is in line with the quantitative scores of Table \ref{tab:Results_Alpine}.

\section{Discussion}
SiROC is an effective method for CD in medium- as well as high-resolution optical imagery which achieves competitive performance on four datasets. In the following, we elaborate on the intuition of SiROC's performance. When contrasting SiROC to image differencing methods, SiROC can be interpreted as an improvement over standard image differencing techniques because it does not assume the same changes in the acquisition conditions across time for the whole image. Rather, it allows for local changes in acquisition conditions. In standard CVA or RCVA, for example, an implicit assumption is that changes in the acquisition conditions across time affect each pixel similarly. SiROC releases this restriction and instead allows for local trends in regions of the image. If a pixel deviates from the local trend around it, it is likely to undergo a change in SiROC. In RCVA or CVA, one would compare this pixel against trends in the whole image and not against its surrounding only. This might be unrealistic in complex scenes where pixels values highly depend on local trends in the surroundings. This is for example the case when a new building casts a shadow on a previously illuminated pixel. Similarly, a cloudy pixel in $t+1$ that was unobstructed in $t$ might not necessarily be changing and is rather influenced by the local trend of a cloud rather than general image trends if large parts of the image are not obstructed by clouds. Hence, SiROC allows for a more granular analysis of deviations from trends in an image time-series because compared to previous methods it makes full use of multi-temporal information in close as well as distant neighbors. 
Although we compare our results to deep learning-based methods, our intention is rather to augment these models than replace them, especially with high-resolution images. SiROC provides an efficient and accurate way to obtain change labels that could also be infused into deep learning models. One application of SiROC could be in self-supervised learning where pseudo-labels are often obtained based on traditional image differencing techniques such as CVA \cite{dong2020self}. SiROC is not only superior in performance compared to image differencing. It also comes with a built-in, well-calibrated uncertainty of predictions. This could be especially beneficial in self-supervised settings since it automatically allows to discriminate pseudo-labels by confidence. For example, one could train only based on pseudo labels with high certainty and discard uncertain data points. 
Similarly, in some unsupervised methods such as MSDRL for VHR imagery, an initial pseudo-classification is separated by confidence where high confidence examples are used for training a classifier that subsequently obtains predictions for leftover uncertain pixels \cite{zhan2020unsupervised}. In these methods, SiROC could also be used to obtain initial predictions and uncertainties to potentially improve not only the initial classification but maybe also the uncertainty categorization. The combination of deep learning-based methods and SiROC may hence open up new potential for CD methods. While we restrict our focus to CD with optical images here, the framework of SiROC may be extended for applications on other multitemporal CD problems in remote sensing as well. 

\section{Conclusion} \label{sec:6}
We present SiROC, an efficient and accurate unsupervised method for CD in medium- and high-resolution optical images. SiROC is inspired by HSR which is used for exoplanet search in astronomy. It models a pixel of interest in $t$ as a linear combination of its neighbors and applies this model to $t+1$ to obtain a prediction for the pixel based on its neighbors. The difference of the prediction for $t+1$ and the actual pixel value in $t+1$ is interpreted as the change signal. If the prediction is far from the actual value, trends in the neighboring pixels divert from the difference in the pixel of interest over time which is seen as an indicator for change on the ground. 

We refine and extend HSR in two major ways to apply it to optical satellite images as SiROC. First, we iterate over several, mutually exclusive neighborhoods and apply HSR with all of these neighborhoods as input to obtain a distribution of change predictions. We combine these predictions with majority voting which improves performance significantly and also returns a heatmap of votes per pixel which can be interpreted as a well-calibrated uncertainty. Second, we use morphological opening and closing at one spatial filter scale to transition from pixel-level to object-level predictions. 

The results of SiROC are validated on four datasets. For urban change detection with medium-resolution images, we verify the effectiveness of our method on OSCD, which contains binary change annotations for 24 cities across the globe. SiROC sets a new state-of-the-art for unsupervised CD on OSCD which surpasses previous methods by 10 p.p. in terms of specificity, 2 p.p. in sensitivity, 11 p.p in precision, and 13 p.p. in F1 score. We further validate the performance of SiROC on high-resolution images with a dataset on the Beirut Harbor Explosion (BHED). Also on this dataset, SiROC surpasses the performance of competing methods and underlines its abilities to segment urban change accurately at several scales. Further, we provide two validation exercises on non-urban data with Sentinel-2 inputs. SiROC segments the effects of a fire in the Italian Alps accurately and in the range of competing methods. On the Agriculture dataset, SiROC falls short of DCVAMR in overall scores but still identifies the changing crop activity correctly.

While SiROC compares well against current deep learning-based unsupervised methods in CD, SiROC should rather be seen as a complement than a substitute to these methods. Since it provides an accurate way to predict change signals with a built-in, well-calibrated uncertainty it may be especially useful in conjunction with deep learning-based methods to generate pseudo-labels. Although we apply SiROC primarily to changes with multispectral data, the model may be applicable to other change detection problems as well which we plan to explore in future research.

\section*{Acknowledgment}
We are grateful to the Center for Satellite Based Crisis Information (ZKI) of the German Aerospace Center for providing us with ground truth of the Beirut Explosion scene.

\ifCLASSOPTIONcaptionsoff
  \newpage
\fi

\bibliographystyle{IEEEtran}
\bibliography{manuscript}

\end{document}